\documentclass[preprint,12pt]{elsarticle}

\usepackage{amssymb}

\usepackage{amsmath}
\usepackage{amsthm}

\usepackage{times}
\usepackage[normalem]{ulem}
\usepackage{xcolor}
\usepackage{url}
\usepackage[hidelinks]{hyperref}
\usepackage[utf8]{inputenc}
\usepackage{marvosym}
\usepackage[small]{caption}
\usepackage{graphicx}
\usepackage{booktabs}
\usepackage{algorithm}
\usepackage[switch]{lineno}

\usepackage{balance} 
\usepackage[T1]{fontenc}    
\usepackage{amsfonts}       
\usepackage{nicefrac}       
\usepackage{microtype}      
\usepackage{xcolor}         

\usepackage{algorithm}      
\usepackage{algpseudocode}  

\usepackage{verbatim}
\usepackage{multirow}
\usepackage{capt-of}
\usepackage{subcaption}
\usepackage{caption}

\usepackage{fancyhdr}
\usepackage{diagbox}
\usepackage{makecell}
\usepackage{rotating}
\usepackage{enumitem}
\usepackage{lineno}

\usepackage[table]{xcolor}

\usepackage[markup=default,authormarkup=none,commandnameprefix=always]{changes}
\definechangesauthor[name={Chenran}, color=red]{cr}



\journal{Neural Networks}

\begin{document}

\begin{frontmatter}

\title{Transferable Delay-Aware Reinforcement Learning via Implicit Causal Graph Modeling}




\author[1]{Chenran Zhao\fnref{equal}}
\author[1,2]{Dianxi Shi\corref{cor1}\fnref{equal}}
\author[3]{Yaowen Zhang}
\author[2]{Chunping Qiu}
\author[1]{Shaowu Yang}

\fntext[equal]{These authors contributed equally to this work.}
\cortext[cor1]{Corresponding author.}

\affiliation[1]{organization={College of Computer Science and Technology, National University of Defense Technology},
	city={Changsha},
	country={China}}

\affiliation[2]{organization={Intelligent Game and Decision Lab (IGDL)},
	city={Beijing},
	country={China}}

\affiliation[3]{organization={Institute of Military Transportation},
	city={Tianjin},
	country={China}}

\affiliation[4]{organization={School of Artificial Intelligence, Hebei University of Technology},
	city={Tianjin},
	country={China}}

\begin{abstract}
Random delays weaken the temporal correspondence between actions and subsequent state feedback, making it difficult for agents to identify the true propagation process of action effects. In cross-task scenarios, changes in task objectives and reward formulations further reduce the reusability of previously acquired task knowledge. To address this problem, this paper proposes a transferable delay-aware reinforcement learning method based on implicit causal graph modeling. The proposed method uses a field-node encoder to represent high-dimensional observations as latent states with node-level semantics, and employs a message-passing mechanism to characterize dynamic causal dependencies among nodes, thereby learning transferable structured representations and environment dynamics knowledge. On this basis, imagination-driven behavior learning and planning are incorporated to optimize policies in the latent space, enabling cross-task knowledge transfer and rapid adaptation. Experimental results show that the proposed method outperforms baseline methods on DMC continuous control tasks with random delays. Cross-task transfer experiments further demonstrate that the learned structured representations and dynamics knowledge can be effectively transferred to new tasks and significantly accelerate policy adaptation.

\end{abstract}

\begin{keyword}
Reinforcement Learning, Delay-Aware, Causality 

\end{keyword}

\end{frontmatter}




\section{Introduction}

In many real-world reinforcement learning scenarios, the feedback caused by an action may not be observed immediately. Instead, its effects can appear after a delay in later states, observations, or rewards. This temporal mismatch makes it difficult for the agent to infer the true influence of its actions on environmental evolution. The problem is further complicated by random delays, where the delay duration changes over time and the correspondence between actions and subsequent feedback becomes uncertain.

This difficulty becomes more pronounced in cross-task transfer. Different tasks may have different objectives and reward functions, while sharing similar underlying environmental dynamics. To adapt efficiently, an agent should learn structural knowledge about the environment that can be reused across tasks, rather than relying only on task-specific policies or correlations. However, random delays may obscure the underlying action-effect propagation process and cause learned knowledge to depend on particular delay patterns. Therefore, how to \textbf{learn stable and transferable structural knowledge in random-delay environments} becomes the central problem addressed in this paper.

Causal model-based reinforcement learning provides an important entry point for this problem. On the one hand, \textbf{world models}\textsuperscript{\cite{hafner2019learning,hafner2019dream,hafner2020mastering,hafner2023mastering}} explicitly characterize environmental dynamics, provide internal representations of environmental evolution for policy learning, and preserve relatively stable and reusable regularities in this process. They therefore have the potential to support transfer and generalization. On the other hand, \textbf{causal models}\textsuperscript{\cite{gao2025causal,poudel2022contrastive,hao2023latent}} help reveal structurally meaningful dependencies among variables, enabling the world model to go beyond black-box fitting of state transitions and to learn dynamic representations that are closer to the intrinsic mechanisms of the environment. However, most existing CMBRL methods are not designed specifically for delayed scenarios, and thus still lack targeted mechanisms for jointly characterizing environmental dynamics and observation-delay effects.

To address the above issues, this paper proposes a transferable delay-aware reinforcement learning method via implicit causal graph modeling, referred to as \textbf{CausalDreamer}. The method represents high-dimensional observations as multiple field nodes with relatively independent semantics, and uses implicit causal dependencies among these nodes to characterize the dynamic evolution of environmental states, as well as the propagation and manifestation of state changes under random delays. As a result, the learned model is no longer limited to superficial state-transition relations constrained by a particular delay pattern in a specific task. Instead, it can further capture reusable structural dynamics shared by related random-delay tasks, thereby supporting cross-task transfer. Specifically, the proposed method consists of two coordinated modules. \textbf{The first is an implicit-causal-graph-based world model}. This module uses a field-node encoder to represent raw high-dimensional observations as latent variables with node-level semantics, and models dynamic dependencies among these nodes in latent space through an implicit causal graph. Meanwhile, a delay-alignment mechanism is incorporated into the state-transition model to characterize state evolution and delayed observation manifestation in random-delay environments. Based on this representation, decoding and prediction modules further perform observation reconstruction and reward and termination prediction. Through this process, the model learns relatively stable structural dependencies and dynamic regularities in the environment, forming a knowledge representation that can be reused within the same dynamics family. \textbf{The second is an imagination-based behavior learning and planning module}. This module rolls out imagined trajectories in latent space using the learned world model, and performs value learning, policy optimization, and exploration based on these trajectories. In this way, the agent can evaluate the long-term effect of actions before delayed feedback is fully observed. Since behavior learning is built upon a structured dynamics representation, the node-level interaction knowledge learned by the model can continue to be reused when the task objective, reward form, or local observation changes, as long as the environmental dynamics and delay-propagation characteristics remain similar. This supports cross-task transfer and rapid adaptation in random-delay environments.

We systematically evaluate CausalDreamer on multiple continuous-control tasks from the DeepMind Control Suite (DMC)~\textsuperscript{\cite{tunyasuvunakool2020dm_control}}, and compares it with baselines including DreamerV3\textsuperscript{\cite{hafner2023mastering}}, CWM\textsuperscript{\cite{yu2023explainable}}, and SAC\textsuperscript{\cite{haarnoja2018soft}}. The experimental results show that CausalDreamer generally outperforms the baselines in standard tasks without delays, and its performance degrades much less after random action delays are introduced. This indicates that the state representations and dynamic knowledge learned by CausalDreamer are more robust to delay disturbances. More importantly, the cross-task transfer experiments show that, whether transferring from easier tasks to harder ones or from harder tasks to easier ones, training initialized with transferred knowledge generally outperforms training from scratch, and the final performance is usually no worse than the latter and is even better in some tasks. This suggests that CausalDreamer learns not merely a local policy for a single task, but structural dynamic knowledge that can be reused within the same dynamics family. Further ablation experiments show that the field-node encoder and the message-passing mechanism are crucial for forming such transferable representations, thereby providing a more stable foundation for knowledge reuse across random-delay tasks.

\section{Related Works}
\subsection{Model-Based Reinforcement Learning}\label{sec:related_work_mbrl}

Model-based reinforcement learning (MBRL) introduces environment modeling into policy learning by explicitly constructing environment dynamics models or reward prediction models. Its core idea is to use the learned model to predict future state trajectories and potential returns, and then perform planning, policy evaluation, or imagination-based training, rather than relying entirely on direct interaction with the real environment. Following this idea, representative methods such as PlaNet\textsuperscript{\cite{hafner2019learning}}, Dreamer\textsuperscript{\cite{hafner2019dream}}, DreamerV2\textsuperscript{\cite{hafner2020mastering}}, and DreamerV3\textsuperscript{\cite{hafner2023mastering}} have gradually established the basic paradigm of learning representations, generating imagined trajectories, and optimizing policies based on latent-space world models. This line of work has also made world models one of the central technical routes in model-based reinforcement learning. On this basis, recent studies have further evolved along several directions, including modeling objectives, task-relevant representation refinement, model-bias mitigation, and joint optimization of models and policies. \textbf{From the perspective of delay problems}, these methods provide an informative modeling viewpoint for understanding long-horizon decision-making under delayed feedback, because they emphasize explicit prediction of future environmental evolution.

\textbf{The first category consists of methods that improve representation learning and modeling objectives.} These methods do not change the basic paradigm of model-based reinforcement learning, but focus on improving the representation quality, task relevance, and training stability of the world model itself, so that it can better support subsequent decision-making. HarmonyDream\textsuperscript{\cite{ma2023harmonydream}} treats world model learning as a dynamic balance between observation modeling and reward modeling, and learns latent representations that are more useful for control by adaptively adjusting the loss weights of the two objectives. AWM\textsuperscript{\cite{ma2024transformer}} analyzes the gradient propagation path and points out that, in long-horizon tasks, action-conditioned prediction is more beneficial for stable optimization than state autoregressive modeling. $\Delta$-IRIS\textsuperscript{\cite{micheli2024efficient}} uses a context-aware incremental tokenization mechanism to compress redundant representations in visual sequences, thereby reducing the cost of long-horizon modeling. SAM\textsuperscript{\cite{ramasubramanianimproving}} further introduces sharpness-aware minimization into world model training to mitigate error accumulation during imagined rollouts. Overall, this category focuses on making the world model more accurate, more stable, and more suitable for supporting policy optimization.

\textbf{The second category consists of selective modeling methods for extracting task-relevant information.} These studies emphasize that the bottleneck of a world model does not always lie in insufficient prediction capability. It may also arise because a large amount of modeling capacity is consumed by visual details irrelevant to control. The key, therefore, is to improve the model's ability to attend to decision-relevant factors. PSP\textsuperscript{\cite{hutson2024policy}} uses policy gradients with respect to input images to identify task-relevant regions, and combines this with segmentation mechanisms to guide the world model to prioritize environmental factors related to the policy. OC-STORM\textsuperscript{\cite{zhang2025objects}} uses object segmentation results obtained from a small amount of manual annotation and a pretrained segmentation network to explicitly inject object-level semantic information into world model learning, thereby shifting the modeling focus from large-scale background reconstruction to key objects and their interactions. In essence, these methods improve the effective capacity utilization of world models in complex visual environments through task-oriented representation selection.

\textbf{The third category focuses on mitigating model bias and improving rollout reliability.} These methods mainly address the usability of model-generated data in policy optimization, namely how to reduce the damage caused by error propagation, distribution shift, and uncontrolled uncertainty during long imagined rollouts. Infoprop\textsuperscript{\cite{frauenknecht2025rollouts}} distinguishes between aleatoric uncertainty inherent in the environment and epistemic uncertainty caused by insufficient model knowledge, suppresses pseudo-random perturbations induced by model errors, and truncates unreliable rollouts when information loss becomes too large. WIMLE\textsuperscript{\cite{aghabozorgi2026wimle}} uses implicit maximum likelihood estimation, model ensembles, and uncertainty-weighted mechanisms to make policy updates rely more on high-confidence model samples. MAC\textsuperscript{\cite{park2025scalable}} targets offline long-horizon tasks and reconstructs dynamics modeling and policy optimization around action chunks, thereby reducing error accumulation while alleviating value bootstrapping bias. Unlike the previous two categories, which focus more on what the model should learn, this category focuses on when model-generated data are reliable and how they can be safely used for decision learning.

\textbf{The fourth category deeply embeds world models into the overall decision-making loop.} These methods no longer treat world models merely as auxiliary predictors, but further use them as core components of perception, planning, control, or policy iteration to support stronger closed-loop decision-making. SINDy-RL\textsuperscript{\cite{zolman2025sindy}} learns interpretable sparse surrogate dynamics and iteratively optimizes policies in a surrogate environment in a Dyna-style manner. DLPA\textsuperscript{\cite{zhang2024model}} targets parameterized action spaces and combines dynamics learning with predictive control, explicitly modeling the coupling between discrete actions and continuous parameters. SAM-RL\textsuperscript{\cite{lv2025sam}} further integrates differentiable physics simulation, differentiable rendering, and active viewpoint selection into model learning and decision-making, enabling the agent to simultaneously refine its environment model and control policy during interaction. VLAW\textsuperscript{\cite{guo2026vlaw}} calibrates the world model using real trajectories, and then uses the improved model to generate high-fidelity synthetic data that feed back into the vision-language-action policy, forming a closed loop in which the model and the policy improve each other alternately. Overall, this line of work reflects the trend that world models are evolving from prediction tools into decision-making infrastructure.

In summary, model-based reinforcement learning incorporates predictions of future state evolution into policy learning by learning environment dynamics models, reward models, or latent-space world models. In recent years, related studies have expanded from early environment dynamics fitting to world model representation learning, task-relevant information extraction, model-error control, and joint optimization of models and policies, forming a relatively rich line of development. \textbf{From the perspective of delay problems}, although these methods are not specifically designed for temporal delays, their forward-looking modeling capability is useful for understanding long-term consequence evaluation and decision-making when the effects of actions appear with delay. However, most existing world models are still essentially data-driven statistical approximations, and they lack explicit characterization of the mechanisms behind state evolution and the transmission of delay effects. When the environment is complex or the delay distribution changes, this limitation may affect model robustness and generalization, which provides a direct motivation for further introducing causal models.

\subsection{Causal Model-Driven Reinforcement Learning}\label{sec:related_work_cmbrl}

In recent years, the development of causal modeling has provided new research perspectives for environment dynamics modeling in reinforcement learning. Causal model-based reinforcement learning (CMBRL) emphasizes explicitly characterizing the causal relationships among actions, intermediate states, and outcomes during environment modeling, rather than merely treating state transitions as statistical mappings to be fitted. Compared with black-box models that only fit state transition probabilities, these methods pay more attention to structural relationships among variables. Therefore, they can not only be used to predict future states, but also help explain the causes of state changes to some extent. Based on these characteristics, existing CMBRL methods can be roughly grouped into three categories according to their technical routes and research focuses: methods that improve world model representation learning and long-horizon modeling through causal constraints; methods that improve mechanism modeling, generalization, and decision-making in model-based reinforcement learning through explicit causal structure learning or utilization; and methods that focus on causal identification in offline settings. It should be noted that these methods are not specifically proposed for delay problems. Nevertheless, \textbf{from the perspective of delay problems}, because they emphasize explicit modeling of functional relationships and effect propagation during state evolution, they still provide useful references for understanding state-change mechanisms in delayed settings. The following paragraphs summarize these related studies accordingly.

\textbf{The first category focuses on improving world model representation learning and long-horizon modeling through causal constraints.} These methods argue that if a world model lacks stable causal-semantic representations, subsequent prediction, planning, and control will all be affected. Causal Dreamer\textsuperscript{\cite{gao2025causal}} is a representative work in this direction. To address the problems of redundant historical observations and difficult long-horizon memory modeling in partially observable environments, this method introduces historical counterfactual reasoning into the Dreamer framework. By removing certain historical observations and comparing their effects on subsequent state transition and reward prediction, it identifies key historical information that truly contributes to dynamics modeling. A gating mechanism is then used to retain useful observations and suppress irrelevant inputs, thereby compressing historical representations and improving multi-step prediction accuracy. It should be noted that although the name of this algorithm is the same as that of the third research point in this dissertation, the two studies were carried out independently in different periods, and their technical ideas and implementations are different. Similarly, WMC\textsuperscript{\cite{poudel2022contrastive}} starts from cross-environment generalization and introduces contrastive unsupervised representation learning based on the invariance principle. By applying style perturbations unrelated to task semantics, it weakens the model's dependence on incidental factors such as texture and color, and combines this with an auxiliary depth prediction task to learn more stable representations, thereby improving out-of-distribution generalization and sim-to-real transfer. LCDM\textsuperscript{\cite{hao2023latent}} further explicitly learns causal structures in latent space. It encodes high-dimensional observations into task-relevant latent factors and makes the update of each latent variable depend only on its causal parents, thereby improving the generalization and interpretability of dynamics modeling. Overall, this category strengthens the adaptability of world models to long-term dependencies, appearance perturbations, and environmental changes by learning compact representations with stronger causal semantics.

\textbf{The second category aims to improve mechanism modeling, systematic generalization, and decision-making in model-based reinforcement learning through explicit causal structure learning or utilization.} Compared with methods that only introduce causal constraints at the representation level, this category places greater emphasis on directly learning or using the causal graph of the environment, and applying it to dynamics modeling, counterfactual rollout, policy optimization, explanation generation, and exploration guidance. Mutti et al.\textsuperscript{\cite{mutti2023provably}} theoretically study systematic generalization by assuming that multiple environments share a sparse and time-invariant causal structure. They estimate a common causal graph and then learn a shared causal transition model on this basis for approximate planning in unknown environments, showing the importance of explicit causal structure for cross-environment generalization. C-MBPO\textsuperscript{\cite{caron2025towards}} incorporates causal modeling into model-based policy optimization. It estimates the local causal structures of state transitions and rewards through conditional independence tests, and performs counterfactual rollouts in a structural causal model to generate pseudo-samples satisfying causal constraints for policy updates, thereby improving robustness under distribution shift. In offline settings, FOCUS\textsuperscript{\cite{zhu2022offline}} learns causal structures of state transitions from offline data and uses them as constraints for world model training, reducing the interference of spurious variables with generalization. BECAUSE\textsuperscript{\cite{lin2024because}} further approaches the problem from causal representation learning. It uses bilinear causal representations and sparse causal masks to characterize structured causal relationships between state-action pairs and next states, and combines behavior policy bias reweighting with pessimistic planning to mitigate the mismatch between world model learning and policy optimization objectives. Beyond prediction and planning, recent studies have also begun to explore the role of causal structure in explanation generation and active exploration. CWM\textsuperscript{\cite{yu2023explainable}} learns sparse causal structures in environment dynamics through causal discovery, and combines them with an inference network to characterize the long-term effects of actions on state evolution and rewards. It can then explain agent decisions based on causal chains and support contrastive explanations between factual and counterfactual actions. ECL\textsuperscript{\cite{cao2025towards}} combines causal structure learning with empowerment maximization. Under the learned causal-structure constraints, it performs empowerment-driven exploration, prioritizes the collection of data that are more controllable and task-relevant, and uses these data to continuously update the causal structure and reward model. Overall, explicit causal structure not only helps alleviate distribution shift, spurious correlation, and objective mismatch, but is also becoming an important foundation for coordinating explanation, exploration, and decision-making in model-based reinforcement learning.

\textbf{The third category focuses on causal identification problems in offline settings caused by unobserved confounding, hidden contexts, and partial observability.} Compared with the preceding methods, this category emphasizes identifiability in the sense of causal inference, rather than only structural generalization. For partially observable Markov decision processes with unobserved confounders, Hong et al.\textsuperscript{\cite{hong2024model}} establish identifiability results for policy values from a model-based perspective. They use reward-observation bridge functions and dynamics-observation bridge functions to characterize the causal effects of actions on immediate rewards and future transitions, and combine two-stage nonparametric estimation with pessimistic policy optimization to obtain finite-sample performance guarantees under weaker coverage conditions. Venkatesh et al.\textsuperscript{\cite{venkatesh2025model}} address unobserved contextual confounding in offline data by transforming contextual Markov decision processes into partially observable decision processes. They combine proximal off-policy evaluation with maximum causal entropy model learning, use observable proxy variables to recover confounded reward expectations, and then construct a plannable proxy MDP. Different from the above methods that emphasize statistical identification, CPRL\textsuperscript{\cite{yu2024causal}} starts from knowledge-guided modeling. It constructs domain causal knowledge as shared causal prompts to guide dynamics model learning, and uses hidden parameters to characterize individual environmental differences, thereby improving robust decision-making and generalization on real-world medical offline data. Overall, this category shows that in offline reinforcement learning, learning an environment model only from surface observations is often insufficient for recovering the true causal mechanism of decision-making. Bridge functions, proxy variables, or domain knowledge are needed to enhance causal identifiability.

In summary, compared with traditional model-based reinforcement learning, causal model-driven reinforcement learning places greater emphasis on explicitly characterizing the structural relationships among actions, states, and outcomes, rather than merely relying on statistical fitting of state transition patterns. Based on this characteristic, these methods seek to further describe the mechanisms behind environment evolution, and use techniques such as causal representation learning and counterfactual reasoning to strengthen the identification and explanation of key influencing factors. They therefore provide a more structured research perspective for state evolution modeling and policy learning in complex environments. \textbf{From the perspective of delay problems}, although these methods are not directly proposed for delay problems, they focus on functional relationships and effect propagation across time scales, and thus remain useful for understanding long-term dependencies, key factor identification, and spurious-correlation interference in delayed settings. However, existing methods remain limited in jointly modeling dynamic time lags, unknown delays, and long-horizon causal dependencies. This also indicates that the application of causal model-driven reinforcement learning to delayed settings still has room for further development.

\section{Formalization}

We consider a partially observable Markov decision process (POMDP) with random delays. At time step $t$, the environment is in a latent state $s_t \in \mathcal{S}$, the agent receives an observation $\mathbf{o}_t \in \mathcal{O}$, executes an action $\mathbf{a}_t \in \mathcal{A}\subseteq \mathbb{R}^{A}$, and obtains a reward $r_t \in \mathbb{R}$ together with a termination signal. Random delay means that the effect of an action on the environment does not necessarily appear immediately at the next time step, but may instead be reflected in later states, observations, and rewards after several time steps. Therefore, the current observation often mixes information about the current environmental state with the delayed effects of several historical actions, and the correspondence between actions and subsequent state changes is no longer direct.

Under this setting, the environmental evolution at time step $t$ can be expressed as Equation~\eqref{eq:5-1-2}:
\begin{equation}
s_{t+1} \sim p(s_{t+1}\mid s_t, \tilde{\mathbf{a}}_t), \qquad
\mathbf{o}_t \sim p(\mathbf{o}_t\mid s_t), \qquad
r_t = r(s_t,\tilde{\mathbf{a}}_t),
\label{eq:5-1-2}
\end{equation}
where $\tilde{\mathbf{a}}_t$ denotes the action that actually takes effect on the environment at time $t$. It may be the current action $\mathbf{a}_t$, or it may be a historical action that takes effect later under the random-delay mechanism.

For proprioceptive tasks in DMC~\textsuperscript{\cite{tunyasuvunakool2020dm_control}}, the observation $\mathbf{o}_t$ usually has a structured form, which can be written as Equation~\eqref{eq:5-2}:
\begin{equation}
\mathbf{o}_t=\{\mathbf{v}_t^{(i)}\}_{i=1}^{N},
\label{eq:5-2}
\end{equation}
where each $\mathbf{v}_t^{(i)} \in \mathbb{R}^{d_i}$ corresponds to a local aspect of the environment, such as position, velocity, angle, or joint state, and $N$ denotes the total number of observation fields. We treat these fields as a set of field nodes and models the relationships among them in latent space. For observations with a dictionary structure, $\mathbf{v}_t^{(i)}$ can directly correspond to the $i$-th observation field. For flat observation vectors, $\mathbf{v}_t^{(i)}$ denotes the $i$-th observation segment obtained according to semantic partitioning.

Based on this setting, this paper aims to learn a world model with an implicit causal inductive bias. The model receives a history of observations and actions, whose input history can be written as Equation~\eqref{eq:5-3}:
\begin{equation}
\mathcal{H}_t=(\mathbf{o}_{1:t},\mathbf{a}_{1:t-1}),
\label{eq:5-3}
\end{equation}
and learns a node-level latent-state representation as shown in Equation~\eqref{eq:5-4-1}:
\begin{equation}
\mathbf{Z}_t=\{\mathbf{z}_t^{(i)}\}_{i=1}^{N},
\label{eq:5-4-1}
\end{equation}
where each $\mathbf{z}_t^{(i)}$ corresponds to the latent representation of one field node. Based on these latent nodes, the world model needs to describe environmental dynamics, reconstruct observations, and predict rewards and termination signals.

Accordingly, the learning objective of the world model can be written as Equation~\eqref{eq:5-5-1}:
\begin{equation}
p(\mathbf{o}_{1:T},r_{1:T},\mathbf{Z}_{1:T}\mid \mathbf{a}_{1:T-1}),
\label{eq:5-5-1}
\end{equation}
where $\mathbf{Z}_{1:T}$ denotes the sequence of latent node states. In practical training, both the world model and the policy are optimized using fixed-length sequence segments, where the batch size is denoted by $B$ and the sequence length by $T$.

In addition to policy learning within a single task, we also consider knowledge reuse among different tasks within the same dynamics family. Let the task set $\mathcal{M}$ be defined as Equation~\eqref{eq:5-6-2}:
\begin{equation}
\mathcal{M}=\{\mathcal{T}_1,\mathcal{T}_2,\dots,\mathcal{T}_K\},
\label{eq:5-6-2}
\end{equation}
where the tasks share the same underlying environmental dynamics, meaning that the state-transition mechanism remains unchanged, while the reward functions or task objectives may differ. The learned world model is expected to preserve structural knowledge that is relatively decoupled from specific task objectives, so that the model can maintain an effective representation of environmental dynamics when switching tasks. This reduces the cost of relearning environmental dynamics and improves adaptation efficiency on new tasks.

Therefore, the problem studied in this paper can be summarized as follows: in a partially observable environment with random delays, how can one use structured observations to learn a world model with an implicit causal inductive bias, so that it can characterize environmental dynamics and observation manifestation under delays, and thereby support sample-efficient policy learning and cross-task transfer?

\section{CausalDreamer Framework and Core Modules}

This section introduces \textbf{CausalDreamer}, a transferable delay-aware reinforcement learning method based on implicit causal graph modeling. The core idea is to represent high-dimensional observations as multiple field nodes with relatively independent semantics, and to characterize environmental state evolution through implicit causal dependencies among nodes, rather than compressing the entire observation into a single latent state. This design preserves relatively stable structural relationships among state factors, learns latent dynamics representations that are closer to the underlying evolution rules, and facilitates world-model reuse when the dynamics mechanism remains the same but the task objective changes. Based on this representation, the agent can track state changes under observation delays and further perform state prediction, imagination learning, and policy planning.

The overall framework of CausalDreamer consists of two tightly coupled modules. \textbf{(1) The implicit-causal-graph-based world model} first maps raw observations into structured representations with node semantics through a field-node encoder. It then models dynamic dependencies among nodes through an implicit causal graph in the causal field-node state-space model, thereby performing latent-state transition modeling and posterior inference. Together with decoding and prediction modules, it reconstructs observations and predicts rewards and termination signals, thereby learning latent representations that characterize environmental dynamics. \textbf{(2) The imagination-based behavior learning and planning module} uses the learned world model to generate imagined trajectories in latent space, supporting subsequent policy optimization and decision planning. The following subsections describe these two modules in detail.

\subsection{Implicit-Causal-Graph-Based World Model}

\begin{figure}
    \centering
    \includegraphics[width=1.0\linewidth]{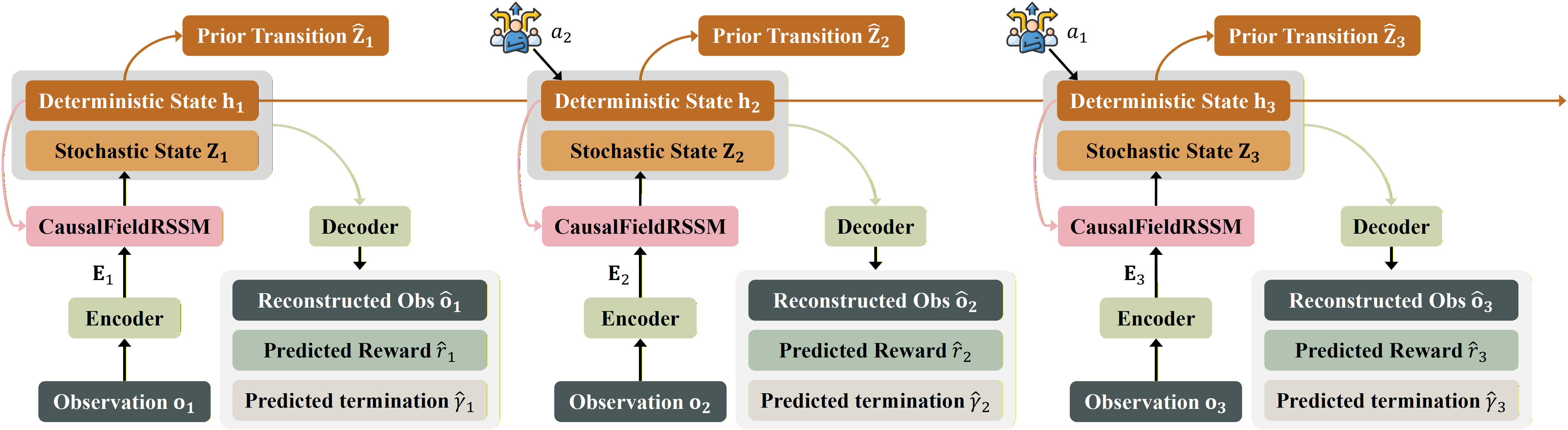}
    \caption{Overall framework of the implicit-causal-graph-based world model}
    \label{fig:worldmodel}
\end{figure}

The implicit-causal-graph-based world model aims to organize raw observations into structured field-node representations and to learn relatively stable latent dynamics through implicit dependencies among nodes, thus providing the basis for policy planning under delays. To this end, the model contains three coordinated components, as shown in Figure~\ref{fig:worldmodel}. First, the \textbf{field-node encoder} maps the input observation into a structured representation composed of multiple field nodes. It preserves semantic boundaries among different state factors at the node level and provides structured inputs for learning reusable latent dynamics. Second, the \textbf{causal field-node state-space model} performs implicit causal modeling of relatively stable dynamic dependencies among field nodes under a joint representation of global deterministic states and field-node stochastic states, and completes latent-state prior transition and posterior inference. This enables the model to reuse underlying state-evolution laws when task objectives or reward functions change. Finally, the \textbf{decoding and prediction modules} reconstruct observations and estimate rewards and termination signals based on the learned latent-state representation. These modules constrain the latent dynamics to retain both environmental-state information and decision-relevant signals, thereby supporting subsequent state prediction, imagination learning, and policy planning.

\subsubsection{Field-Node Encoder}

The field-node encoder $\mathbf{E}_t \sim p_{\varphi}(\mathbf{o}_t)$ converts high-dimensional raw observations into structured node representations for subsequent dynamics modeling. Its key idea is to first determine a set of nodes according to the physical meaning or task semantics of observation variables, and then map each node into a latent representation space with a unified dimension. In this way, different physical or semantic factors in the raw observation can maintain relatively clear boundaries in latent space, providing a basis for subsequent node-level posterior inference and implicit causal graph modeling.

The construction of the node set depends on the form of the observation. For environments with dictionary-structured observations, each observation field, such as velocity or position, can naturally correspond to an independent node. For flat observation vectors without explicit field partitions, the vector is divided into several semantic segments according to physical constraints or task semantics, and each segment is treated as a node. For example, in multi-joint robotic control tasks, the state of each joint can form one node. In more general scenarios, states belonging to the same type of attribute can be grouped into one node, while different categories of attributes correspond to different nodes. After this processing, the raw observation is organized as a node set with explicit physical or semantic meaning.

After determining the node set, the field-node encoder maps the raw features of each node into a unified node-embedding space. For the raw feature $\mathbf{v}_t^{(i)}\in\mathbb{R}^{d_i}$ of the $i$-th node in the current observation $\mathbf{o}_t$, the model uses the corresponding encoding network to obtain the node embedding, as shown in Equation~\eqref{eq:6-node-encoder}:
\begin{equation}
    \mathbf{e}_t^{(i)}=\mathrm{Enc}_i(\mathbf{v}_t^{(i)}),\quad \mathbf{e}_t^{(i)}\in\mathbb{R}^{D_{node}}.
    \label{eq:6-node-encoder}
\end{equation}
Here, $\mathrm{Enc}_i(\cdot)$ denotes the encoding mapping for the $i$-th node, implemented by a multilayer perceptron and a linear projection. Different nodes may have different input dimensions $d_i$, but they are all mapped to the same node-embedding dimension $D_{node}$ after encoding. This design enables the model to process heterogeneous observation fields while maintaining consistency in subsequent node-level computation.

Finally, all node embeddings are stacked along the node dimension to form the field-node embedding representation at the current time step, as shown in Equation~\eqref{eq:6-node-embedding}:
\begin{equation}
    \mathbf{E}_t=[\mathbf{e}_t^{(1)},\dots,\mathbf{e}_t^{(N)}]\in\mathbb{R}^{N\times D_{node}}.
    \label{eq:6-node-embedding}
\end{equation}
This representation organizes the raw observation into a node-based structure, enabling the model to learn potential dependencies among observation variables through node-level message passing in subsequent posterior inference.

\subsubsection{Causal Field-Node State-Space Model}    

The causal field-node state-space model builds on the structured node representation to model the temporal evolution of latent states and to characterize dynamic dependencies among nodes and their propagation across time. This section decomposes the latent state into a global deterministic state and field-node stochastic states, and organizes prior transition and posterior inference around their joint representation. Through this state decomposition, the model can use the global deterministic state to describe the overall dynamic context accumulated from historical information, while using node-level stochastic states to characterize the uncertainty and local dependencies of different state factors at the current time step. This representation facilitates the learning of relatively stable state-evolution laws and provides a clearer dynamics basis for world-model reuse in related tasks.

The causal field-node state-space model decomposes the latent state into two parts. \textbf{(1) The global deterministic state} $\mathbf{h}_t \in \mathbb{R}^{D_d}$ is a deterministic recurrent state based on a recurrent model. It systematically integrates historical observations and agent actions, and is used to characterize deterministic dynamics and potential confounders in the environment. This state represents relatively stable background factors and complex interaction effects in the environment, providing key structured context for subsequent state-transition prediction. \textbf{(2) The field-node stochastic state} $\mathbf{Z}_t \in \mathbb{R}^{N \times D_{node}}$ consists of $N$ node-level random variables, where each node $\mathbf{z}_t^{(i)} \in \mathbb{R}^{D_{node}}$ corresponds to one semantic observation unit. The stochasticity of each node is modeled by a Gaussian distribution to capture the local uncertainty of different semantic units. In random-delay scenarios, this node-based representation can, to some extent, separate observation changes from different sources, thereby improving the robustness of state estimation against delay disturbances.

\textbf{Prior transition $\hat{\mathbf{Z}}_t \sim p_\psi(\hat{\mathbf{Z}}_t \mid \mathbf{h}_{t-1}, \mathbf{Z}_{t-1}, \mathbf{a}_{t-1:t-K})$\quad} generates the prior state representation at the current time step based on the latent state at the previous time step and historical action information, and provides dynamic context for subsequent posterior inference. This process improves the robustness of the model to temporal misalignment between action effects and the currently observable state under random observation delays by aligning historical actions.

First, to mitigate action--state misalignment caused by random action delays, this section introduces an action-history gating mechanism into the state-transition process. Rather than directly assuming that the most recent action $\mathbf{a}_{t-1}$ necessarily corresponds to the current state change, the mechanism adaptively estimates the action component that is more effective for the current transition from the most recent $K$ historical actions $\{\mathbf{a}_{t-1},\dots,\mathbf{a}_{t-K}\}$ according to the previous latent-state context. Specifically, the model uses the global deterministic state $\mathbf{h}_{t-1}$ and the field-node stochastic state $\mathbf{Z}_{t-1}$ to generate delay weights, as shown in Equation~\eqref{eq:5-delay-weight}:
\begin{equation}
\mathbf{w}_t=\mathrm{softmax}\left(\mathrm{MLP}_{delay}\left([\mathbf{h}_{t-1},\mathrm{Pool}(\mathbf{Z}_{t-1})]\right)\right),\quad \mathbf{w}_t\in\mathbb{R}^{K}.\label{eq:5-delay-weight}
\end{equation}
Here, $\mathrm{Pool}(\cdot)$ denotes mean pooling. The weight vector captures the dependency of the current state transition on different historical actions. Based on these weights, the model forms a weighted combination of historical actions to obtain a delay-aligned action representation, as shown in Equation~\eqref{eq:5-delay-action}:
\begin{equation}
    \bar{\mathbf{a}}_{t-1}=\sum_{k=1}^{K}w_t^{(k)}\mathbf{a}_{t-k}.
    \label{eq:5-delay-action}
\end{equation}
where $w_t^{(k)}$ denotes the $k$-th component of the delay-weight vector $\mathbf{w}_t$. This design explicitly enhances the model's robustness to action--state misalignment under random delays. The state transition no longer relies on a fixed-delay assumption, but can dynamically adjust the action source according to the latent-state context.

Next, the model feeds the delay-aligned action representation $\bar{\mathbf{a}}_{t-1}$ and the previous field-node stochastic state $\mathbf{Z}_{t-1}$ into a GRU recurrent network to update the global deterministic state, as shown in Equation~\eqref{eq:5-deter-update}:
\begin{equation}
    \mathbf{h}_t=\mathrm{GRU}\left(\mathrm{MLP}_{trans} \left(\mathrm{Pool}(\mathbf{Z}_{t-1}),\bar{\mathbf{a}}_{t-1}\right),\mathbf{h}_{t-1}\right).
    \label{eq:5-deter-update}
\end{equation}
Here, $\mathrm{MLP}_{trans}$ denotes the transformation that encodes the node-state representation and the action representation. Through this recurrent update, $\mathbf{h}_t$ compresses historical states, node-level stochastic changes, and delay-aware action information into a shared dynamic context, which is used to characterize global factors in environmental evolution.

Finally, in the node-prior prediction stage, the model generates a prior distribution for the stochastic state of each node conditioned on the global deterministic state $\mathbf{h}_t$. To prevent all nodes from sharing only the same global context and thereby losing node-level semantic distinctions, this section further introduces learnable node-ID embeddings so that different nodes can still form differentiated prior parameters under the same global state, as shown in Equation~\eqref{eq:5-node-prior}:
\begin{equation}
    p_\psi(\mathbf{z}_t^{(i)}\mid \mathbf{h}_t)=\mathcal{N}(\boldsymbol{\mu}_t^{(i)},\boldsymbol{\sigma}_t^{(i)}),\ 
    [\boldsymbol{\mu}_t^{(i)},\boldsymbol{\sigma}_t^{(i)}]=\mathrm{MLP}_{prior}([\mathbf{h}_t,\mathbf{e}_i]),\ 
    \mathbf{e}_i=\mathrm{Embed}(i).
    \label{eq:5-node-prior}
\end{equation}

The prior-transition module models action effects under random delays while preserving semantic differences among nodes, thereby providing a dynamically constrained structured prior for subsequent posterior inference.

\begin{figure}
    \centering
    \includegraphics[width=0.65\linewidth]{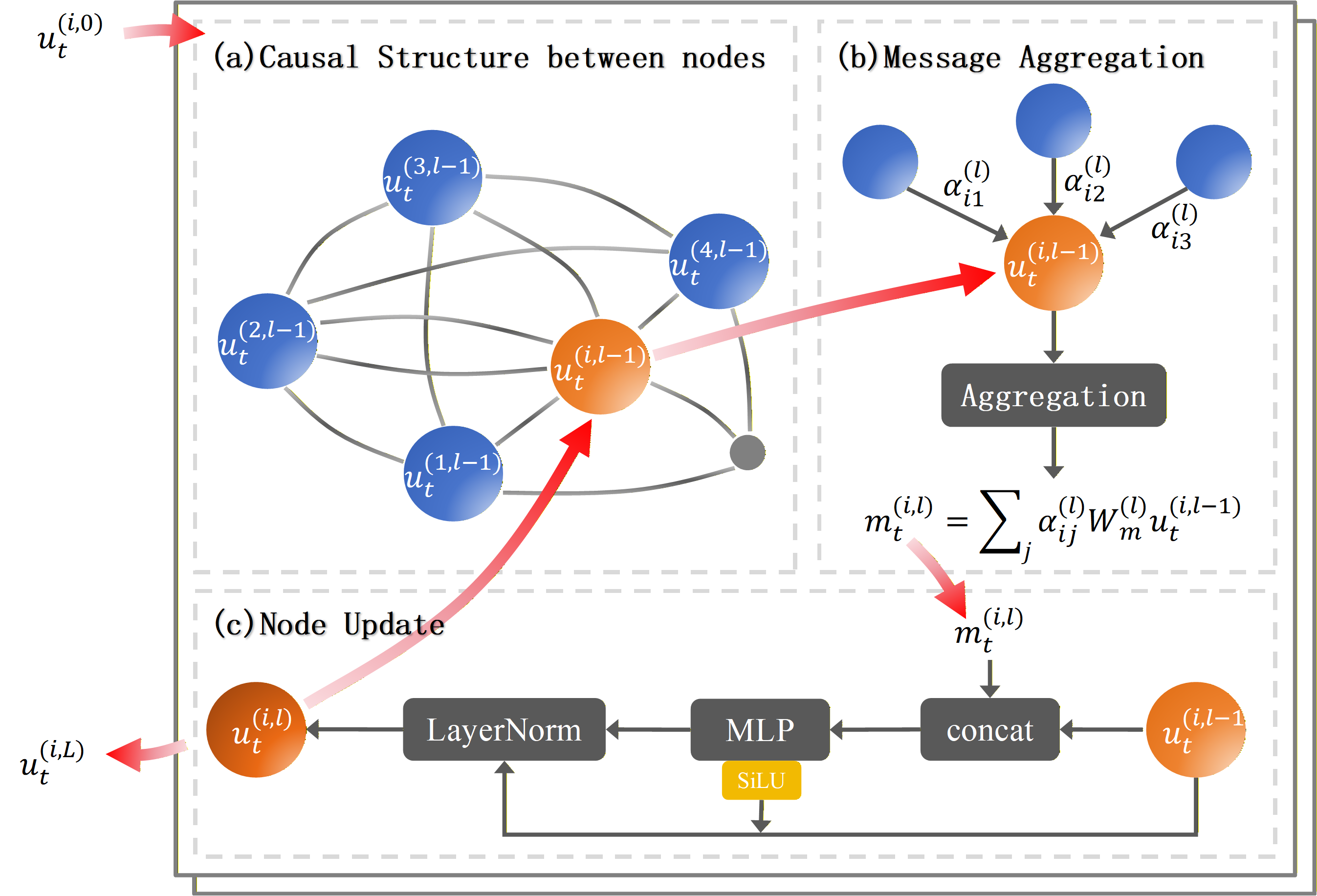}
    \caption{Workflow of the message-passing mechanism}
    \label{fig:message}
\end{figure}

\textbf{Posterior inference $\mathbf{Z}_t \sim q_\phi(\mathbf{Z}_t \mid \mathbf{h}_t, \mathbf{E}_t)$\quad} corrects the stochastic state of each field node given the global deterministic state $\mathbf{h}_t$ and the current observation-node embedding $\mathbf{E}_t$, yielding a state estimate that is closer to the current observation than the prior prediction. Under a node-based state representation, different observation variables or field nodes usually have latent dependencies. If the posterior distribution of each node is estimated independently, the model has difficulty capturing mutual influences among variables. Therefore, this section introduces a node-level message-passing mechanism into posterior inference, modeling information exchange among nodes as an implicit causal graph, as shown in Figure~\ref{fig:message}. This mechanism does not predefine connections among nodes. Instead, it adaptively estimates the influence strength among nodes through learnable message weights, allowing each node to incorporate observation context from other relevant nodes when inferring its own state and thereby better reflect latent structural dependencies.

Specifically, at the $l$-th message-passing layer, node $i$ uses its representation from the previous layer $\mathbf{u}_t^{(i,l-1)}$ as a query and aggregates related contextual information from other node representations $\mathbf{u}_t^{(j,l-1)}(i \neq j)$ to obtain the message vector $\mathbf{m}_t^{(i,l)}$, as shown in Equation~\eqref{eq:5-message-agg}:
\begin{equation}
\mathbf{m}_t^{(i,l)}=\sum_{j=1}^{N}\alpha_{ij}^{(l)}\mathbf{W}_m^{(l)}\mathbf{u}_t^{(j,l-1)}.\label{eq:5-message-agg}
\end{equation}
Here, $\mathbf{W}_m^{(l)}$ is the message transformation matrix at the $l$-th layer, which maps node representation $\mathbf{u}_t^{(j,l-1)}$ into the message space. $\alpha_{ij}^{(l)}$ denotes the influence weight of node $j$ on node $i$. This weight can be interpreted as the edge strength in the implicit causal graph and is used to characterize the relevance of different nodes in posterior inference.

To allow the node representation to retain its own observation features while reflecting structural dependencies among nodes, after obtaining the aggregated message $\mathbf{m}_t^{(i,l)}$, the model feeds it together with the node's own representation $\mathbf{u}_t^{(i,l-1)}$ into a shared update network, and obtains a new node representation through a residual update, as shown in Equation~\eqref{eq:5-node-update}:
\begin{equation}
    \mathbf{u}_t^{(i,l)}=\mathrm{Update}^{(l)}\left(\mathbf{u}_t^{(i,l-1)},\mathbf{m}_t^{(i,l)}\right),\quad l=1,\dots,L.
    \label{eq:5-node-update}
\end{equation}
where $\mathrm{Update}^{(l)}(\cdot)$ denotes the node update network at the $l$-th layer.

Finally, after message passing, the model generates the posterior Gaussian distribution of each node based on the final node representation, as shown in Equation~\eqref{eq:5-posterior}:
\begin{equation}
    q_\phi(\mathbf{z}_t^{(i)}\mid \mathbf{h}_t,\mathbf{E}_t)=\mathcal{N}\left(\tilde{\boldsymbol{\mu}}_t^{(i)},\tilde{\boldsymbol{\sigma}}_t^{(i)}\right),\quad [\tilde{\boldsymbol{\mu}}_t^{(i)},\tilde{\boldsymbol{\sigma}}_t^{(i)}]=\mathrm{MLP}_{out}(\mathbf{u}_t^{(i,L)}).
    \label{eq:5-posterior}
\end{equation}
where $\tilde{\boldsymbol{\mu}}_t^{(i)},\tilde{\boldsymbol{\sigma}}_t^{(i)}\in\mathbb{R}^{D_{node}}$ denote the mean and standard deviation of the posterior distribution of the $i$-th node, respectively.

The posterior-inference module uses the current observation to correct the prior state obtained from historical information, and integrates information from relevant nodes on the implicit causal graph through multi-layer message passing. As a result, the stochastic-state estimate of each node no longer depends only on its own observation, but is also constrained by the global dynamic context and the contexts of other nodes. This mechanism enables the model to preserve stable dependencies among state factors at the node level, thereby learning latent dynamics representations that are closer to the underlying state-evolution laws. For scenarios where the dynamics mechanism is the same but the task objective differs, such representations help reduce interference from task-specific observations or reward changes and improve the reusability of the world model in cross-task transfer.

\subsubsection{Decoding and Prediction Modules}

The decoding and prediction modules transform the latent-state representation described above into prediction results directly related to observation recovery and decision evaluation. After the field-node encoder and the causal field-node state-space model, the world model has obtained latent-state representations that characterize environmental dynamics. Based on these representations, it is further necessary to examine whether they can support raw observation reconstruction and reward and termination prediction.

To this end, this section flattens $\mathbf{Z}_t$ into $\mathbf{z}_t \in \mathbb{R}^{N \cdot D_{node}}$ and concatenates it with $\mathbf{h}_t \in \mathbb{R}^{D_d}$ to obtain the complete latent feature $\mathbf{f}_t=[\mathbf{z}_t,\mathbf{h}_t]\in\mathbb{R}^{N\cdot D_{node}+D_d}$. This feature contains both node-level stochastic states and the global deterministic state, and serves as the unified input to subsequent prediction heads for raw observation reconstruction, reward prediction, and termination prediction. Specifically, the \textbf{observation decoder} $\hat{\mathbf{o}}_t \sim p_\theta(\hat{\mathbf{o}}_t \mid \mathbf{f}_t)$ recovers observation signals from latent features and constrains the learned representation to retain key information related to the environmental state. The \textbf{reward predictor} $\hat{r}_t \sim p_\xi(\hat{r}_t \mid \mathbf{f}_t)$ estimates the immediate reward under the current latent state. The \textbf{termination predictor} $\hat{\gamma}_t \sim p_\alpha(\hat{\gamma}_t \mid \mathbf{f}_t)$ determines whether the current state is close to termination. Together, these modules constrain the latent representation so that the world model can characterize environmental state evolution and provide decision-relevant prediction signals for subsequent imagination learning and behavior planning.

\subsubsection{Training Objective}

This section constrains the representation quality, prediction capability, and temporal consistency of the world model from the optimization perspective. After obtaining observation reconstruction, reward prediction, and termination prediction results, model training needs not only to ensure sufficient output accuracy, but also to maintain the compactness and generalizability of latent-state representations through regularization between the prior and posterior distributions. In addition, to address the unstable correspondence between action effects and currently observable states caused by random observation delays, this section introduces an additional delay-aware regularizer to enhance temporal discrimination under random delays.

Based on this design, the training objective of the world model consists of two parts: the standard reconstruction and regularization loss for the world model, and the entropy regularization term introduced for the delay-alignment mechanism. The overall loss function is shown in Equation~\eqref{eq:5-14}:
\begin{equation}
    \mathcal{L} = \mathcal{L}_{WM} + \mathcal{L}_{delay},
    \label{eq:5-14}
\end{equation}
where the world-model loss $\mathcal{L}_{WM}$ consists of the observation reconstruction term, reward prediction term, termination prediction term, and the KL regularization term between the prior and posterior distributions, as shown in Equation~\eqref{eq:5-15}:
\begin{equation}
\begin{aligned}
    \mathcal{L}_{WM}
    =
    \mathbb{E}_{ \mathcal{D}}
    \Bigg[
    \sum_{t=1}^{T}
    \Big(
        &-\log p_\theta(\mathbf{o}_t \mid \mathbf{f}_t)
        -\log p_\xi(r_t \mid \mathbf{f}_t)
        -\log p_\gamma(\gamma_t \mid \mathbf{f}_t) \\
        &+ \beta \, \mathrm{KL}\big(q_\phi(\mathbf{Z}_t \mid \mathbf{h}_t, \mathbf{E}_t)\,\|\,p_\psi(\mathbf{Z}_t \mid \mathbf{h}_t)\big)
    \Big)
    \Bigg],
\end{aligned}
\label{eq:5-15}
\end{equation}
where $\mathcal{D}$ denotes the dataset that stores transition data, and $\beta$ denotes the coefficient of the KL regularization term. The first three terms in Equation~\eqref{eq:5-15} constrain the quality of observation reconstruction, reward prediction, and termination prediction, respectively. The KL term regularizes the prior and posterior distributions and maintains the compactness and generalizability of the latent-state representation.

For random-delay scenarios, this section further introduces an entropy penalty on delay weights to improve the discriminability of the delay-alignment mechanism when selecting historical actions, as shown in Equation~\eqref{eq:5-16}:
\begin{equation}
    \mathcal{L}_{delay}
    =
    \lambda_{ent}
    \sum_{t=1}^{T}
    \left(
        -\sum_{k=1}^{K} w_t^{(k)} \log\big(w_t^{(k)} + \epsilon\big)
    \right),
    \label{eq:5-16}
\end{equation}
where \(\lambda_{ent}\) is the weight coefficient of the entropy regularization term, \(\mathbf{w}_t \in \mathbb{R}^K\) is the historical-action gating weight vector generated by the delay-alignment mechanism at time step \(t\), and its $k$-th component is denoted by \(w_t^{(k)}\). \(\epsilon > 0\) is a small constant introduced to avoid numerical instability. This regularizer encourages the model to focus on a small number of critical historical time steps, making the delay selection sharper and thereby enhancing the stability of dynamics modeling under random delays.

Through the above joint optimization objective, the causal field-node state-space model can learn observation reconstruction, reward prediction, termination prediction, and latent-state transition within a unified framework. Specifically, the global deterministic state $\mathbf{h}_t$ integrates historical actions and observations, the structured message passing among nodes captures dependencies among variables, and the field-node stochastic states preserve local uncertainty of each semantic unit. In addition, the delay-alignment mechanism and entropy regularization allow the model to maintain relatively stable temporal-structure modeling ability even under random observation delays.

\refstepcounter{algorithm}

\textbf{Algorithm workflow\quad} The training process of the world model is shown in Algorithm~\ref{alg:world_model}. First, sequence segments of length $L$ are sampled from the dataset $\mathcal{D}$. For each sequence, the following operations are performed at each time step: the field-node encoder maps the raw observation into node embeddings; the causal field-node state-space model performs delay alignment under historical action inputs, and updates the global deterministic state and field-node stochastic states; then the joint feature is fed into the decoder, reward predictor, and termination predictor to obtain observation reconstruction, reward prediction, and termination prediction, respectively. After the forward pass, the observation reconstruction loss, reward prediction loss, termination prediction loss, KL regularization between the prior and posterior, and entropy regularization for the delay-gating weights are computed. The weighted sum of these losses is then backpropagated to update the model parameters. This process is repeated until the maximum number of training steps is reached, yielding the trained world model.

\begin{algorithm}
\caption{Training Algorithm of the CausalDreamer World Model}
\label{alg:world_model}
\begin{algorithmic}[1]
    \Require Dataset $\mathcal{D}$, batch size $B$, sequence length $L$, historical action window length $K$, KL weight $\beta$, delay regularization weight $\lambda_{ent}$, learning rate $\eta$
    \Ensure Trained world-model parameters $\Phi=\{\varphi,\psi,\phi,\theta,\xi,\alpha\}$, where $\varphi$, $\psi$, $\phi$, $\theta$, $\xi$, and $\alpha$ denote the parameters of the field-node encoder, prior-transition model, posterior-inference model, decoder, reward predictor, and termination predictor, respectively

    \State Initialize world-model parameters $\Phi$
    \While{the maximum number of training steps is not reached}
        \State Sample a batch of sequences $\{(\mathbf{o}_{1:L},\mathbf{a}_{1:L},r_{1:L},\gamma_{1:L})\}_{b=1}^{B}$ from $\mathcal{D}$
        \State Initialize latent states $\mathbf{h}_0,\mathbf{Z}_0$, and set the total loss $\mathcal{L}\gets0$

        \For{$t=1$ \textbf{to} $L$}
            \State Encode the observation into field-node embeddings: $\mathbf{E}_t\gets p_{\varphi}(\mathbf{o}_t)$
            \State Perform delay-aware prior transition: $\hat{\mathbf{Z}}_t\gets p_{\psi}(\mathbf{h}_{t-1},\mathbf{Z}_{t-1},\mathbf{a}_{t-1:t-K})$
            \State Perform message-passing posterior inference: $\mathbf{Z}_t\gets q_{\phi}(\mathbf{h}_t,\mathbf{E}_t)$
            \State Construct the latent feature: $\mathbf{f}_t\gets[\mathrm{Flatten}(\mathbf{Z}_t),\mathbf{h}_t]$
            \State Reconstruct the observation: $\hat{\mathbf{o}}_t\gets p_{\theta}(\mathbf{f}_t)$
            \State Predict the reward: $\hat{r}_t\gets p_{\xi}(\mathbf{f}_t)$
            \State Predict termination: $\hat{\gamma}_t\gets p_{\alpha}(\mathbf{f}_t)$
            \State Accumulate the training loss: $\mathcal{L}\gets\mathcal{L}+\mathcal{L}_{WM}+\mathcal{L}_{delay}$
        \EndFor

        \State Update world-model parameters: $\Phi\gets\Phi-\eta\nabla_{\Phi}\mathcal{L}$
    \EndWhile
\end{algorithmic}
\end{algorithm}

\subsection{Imagination-Based Behavior Learning and Planning}

Imagination-based behavior learning and planning performs policy optimization and value estimation on top of the learned world model, with the main goal of reducing dependence on real environment interactions. Unlike direct trial and error in the environment, this method uses the world model to roll out future trajectories in latent space, and updates the policy network and value network based on these imagined trajectories. This section follows the imagination-learning paradigm of DreamerV3\textsuperscript{\cite{hafner2023mastering}} and organizes behavior learning into three stages: trajectory imagination, value learning, and policy optimization.

The first stage is trajectory imagination. This stage generates latent trajectories for policy learning without accessing the real environment. Starting from the latent state $\mathbf{f}_t$ corresponding to a real interaction trajectory, the policy network $\pi_{\Theta_\pi}$ samples an action $\mathbf{a}_t$, and the prior transition of the causal field-node state-space model predicts the next latent state $\mathbf{f}_{t+1}$. Repeating this process rolls out an imagined trajectory of length $H$ in latent space, as shown in Equation~\eqref{eq:5-16-2}:
\begin{equation}
    \tau = (\mathbf{f}_t,\mathbf{a}_t,\hat{r}_t,\hat{\gamma}_t,\mathbf{f}_{t+1},\ldots,\mathbf{f}_{t+H}).
    \label{eq:5-16-2}
\end{equation}
Here, $\hat{r}_t$ and $\hat{\gamma}_t$ are given by the reward predictor and termination predictor, respectively. Since imagined trajectories are generated entirely in latent space, the policy network can obtain more training signals with fewer real interactions.

The second stage is value learning. The value network $V_{\Theta_V}(\mathbf{f}_t)$ estimates the long-term return of the current latent state under policy $\pi_{\Theta_\pi}$ and provides evaluation signals for subsequent policy updates. This section uses the $\lambda$-return as the training target, and updates the value network by minimizing the discrepancy between value prediction and the target return, as shown in Equation~\eqref{eq:5-17}:
\begin{equation}
    \mathcal{L}_V=\mathbb{E}_{\tau}\left[\sum_t \frac{1}{2}\left(V_{\Theta_V}(\mathbf{f}_t)-R_t^\lambda\right)^2\right].
    \label{eq:5-17}
\end{equation}
where $R_t^\lambda$ denotes the $\lambda$-return target constructed from predicted rewards, discount factors, and value estimates along the imagined trajectory.

The final stage is policy optimization. The objective of the policy network is to select actions that yield higher long-term returns on imagined trajectories generated by the world model. This section uses dynamics-path gradients\textsuperscript{\cite{hafner2019dream}} to update the policy network. That is, the policy loss depends not only on the return targets along the imagined trajectory, but also backpropagates gradients to the policy parameters through the differentiable transition process of the world model. Specifically, the $\lambda$-return $R_t^\lambda$ is first computed from the imagined trajectory, and the prediction of the value network is used as a baseline to obtain the advantage estimate, as shown in Equation~\eqref{eq:5-actor-adv}:
\begin{equation}
    A_t = R_t^\lambda - V_{\Theta_V}(\mathbf{f}_t).
    \label{eq:5-actor-adv}
\end{equation}
Based on this, the policy network is updated by maximizing the advantage objective, together with a policy-entropy regularizer to maintain necessary exploration, as shown in Equation~\eqref{eq:5-18}:
\begin{equation}
    \mathcal{L}_\pi=-\mathbb{E}_{\tau}\left[\sum_t A_t\right]-\kappa \mathcal{H}\left(\pi_{\Theta_\pi}(\cdot\mid\mathbf{f}_t)\right).
    \label{eq:5-18}
\end{equation}
where $\mathcal{H}(\cdot)$ denotes the entropy of the policy distribution, and $\kappa$ is the entropy regularization coefficient. The policy parameters $\Theta_\pi$ are updated by minimizing $\mathcal{L}_\pi$, as shown in Equation~\eqref{eq:5-actor-update}:
\begin{equation}
    \Theta_\pi \leftarrow \Theta_\pi - \eta_\pi \nabla_{\Theta_\pi}\mathcal{L}_\pi.
    \label{eq:5-actor-update}
\end{equation}
Since imagined trajectories are generated by a differentiable prior-transition model, $\nabla_{\Theta_\pi}\mathcal{L}_\pi$ can be backpropagated along the paths through which actions affect latent states, reward predictions, and value estimates, thereby realizing dynamics-path-gradient-based policy optimization.

Through the above three stages, the behavior-learning module completes a closed-loop optimization process from trajectory generation, return estimation, and policy update within the learned world model, thereby enabling model-based policy learning.

\section{Experimental Setup and Results Analysis}

\subsection{Experimental Setup}

\subsubsection{Experimental Environment}

We evaluate CausalDreamer and the baseline methods on continuous-control benchmark tasks from the DeepMind Control Suite (DMC)\textsuperscript{\cite{tassa2018deepmind}}. Built on the MuJoCo physics engine, DMC provides standardized continuous-control tasks such as walker-walk and cheetah-run. Because it offers high-fidelity physical simulation and structured proprioceptive observations composed of low-dimensional vectors such as joint angles and angular velocities, it is well suited for evaluating the modeling capability of structured world models.

To better reflect delay effects that commonly exist in real-world environments, we introduce a random observation-delay wrapper at the environment level to examine the robustness of the proposed method under non-ideal interaction conditions. Let the maximum observation delay be $\tau_{max}=5$. Then the observation received by the agent at time $t$ is not the immediate observation corresponding to the current environmental state, but a delayed observation from the historical observation queue:
\begin{equation}
    \delta_t \sim \mathcal{U}\{1,\dots,\tau_{max}\}, \qquad o_t^{obs} = o_{t-\delta_t}.
\end{equation}
In other words, the agent must make decisions based on the delayed observation $o_t^{obs}$, while the environmental state continues to evolve under the current action. This creates a random time lag between the input observation and the true environmental state, making the setting closer to delayed perception in real systems.

In addition, we use structured observations rather than pixel observations for training and testing, so as to focus on the use of proprioceptive information and the implicit modeling of causal relations.

\subsubsection{Evaluation Metric}

We use average return as the evaluation metric, namely the average cumulative reward obtained by the agent within a finite episode. All selected DMC tasks have an episode length of 1000 steps and a maximum episode reward of 1000. Unless otherwise specified, all experimental results are averaged over 5 random seeds. A higher average return indicates better task performance.

\subsubsection{Baseline Methods}

The selected baselines include the model-based reinforcement learning method DreamerV3\textsuperscript{\cite{hafner2023mastering}}, the causal-model-based reinforcement learning method CWM\textsuperscript{\cite{yu2023explainable}}, and the model-free reinforcement learning method SAC\textsuperscript{\cite{haarnoja2018soft}}. These baselines are described as follows:
\begin{itemize}
    \item \textbf{DreamerV3}\textsuperscript{\cite{hafner2023mastering}} is one of the current state-of-the-art model-based reinforcement learning methods. Its recurrent state-space model jointly models stochastic and deterministic features of the environment, has strong world-model representation ability, and performs well on a range of tasks.
    \item \textbf{CWM}\textsuperscript{\cite{yu2023explainable}} is a reinforcement learning method based on explicit causal modeling. It identifies causal relations in the environment using conditional-independence tests, and quantifies the influence strength of different causes on effects through an attention mechanism. The learned causal world model is embedded into a model-based reinforcement learning framework, providing both interpretability and policy-learning capability.
    \item \textbf{SAC}\textsuperscript{\cite{haarnoja2018soft}} is a leading model-free reinforcement learning method based on the maximum-entropy reinforcement learning framework. It has strong sample efficiency and policy stability in continuous action spaces, and is included to compare model-based and model-free reinforcement learning methods in terms of sample efficiency, final performance, and training stability.
\end{itemize}

\subsection{Experimental Results and Analysis}

To verify the effectiveness of CausalDreamer, this section systematically compares and analyzes CausalDreamer and the baselines from the following five aspects.

\subsubsection{Performance of Different Methods in Environments without Delays}

\begin{figure}
    \centering
    \setlength{\tabcolsep}{0pt}
    \includegraphics[width=0.65\textwidth]{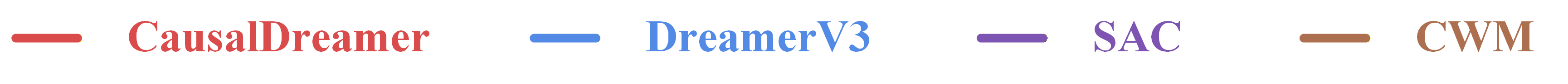}
    
    \begin{tabular}{cccc}
        \subfloat[\textit{Cheetah Run}]{
            \centering
            \includegraphics[width=0.24\textwidth]{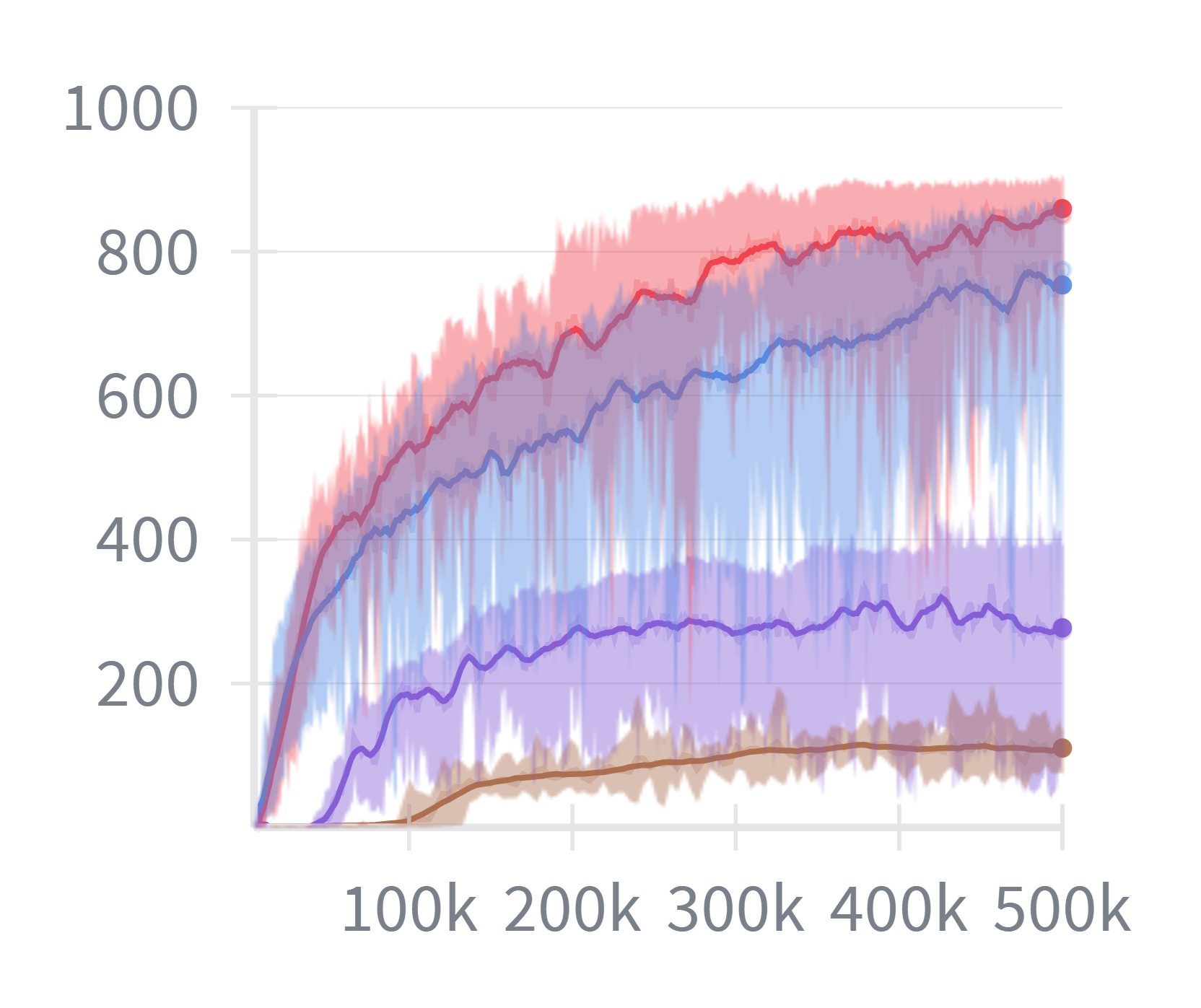}} &
        \subfloat[\textit{Walker Run}]{
            \centering
            \includegraphics[width=0.24\textwidth]{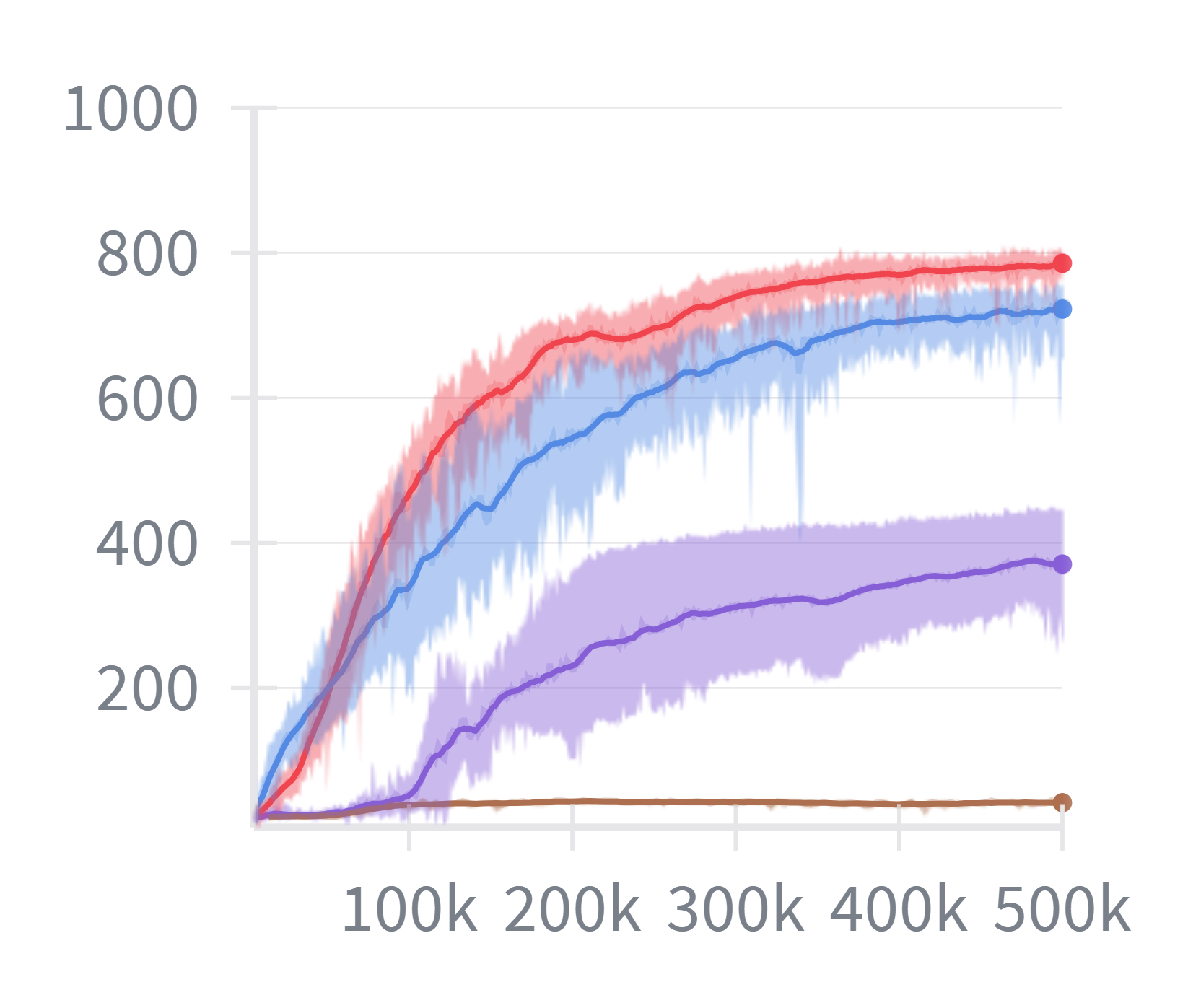}} &
        \subfloat[\textit{Walker Stand}]{
            \centering
            \includegraphics[width=0.24\textwidth]{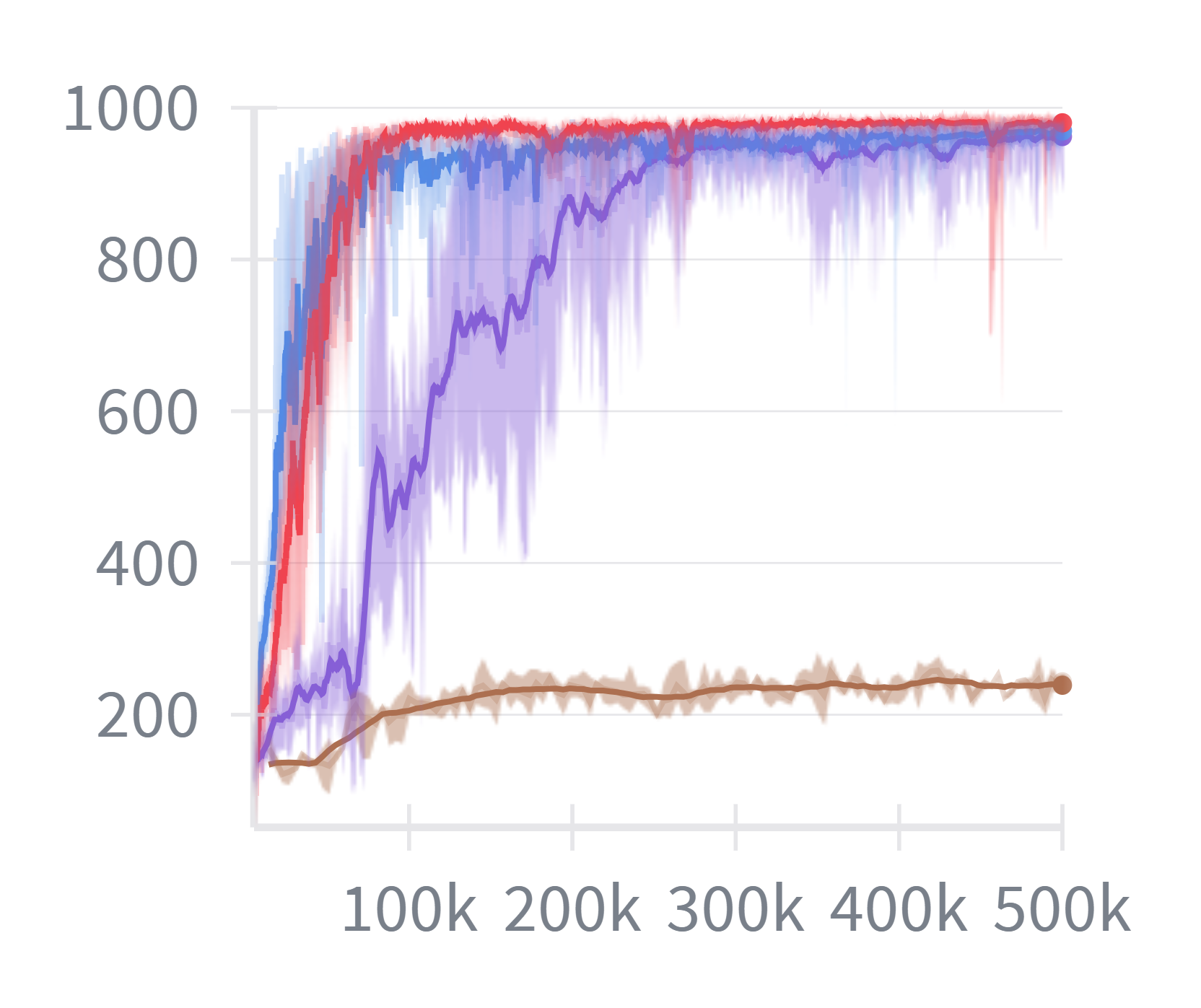}} &
        \subfloat[\textit{Walker Walk}]{
            \centering
            \includegraphics[width=0.24\textwidth]{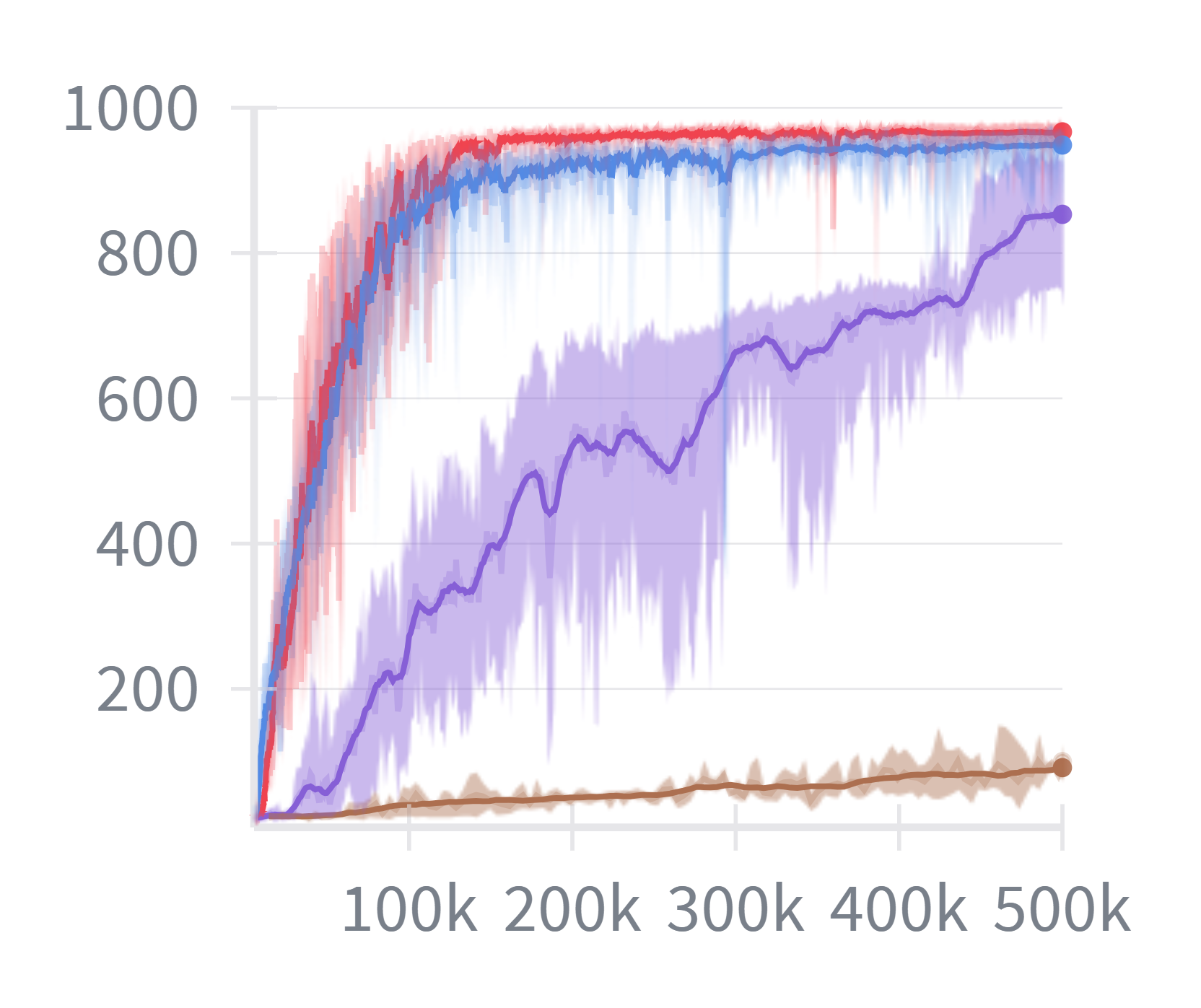}} \\
        
        \subfloat[\textit{Reacher Easy}]{
            \centering
            \includegraphics[width=0.24\textwidth]{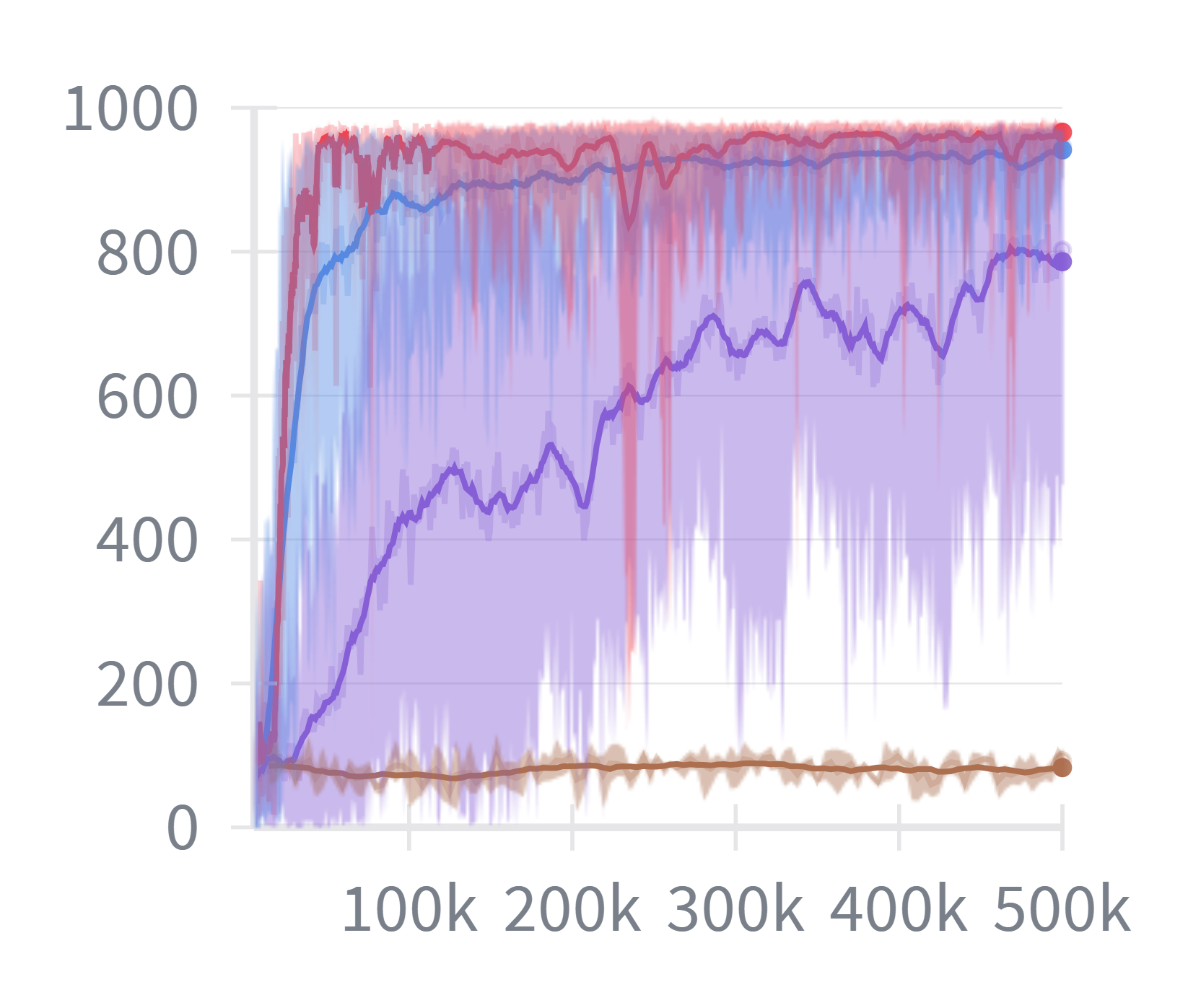}} &
        \subfloat[\textit{Reacher Hard}]{
            \centering
            \includegraphics[width=0.24\textwidth]{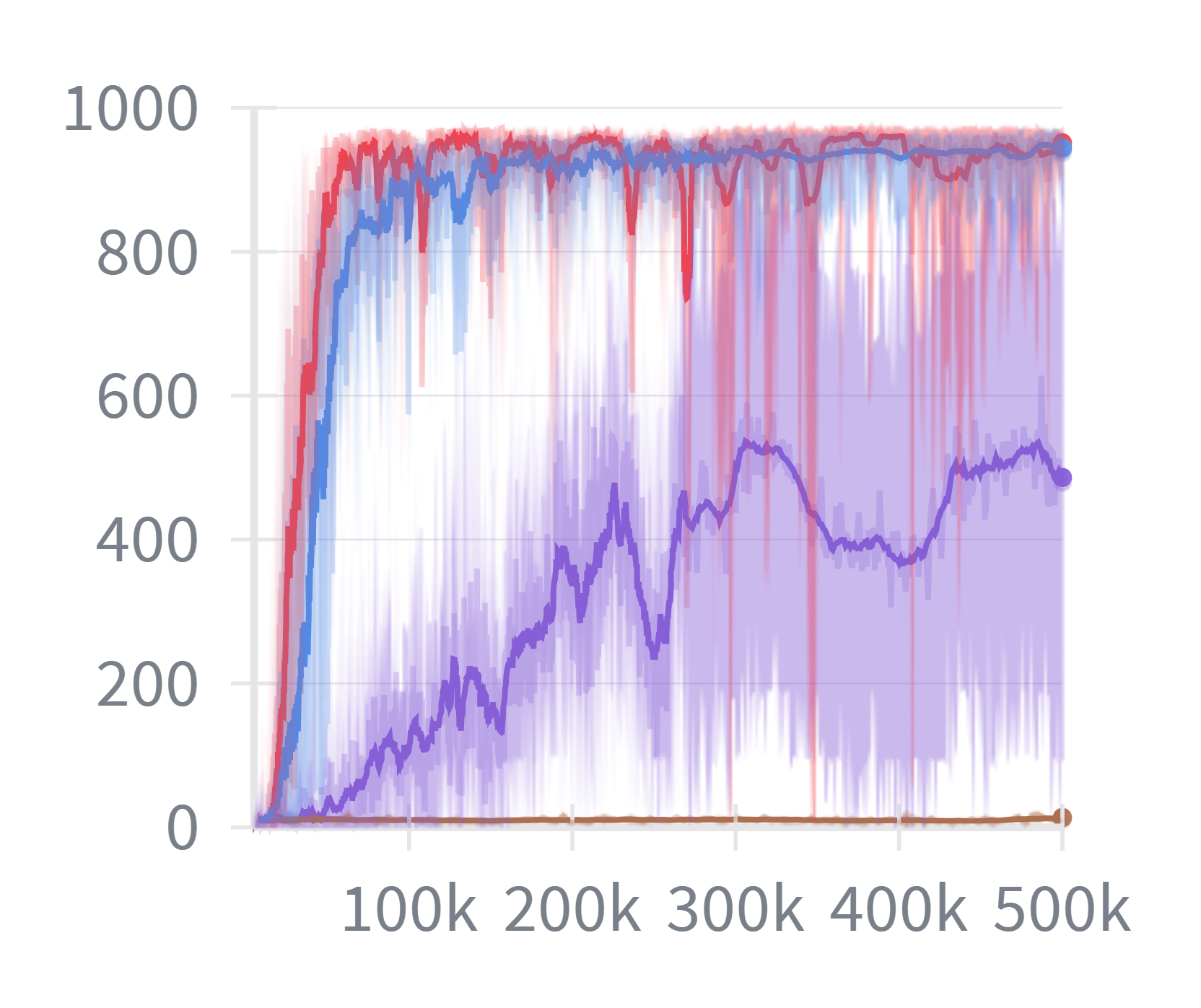}} &
        \subfloat[\textit{Hopper Hop}]{
            \centering
            \includegraphics[width=0.24\textwidth]{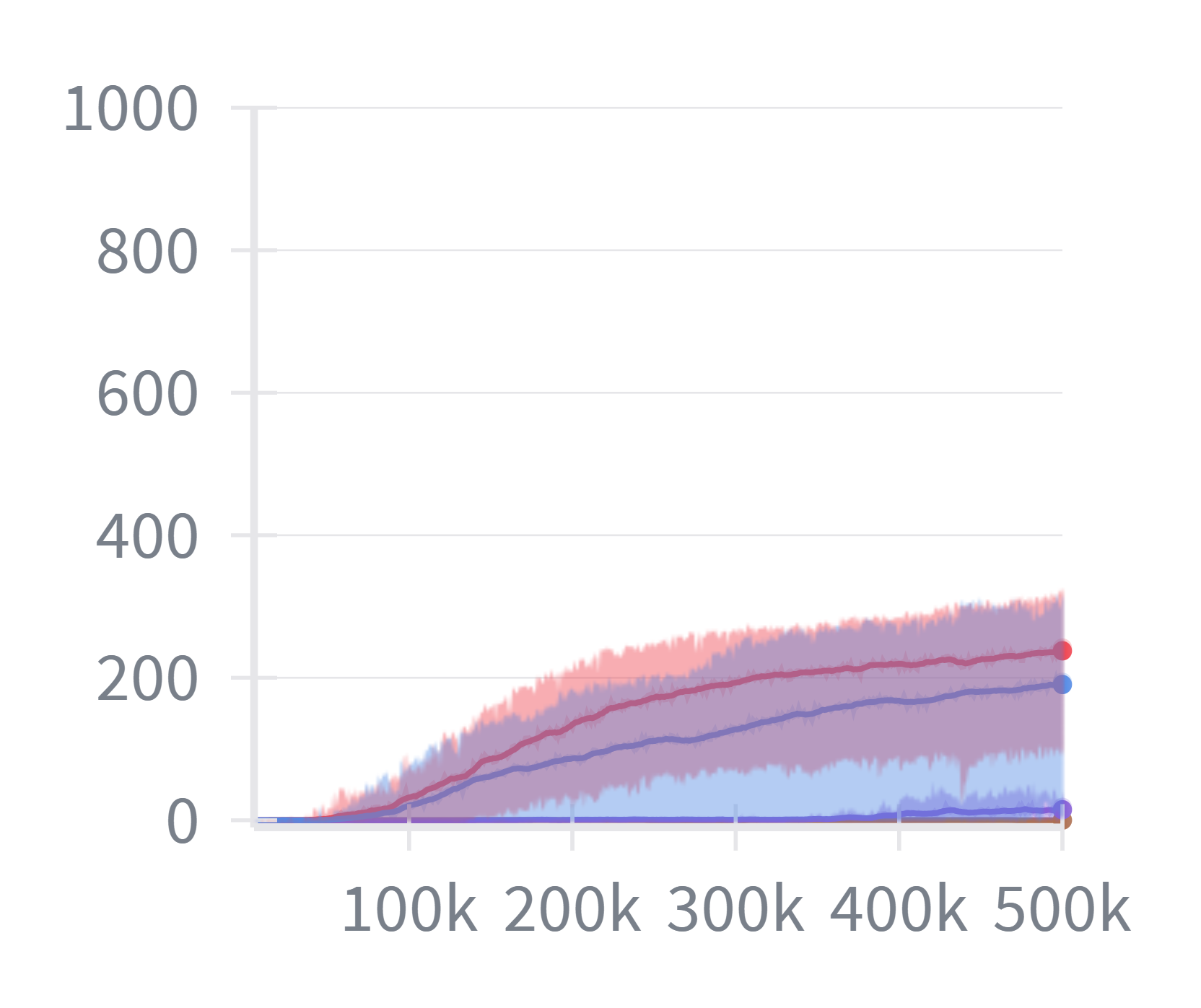}} &
        \subfloat[\textit{Cartpole Swingup}]{
            \centering
            \includegraphics[width=0.24\textwidth]{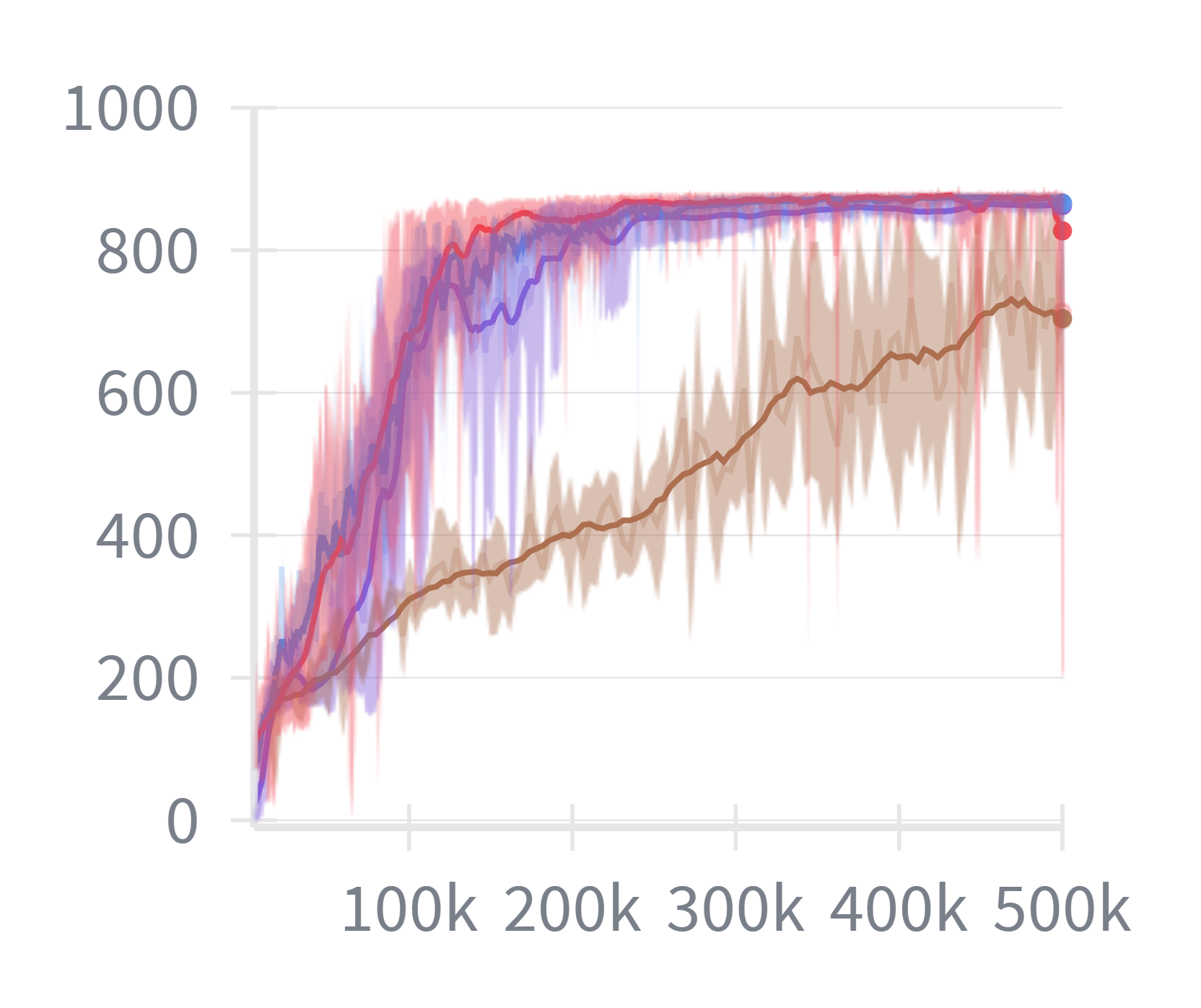}} \\
    \end{tabular}
    \caption{Performance of different methods on DMC tasks without delays}
    \label{fig:ave_success_ratio_delay0}
\end{figure}

\textbf{Experimental design\quad}
To examine the basic learning capability of different methods in standard no-delay environments and compare their sample efficiency and training stability, this section conducts comparative experiments on 8 no-delay continuous-control tasks from DMC. The results are shown in Figure~\ref{fig:ave_success_ratio_delay0}, where the horizontal axis denotes the number of real episodes used for training, and the vertical axis denotes average return.

\textbf{Experimental results\quad}
As shown in Figure~\ref{fig:ave_success_ratio_delay0}, \textbf{CausalDreamer} achieves the strongest overall performance in no-delay tasks. In most tasks, it increases the return rapidly and maintains relatively stable training curves. Its final performance is also slightly better than that of the other methods in most tasks. This indicates that the implicit causal graph modeling adopted by CausalDreamer not only avoids the complexity and instability that may arise from explicit causal discovery, but also more naturally captures structural dependencies in the environment inside the world model. As a result, the method remains lightweight while exhibiting strong sample efficiency and stable learning ability.

\textbf{DreamerV3} is also competitive and can converge quickly to high returns in most tasks, but its overall performance is still slightly weaker than that of CausalDreamer. This is mainly related to its latent-state representation. DreamerV3 models environmental dynamics through deterministic and stochastic states and has strong basic modeling capability. However, it usually encodes the observation into a single holistic latent state, lacking field-node decomposition of different state factors and further modeling of implicit causal dependencies among nodes. Therefore, in tasks where observation variables are strongly coupled or where different state factors affect dynamic evolution differently, its latent representation is less structured. In contrast, CausalDreamer preserves structured state relationships through the field-node encoder and implicit causal graph modeling, thereby further improving training stability and final performance in some tasks.

\textbf{SAC}, as a model-free method, can also reach high returns in some tasks, but its overall convergence speed is significantly slower than the two model-based methods, and its training curves fluctuate more. As shown in the figure, its curves improve slowly in the early stage on many tasks, and also show noticeable oscillations in the middle and late stages. This is consistent with the characteristics of SAC: because it does not rely on an explicit world model, its policy optimization is mainly based on a large number of real environment interactions, which puts it at a relative disadvantage in sample efficiency.

\textbf{CWM} has the weakest overall performance. In most tasks, both its return level and training stability are clearly worse than those of CausalDreamer and DreamerV3, and it performs reasonably only on a few tasks with relatively simple dynamics. The results suggest that although the explicit causal modeling adopted by CWM has desirable theoretical interpretability, the stability of causal discovery and the accumulation of model errors in continuous-control scenarios can substantially affect final control performance, limiting its practical performance in complex dynamic environments.

\subsubsection{Performance of Different Methods in Environments with Random Delays}

\begin{figure}
    \centering
    \setlength{\tabcolsep}{0pt}
    \includegraphics[width=0.65\textwidth]{general_legand_chapter5.png}
    
    \begin{tabular}{cccc}
        \subfloat[\textit{Cheetah Run}]{
            \centering
            \includegraphics[width=0.24\textwidth]{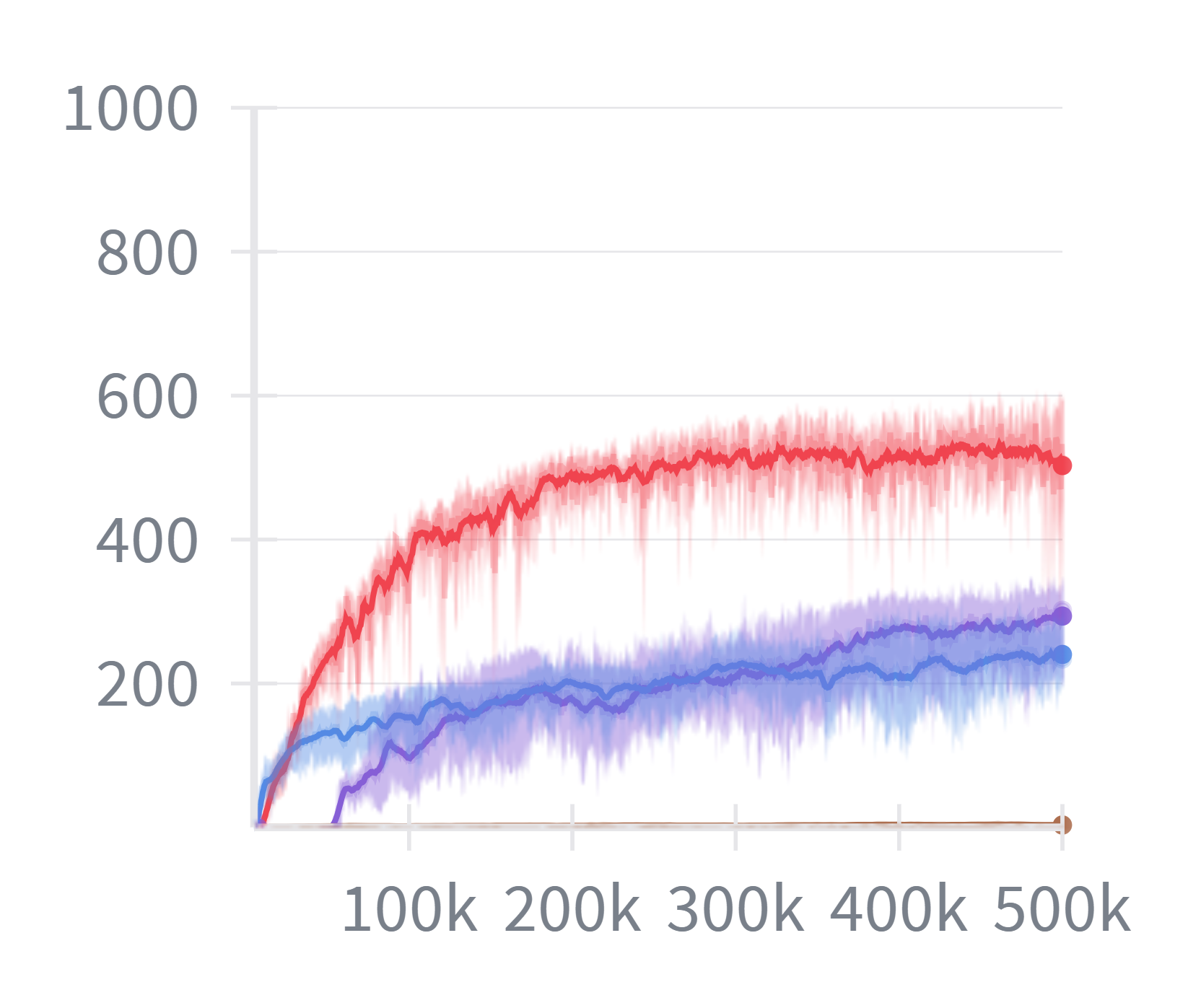}} &
        \subfloat[\textit{Walker Run}]{
            \centering
            \includegraphics[width=0.24\textwidth]{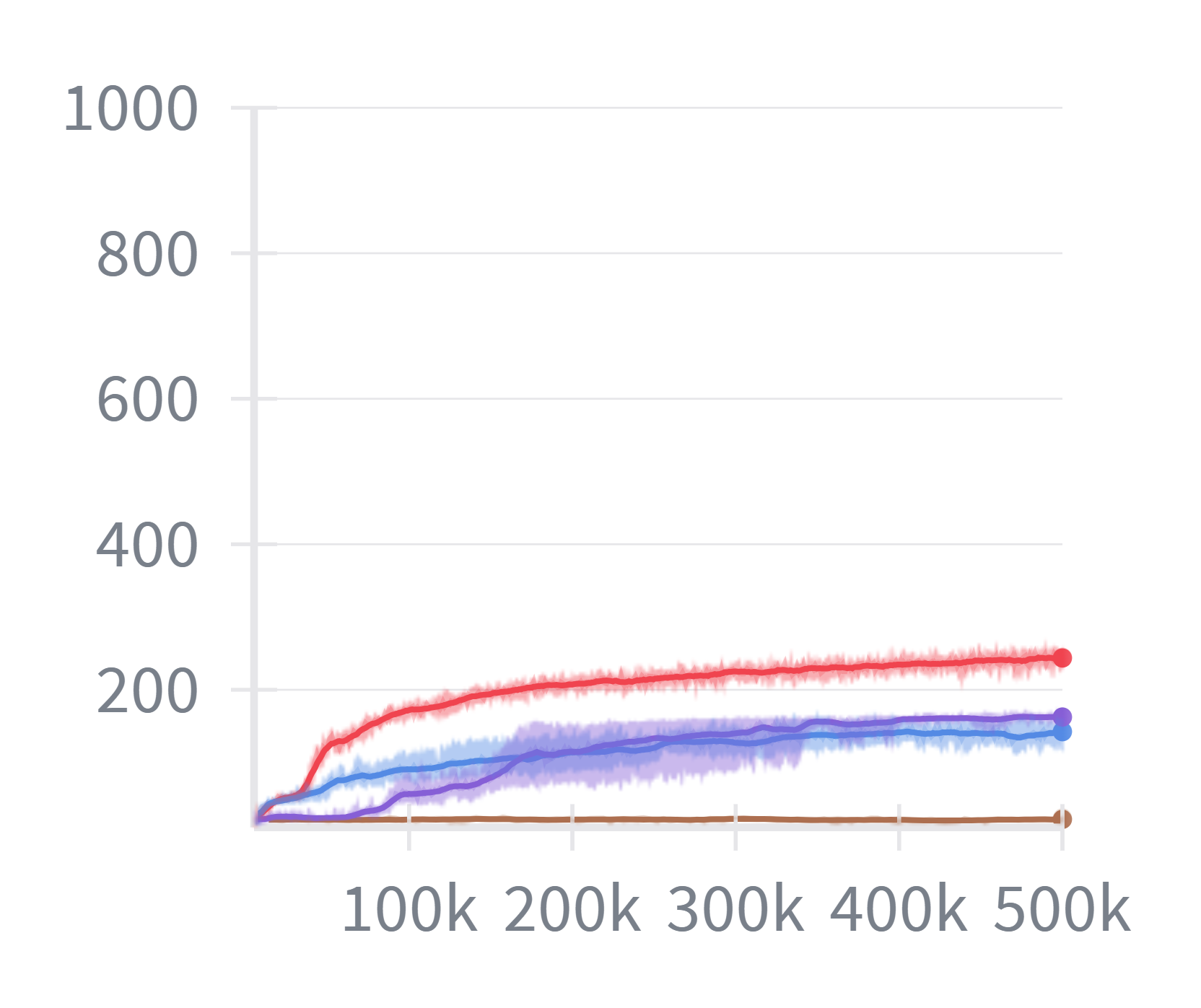}} &
        \subfloat[\textit{Walker Stand}]{
            \centering
            \includegraphics[width=0.24\textwidth]{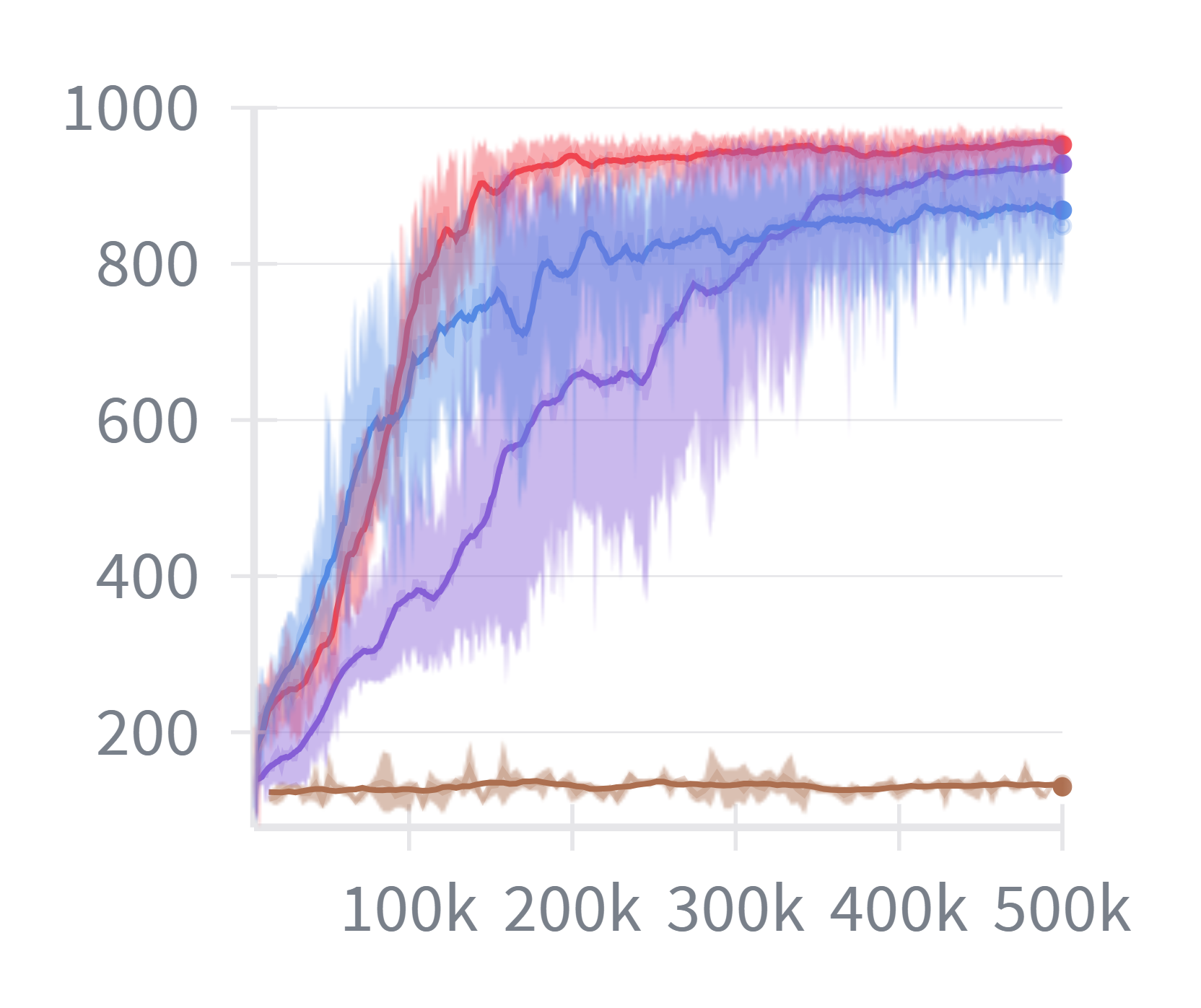}} &
        \subfloat[\textit{Walker Walk}]{
            \centering
            \includegraphics[width=0.24\textwidth]{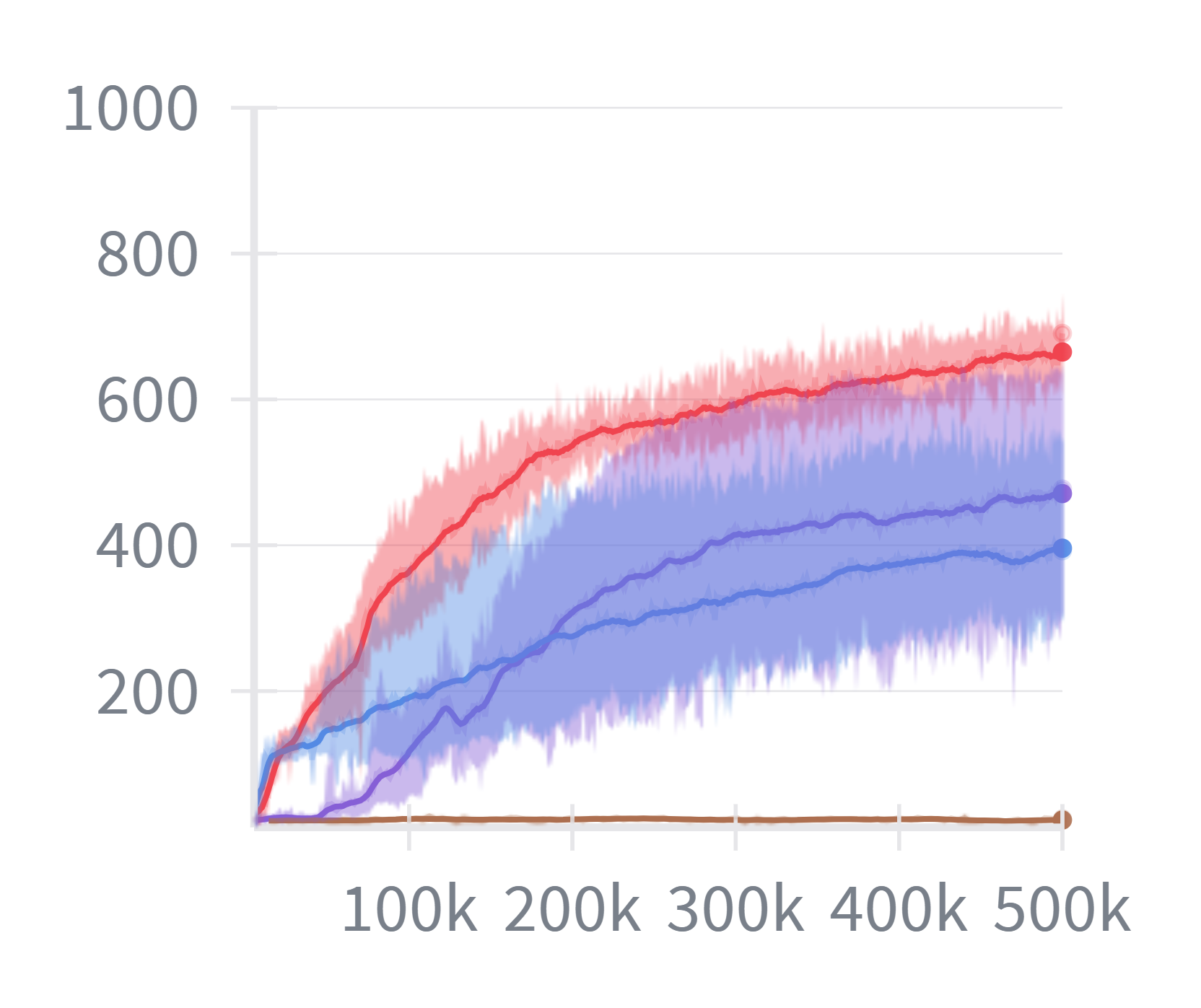}} \\
        
        \subfloat[\textit{Reacher Easy}]{
            \centering
            \includegraphics[width=0.24\textwidth]{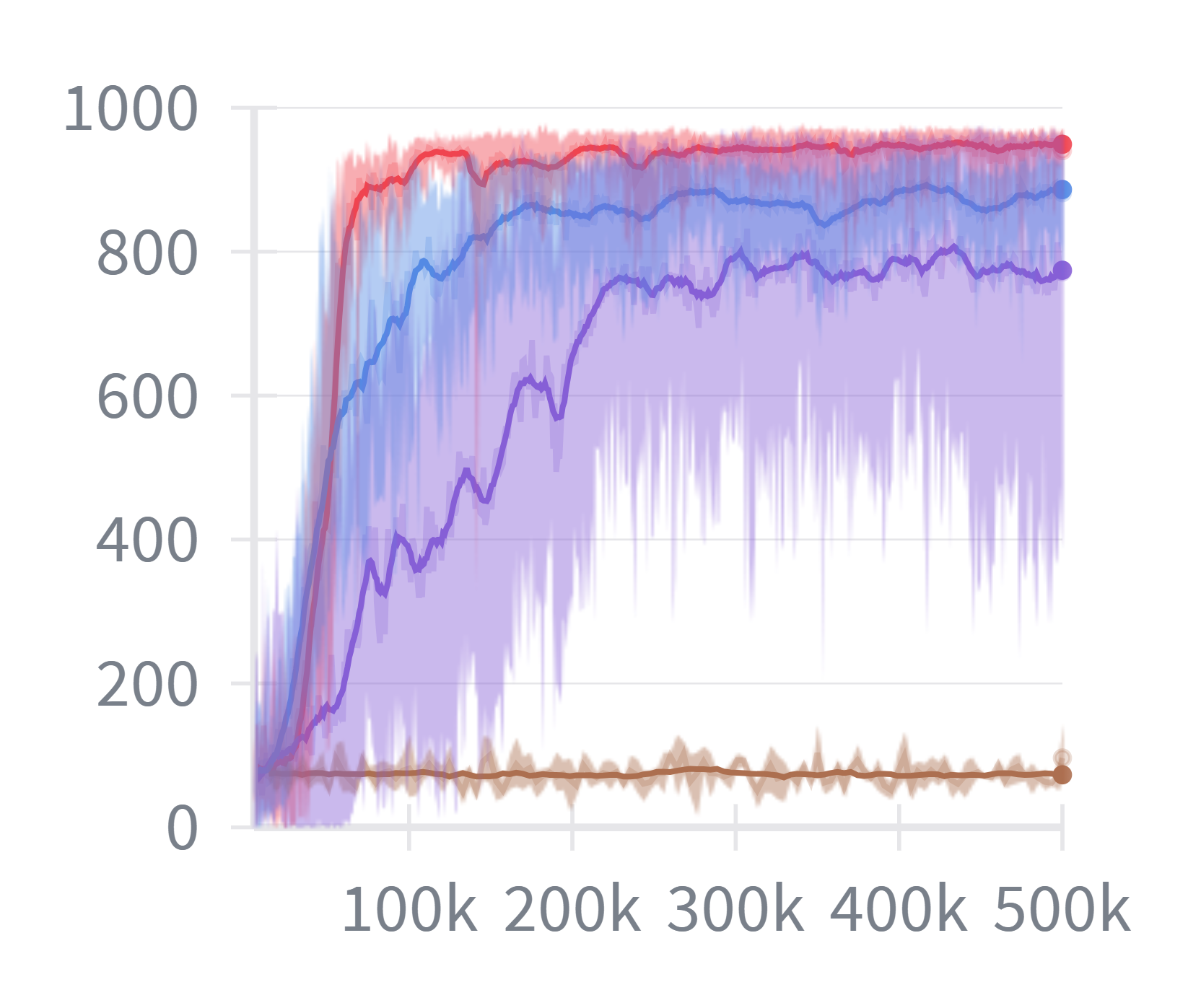}} &
        \subfloat[\textit{Reacher Hard}]{
            \centering
            \includegraphics[width=0.24\textwidth]{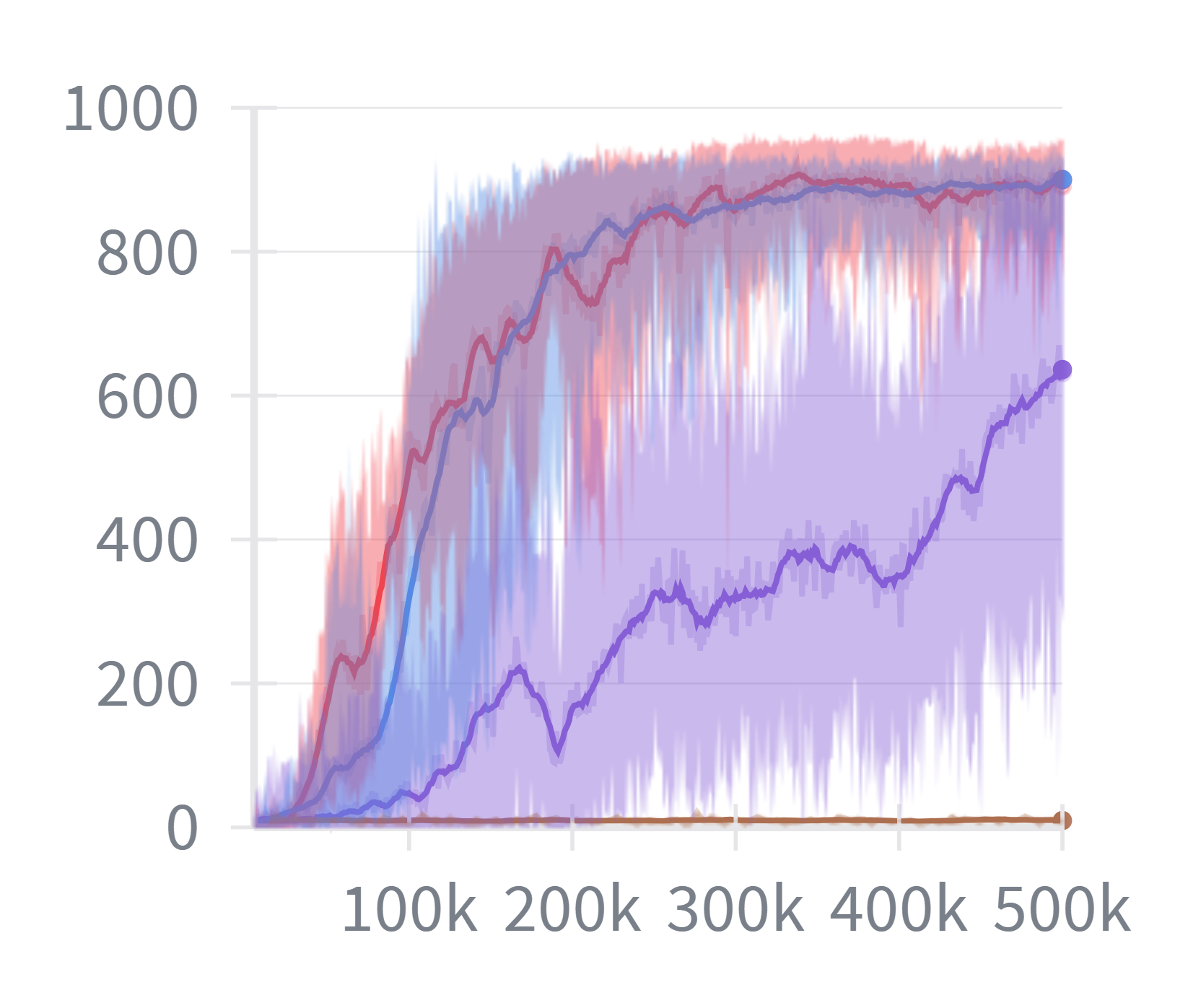}} &
        \subfloat[\textit{Hopper Hop}]{
            \centering
            \includegraphics[width=0.24\textwidth]{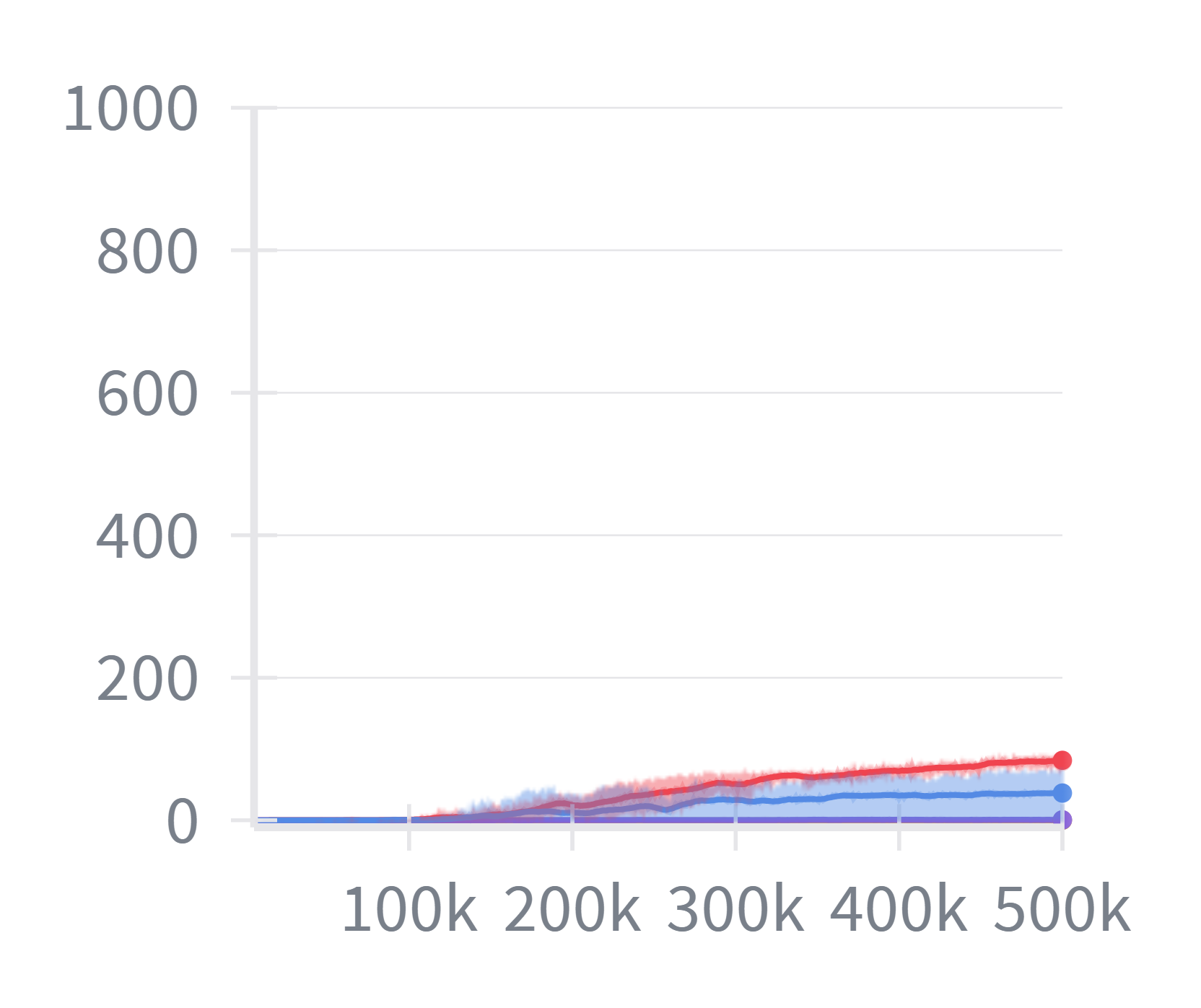}} &
        \subfloat[\textit{Cartpole Swingup}]{
            \centering
            \includegraphics[width=0.24\textwidth]{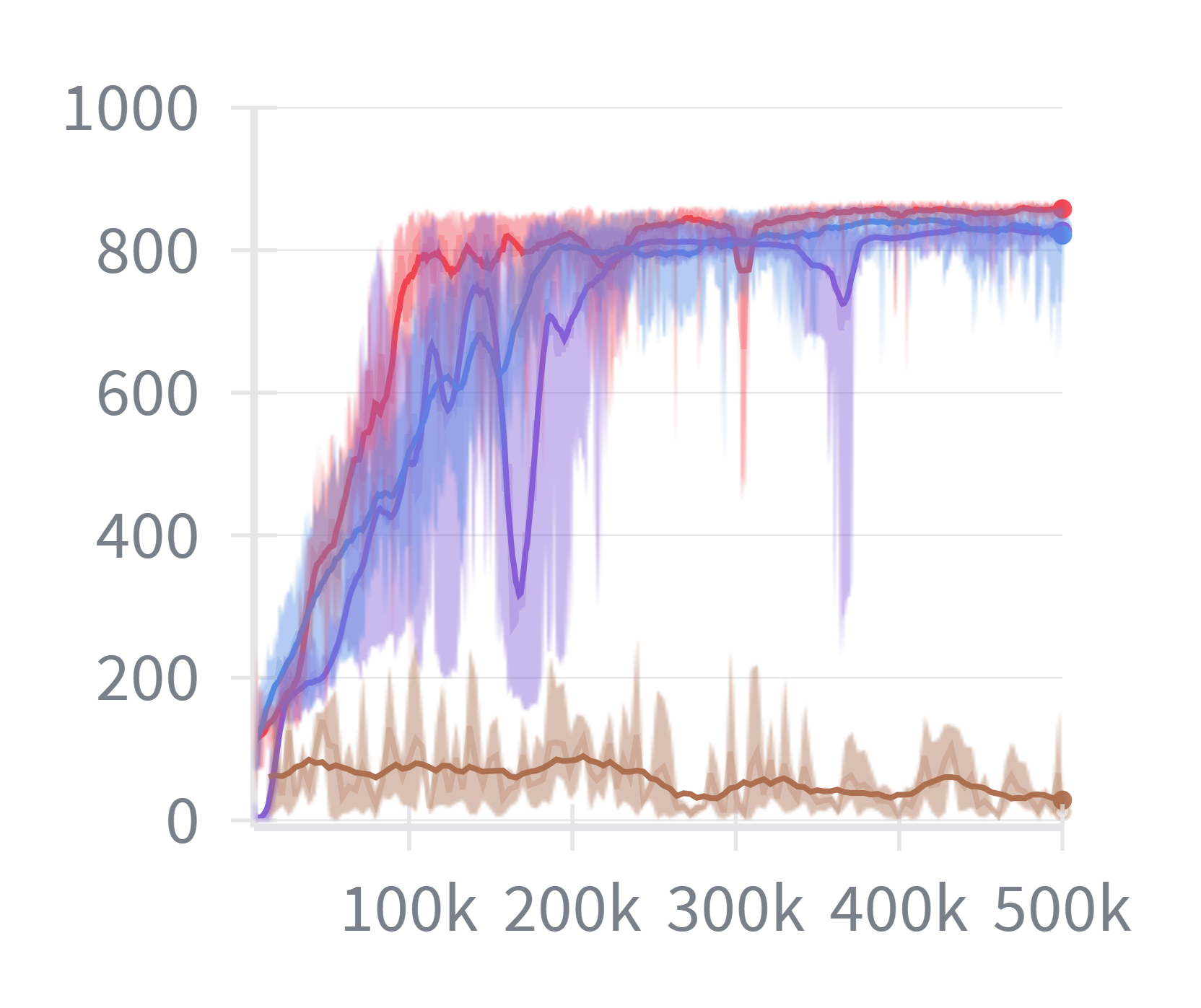}} \\
    \end{tabular}
    \caption{Performance of different methods on delayed DMC tasks}
    \label{fig:ave_success_ratio_delay5}
\end{figure}

\textbf{Experimental design\quad}
To further examine the adaptability of different methods under random delays, this section systematically compares the methods on 8 DMC benchmark tasks with random delays. The results are shown in Figure~\ref{fig:ave_success_ratio_delay5}, where the horizontal axis denotes the number of real episodes used for training, and the vertical axis denotes average return. Compared with the no-delay experiments, this group of experiments focuses more on learning stability, state-modeling capability, and long-term decision robustness when observations and action feedback are temporally misaligned.

\textbf{Experimental results\quad}
As shown in Figure~\ref{fig:ave_success_ratio_delay5}, after random delays are introduced, all methods show performance degradation to varying degrees, but the degree of degradation differs. Overall, \textbf{CausalDreamer} still maintains the best comprehensive performance. In most tasks, its convergence speed, final return, and training stability remain leading. Even on more difficult tasks where the overall returns are generally low, it still maintains a relative advantage. This shows that the implicit causal modeling and structured state representation adopted by CausalDreamer are not only effective in standard environments. When observations are disturbed by random delays, CausalDreamer can still provide relatively stable support for state tracking and world-model prediction, resulting in a smaller overall performance drop.

\textbf{DreamerV3} also remains competitive. It can still converge to relatively high returns on most tasks, and its overall performance is clearly better than SAC and CWM. From the curves in the figure, DreamerV3 still shows good early-stage learning speed and final performance on some tasks. However, compared with the no-delay scenario, its fluctuations increase and the performance degradation becomes more pronounced in some tasks. This indicates that the recurrent state modeling provided by the recurrent state-space model can alleviate the influence of observation lag to some extent, but when delays further disrupt temporal correspondence, the recovery capability based only on a unified latent representation remains limited.

\textbf{CWM} becomes substantially less stable. Its returns remain low on most tasks, and the training process lacks sufficient competitiveness. The reason is that delays further disturb the original temporal correspondence among states, increasing the difficulty of explicit structure discovery. Once causal-structure learning is biased, the errors can continue to propagate into subsequent prediction and control processes, ultimately affecting overall performance.

\textbf{SAC}, as a model-free method, is also strongly affected under random delays. Although it can still improve gradually on a few tasks, its overall convergence is slower, training fluctuations are larger, and final performance generally lags behind the two world-model-based methods. This is consistent with the method's characteristics. When input observations are randomly misaligned with the true environmental state, the absence of state recovery and dynamic compensation from a world model makes its value estimation and policy update more directly affected by the disturbance.

\subsubsection{Structured Representation Ability of the Field-Node Encoder}

\begin{figure}
    \centering
    \setlength{\tabcolsep}{0pt}
    \includegraphics[width=0.85\textwidth]{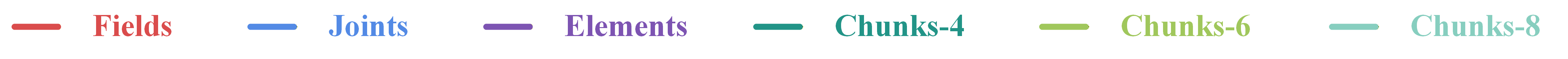}
    
    \begin{tabular}{cccc}
        \subfloat[\textit{Cheetah Run}]{
            \centering
            \includegraphics[width=0.24\textwidth]{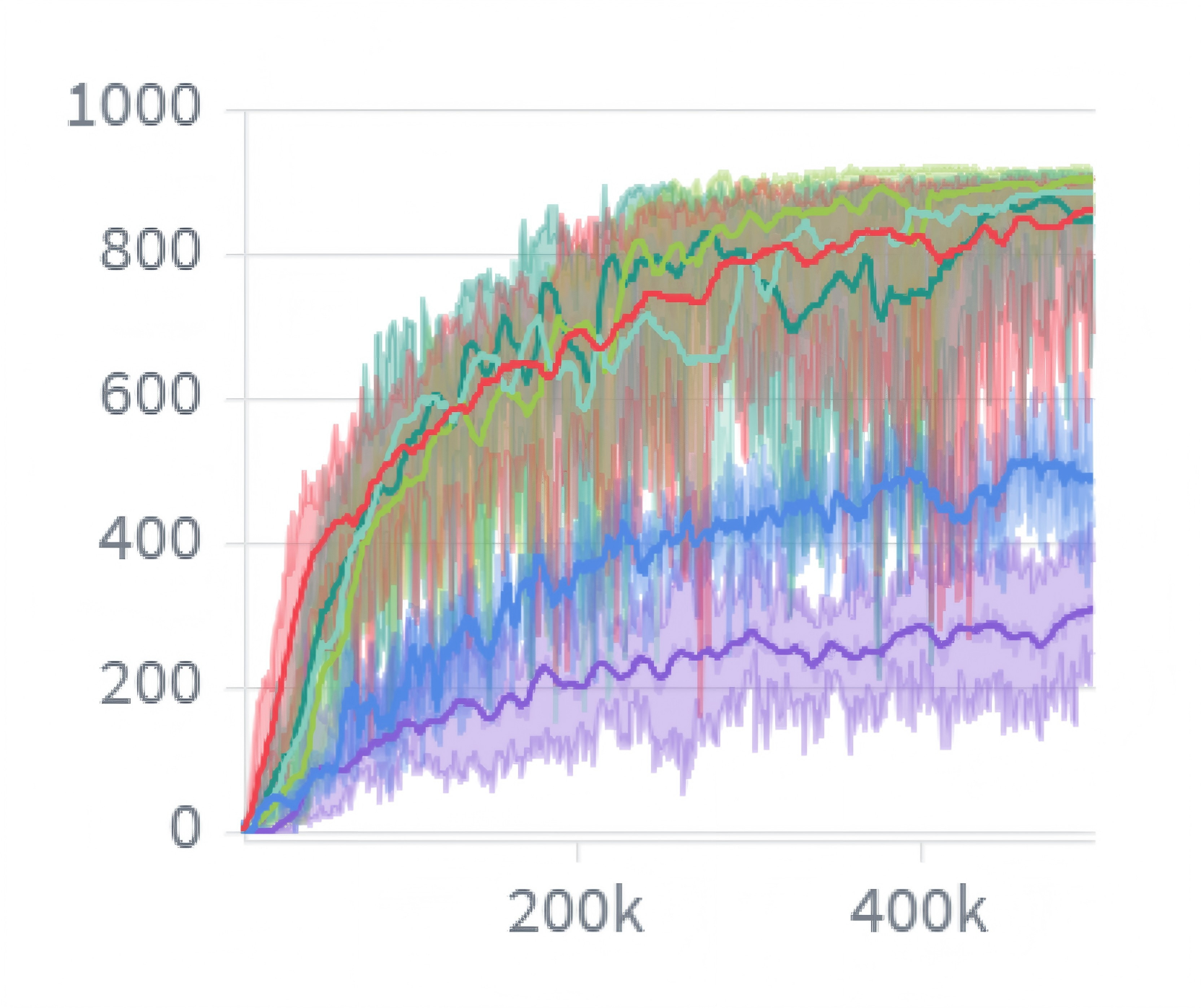}} &
        \subfloat[\textit{Walker Run}]{
            \centering
            \includegraphics[width=0.24\textwidth]{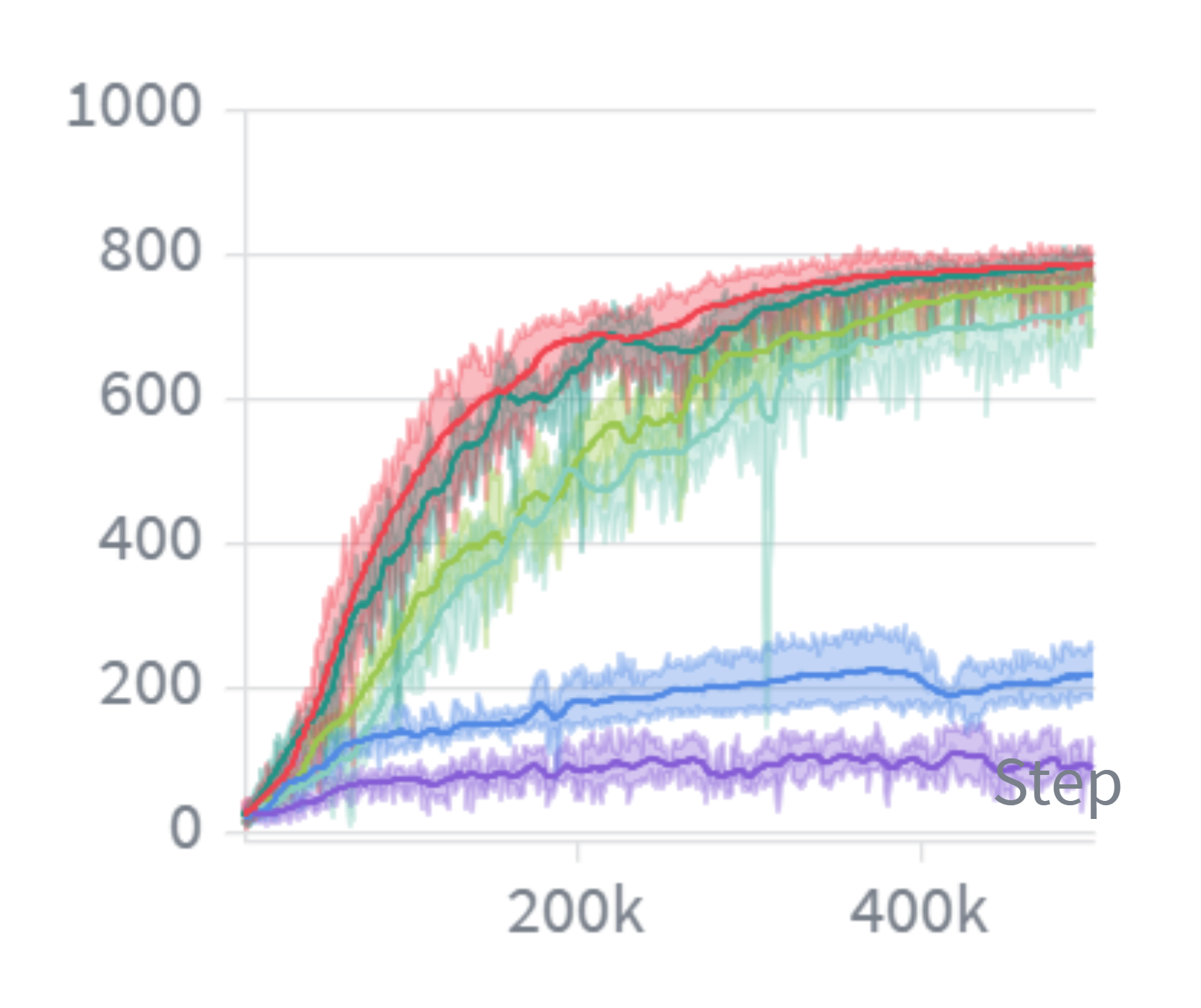}} &
        \subfloat[\textit{Hopper Hop}]{
            \centering
            \includegraphics[width=0.24\textwidth]{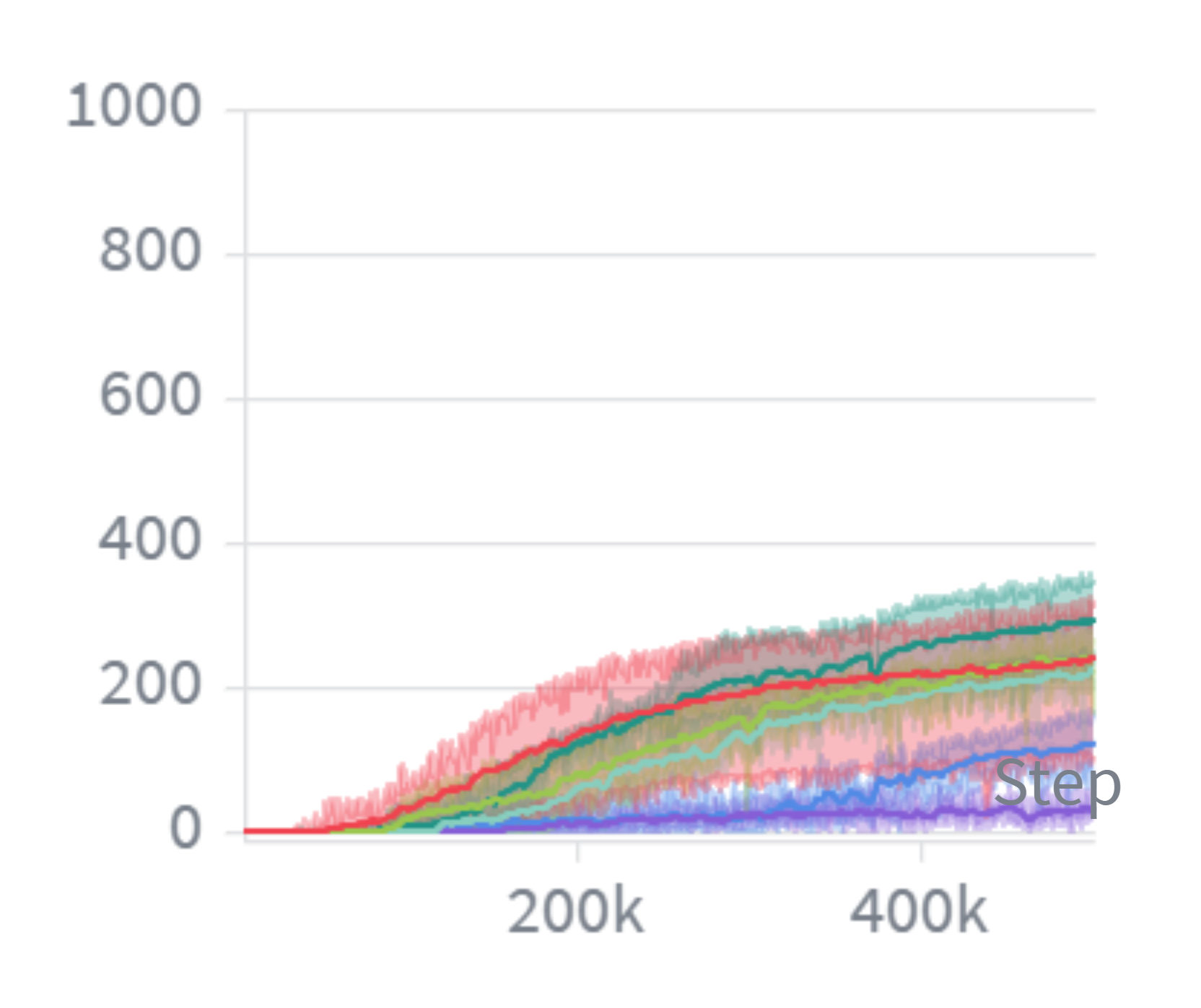}} &
        \subfloat[\textit{Cartpole Swingup}]{
            \centering
            \includegraphics[width=0.24\textwidth]{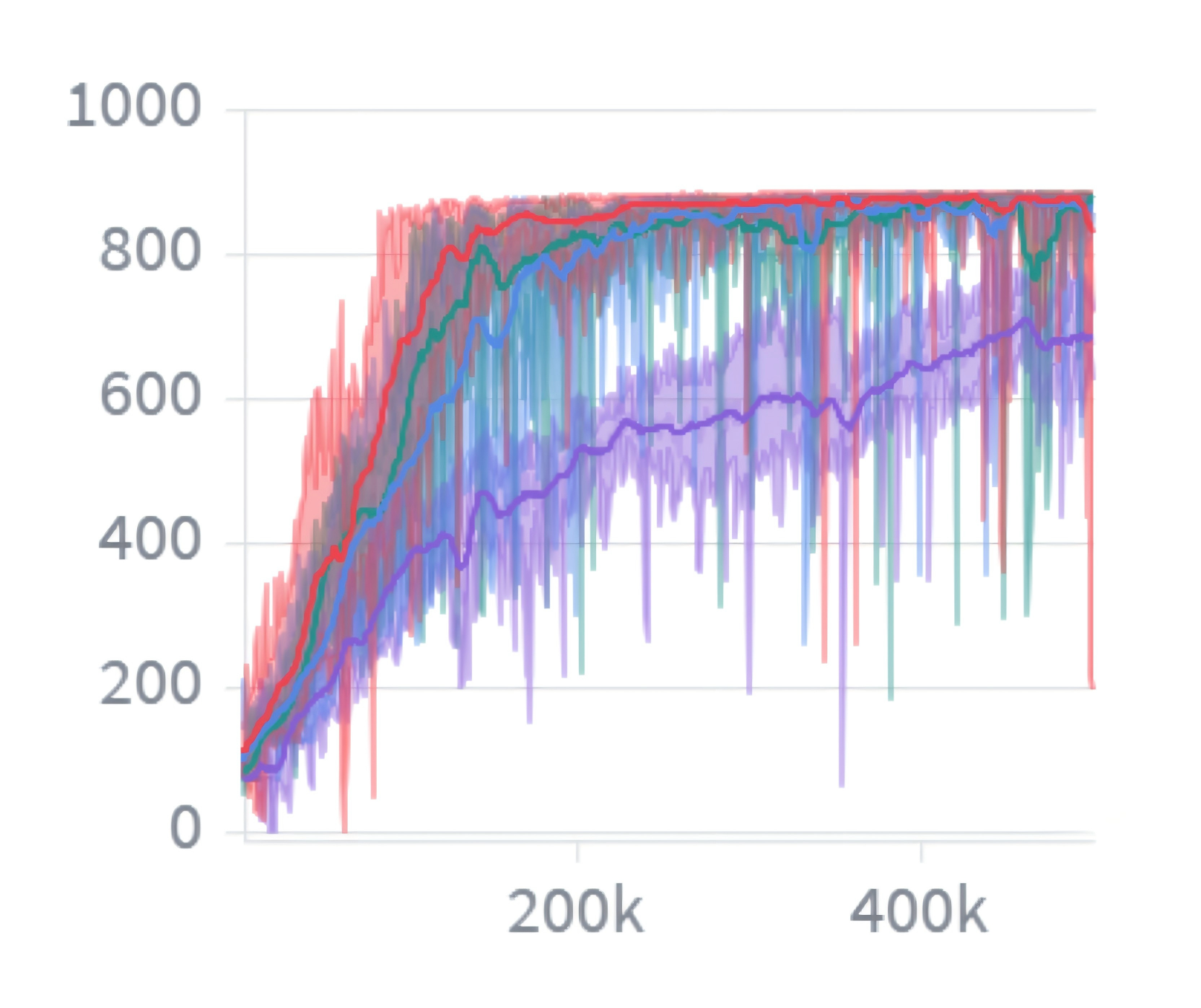}} \\
    \end{tabular}
    \caption{Overall performance comparison of CausalDreamer under different encoder partition strategies}
    \label{fig:different_encoder}
\end{figure}

\textbf{Experimental design\quad}
To evaluate the structured representation ability of the field-node encoder, this section compares four node-partition strategies:
\begin{itemize}
    \item \textbf{Field-wise}: Explicit semantic fields provided by the environment, such as position, velocity, and angle, are directly treated as independent nodes. Each field corresponds to one node, preserving the semantic integrity of the original observation as much as possible.
    \item \textbf{Joint-wise}: According to the physical structure of the robot, observation attributes related to the same joint are aggregated into one node. For example, for a robot with $M$ joints, the position, velocity, and angle information of each joint can be combined into one node, forming $M$ nodes. This strategy emphasizes physical integrity at the joint level.
    \item \textbf{Chunk-wise}: All observation fields are concatenated into a unified vector and then uniformly divided by fixed dimensions. Let the total observation dimension be $d$ and the number of chunks be $n$; then each chunk has dimension $\lfloor d/n \rfloor$, and the last chunk contains the remaining dimensions. This strategy provides a uniform partitioning baseline that does not rely on semantic information.
    \item \textbf{Element-wise}: Each scalar element in the observation vector is treated as an independent node, so each dimension corresponds to one node. For a $d$-dimensional observation, this forms $d$ one-dimensional nodes. This strategy corresponds to the finest-grained structured representation, but it does not preserve the original local correlations among observation elements.
\end{itemize}
While keeping the overall encoder architecture unchanged, the above four strategies only change the node-partition method, so as to fairly compare the impact of different degrees of encoder structuring on the overall performance of CausalDreamer. The experimental results are shown in Figure~\ref{fig:different_encoder}, where the horizontal axis denotes the number of real episodes used for training, and the vertical axis denotes average return.

\textbf{Experimental results\quad}
Figure~\ref{fig:different_encoder} shows that different node-partition strategies indeed affect model performance, but the effect does not follow a simple rule such as finer granularity always being better or more intuitive partitioning always being better. Overall, field-wise and chunk-wise partitioning are the most stable in most tasks, and their results are relatively close. In contrast, joint-wise and element-wise partitioning perform worse overall.

\textbf{Field-wise partitioning} generally performs best or close to the best. Field-wise partitioning is usually consistent with the original semantic structure of the observation, and therefore can naturally preserve existing local correlations among state variables. Based on this partitioning, the model can more easily form stable node representations during learning and further characterize dynamic dependencies among nodes. Therefore, this partitioning usually provides a good modeling starting point for the world model.

\textbf{Chunk-wise partitioning}, although not relying on environmental semantics or physical structure, performs comparably to field-wise partitioning in most tasks, with only small gaps in some cases. This result indicates that CausalDreamer does not strictly rely on precise prior structural information. As long as the initial partitioning does not completely destroy local correlations, the model can still gradually extract effective implicit relations through subsequent learning. In other words, the field-node encoder and the subsequent dynamics modeling process have a certain degree of adaptability, and can recover useful structural information for the task even from coarse-grained partitions without explicit semantics.

\textbf{Joint-wise partitioning} performs relatively worse. Although this strategy has intuitive physical meaning, it assumes that observation variables should mainly be aggregated within each joint, which may not always hold. In some tasks, different joints are themselves strongly coupled, and some key state variables may not be well organized by simple joint boundaries. Therefore, although joint-wise partitioning seems reasonable, it may restrict the model's ability to represent cross-unit interactions to some extent.

\textbf{Element-wise partitioning} performs the worst, which more clearly demonstrates the problem caused by overly fine node partitioning. When each node contains only a single scalar, local semantic consistency is substantially weakened, and it becomes more difficult for the model to form stable and discriminative node-level representations. As a result, subsequent dynamics modeling and information exchange lack sufficient local semantic support, ultimately affecting the overall modeling quality.

This group of experiments shows that CausalDreamer is better suited to node partitions with certain local organization, but it does not require the partitioning to depend on precise manually designed priors. Field-wise partitioning performs best, but chunk-wise partitioning can still achieve comparable results when explicit semantic information is unavailable. This indicates that the method remains applicable when structural information is incomplete. In contrast, overly rigid or overly fragmented partitioning is not conducive to fully exploiting the model's structured modeling capability.

\subsubsection{Effectiveness of the Message-Passing Mechanism in the Causal Field-Node State-Space Model}

\begin{figure}
    \centering
    \setlength{\tabcolsep}{0pt}
    \includegraphics[width=0.45\textwidth]{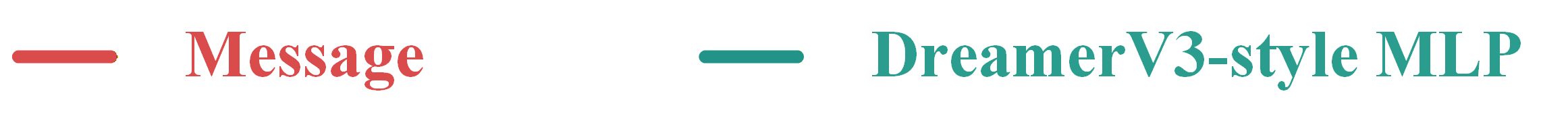}
    
    \begin{tabular}{cccc}
        \subfloat[\textit{Cheetah Run}]{
            \centering
            \includegraphics[width=0.24\textwidth]{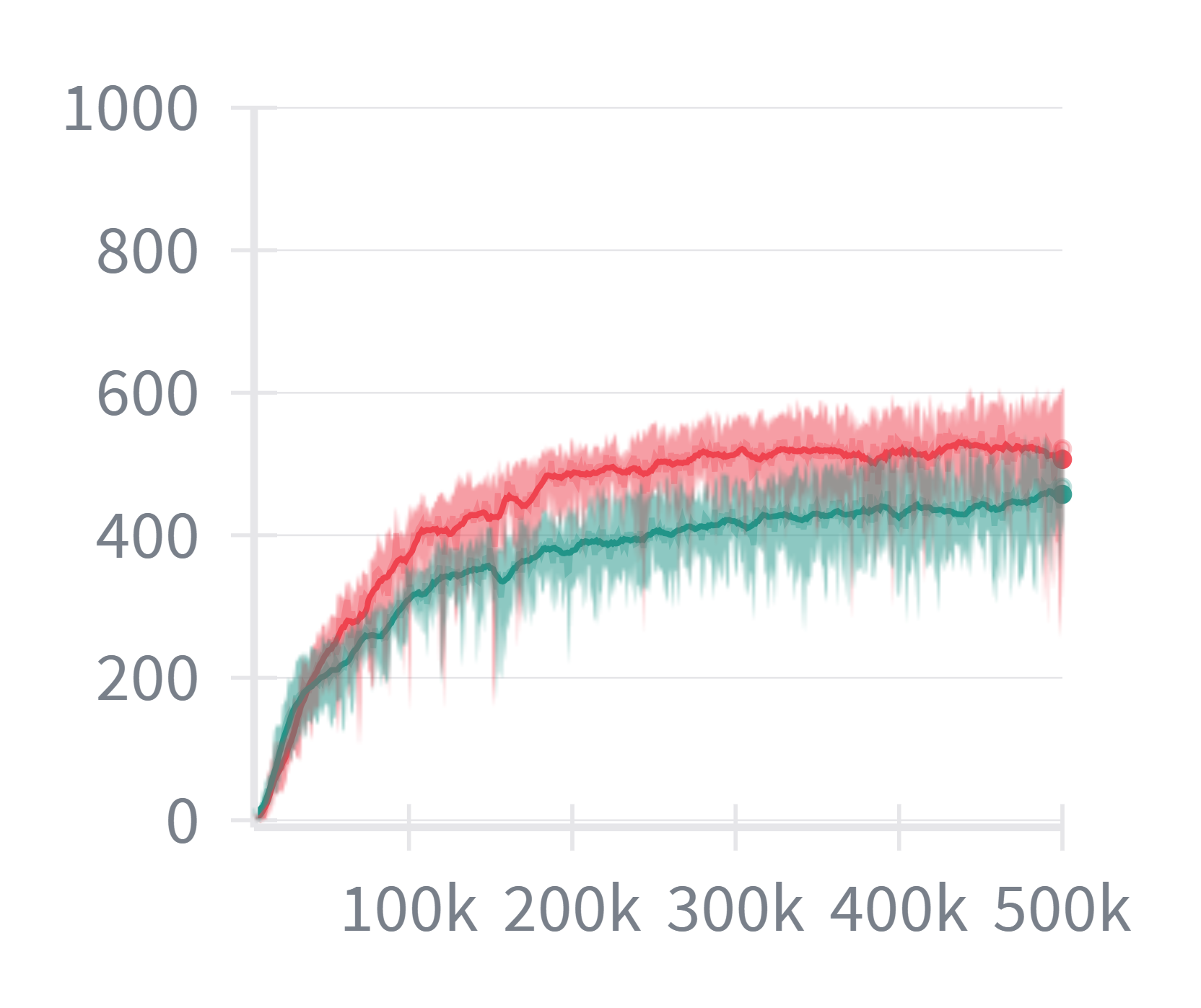}} &
        \subfloat[\textit{Walker Run}]{
            \centering
            \includegraphics[width=0.24\textwidth]{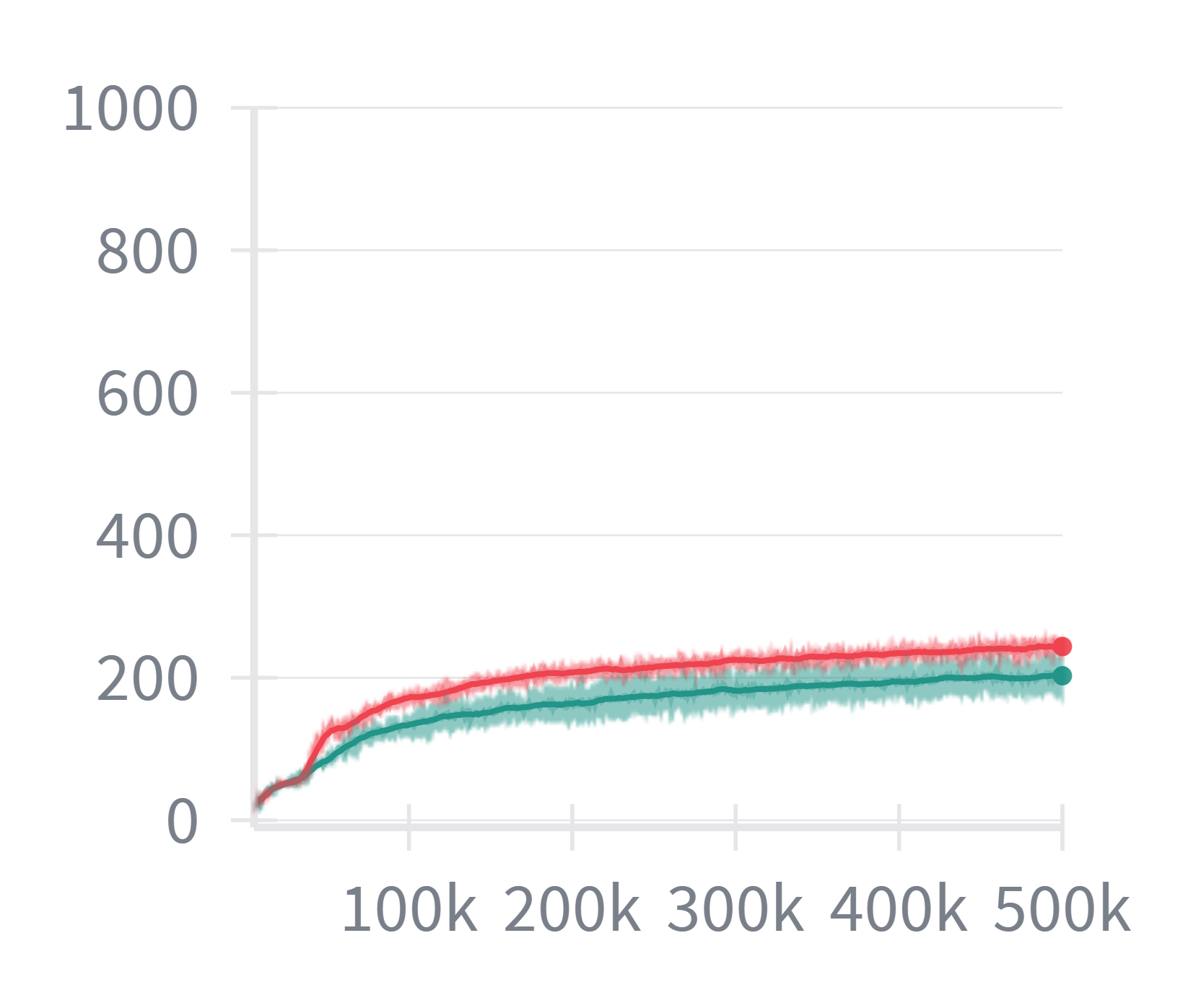}} &
        \subfloat[\textit{Walker Stand}]{
            \centering
            \includegraphics[width=0.24\textwidth]{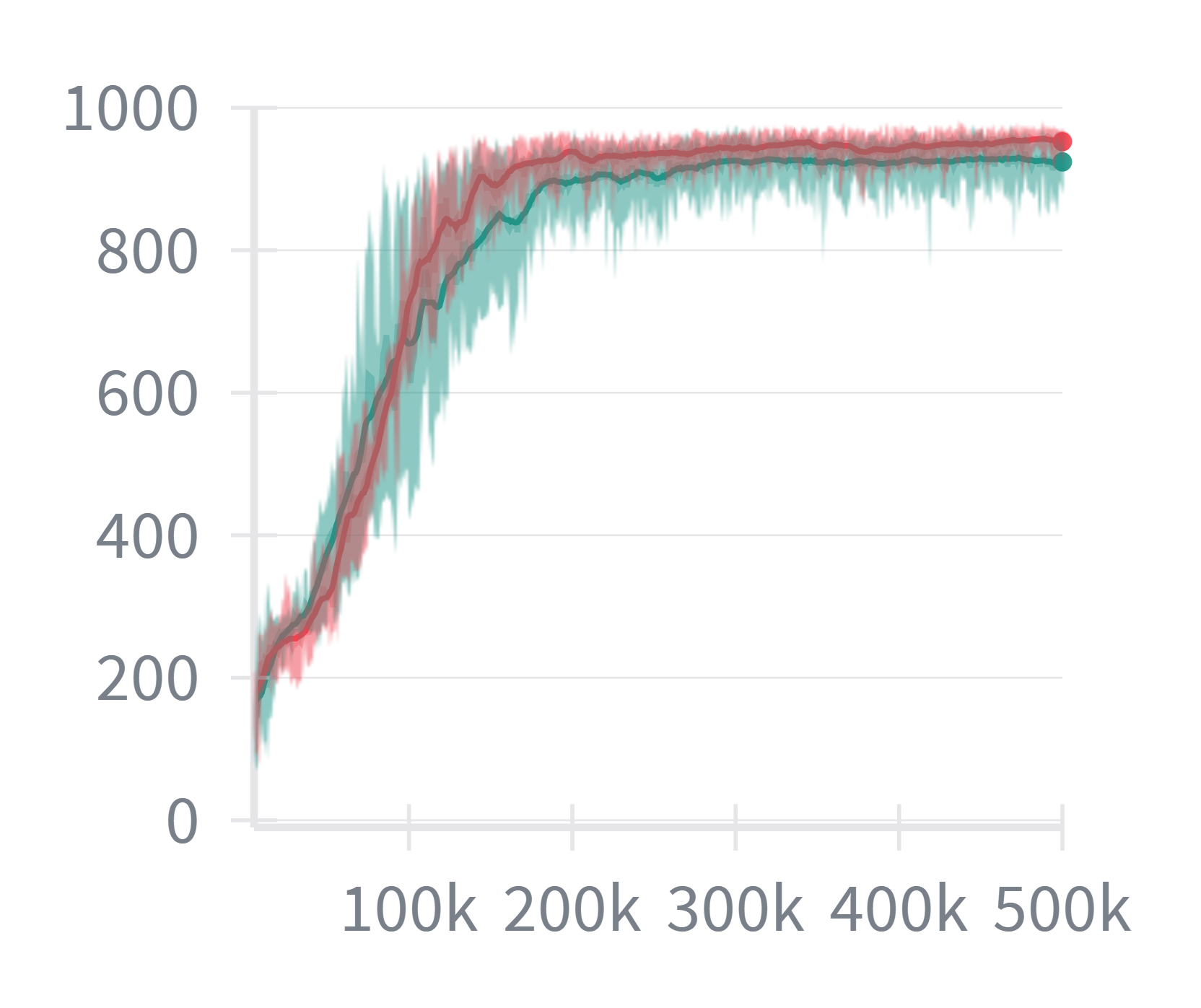}} &
        \subfloat[\textit{Walker Walk}]{
            \centering
            \includegraphics[width=0.24\textwidth]{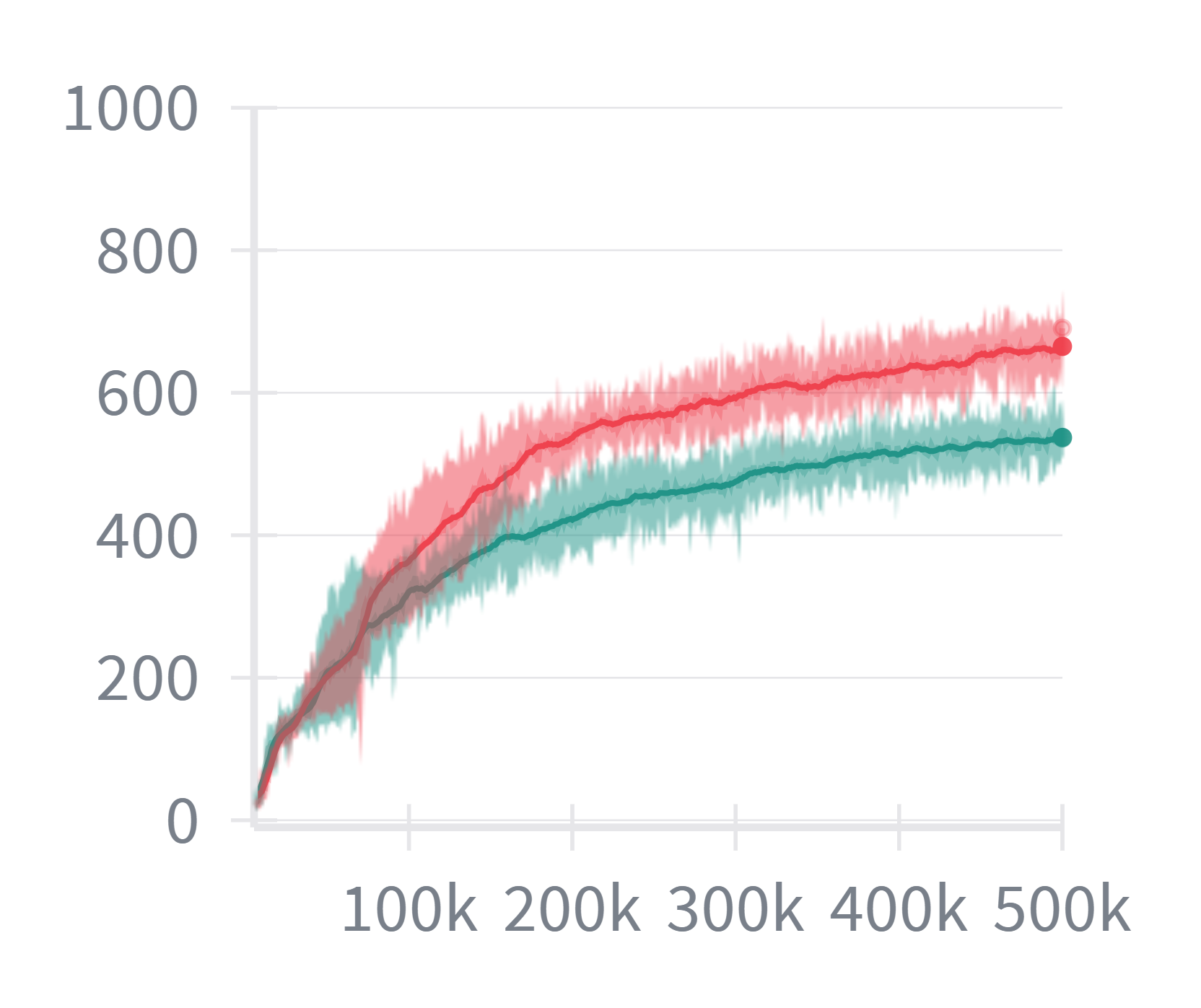}} \\
        
        \subfloat[\textit{Reacher Easy}]{
            \centering
            \includegraphics[width=0.24\textwidth]{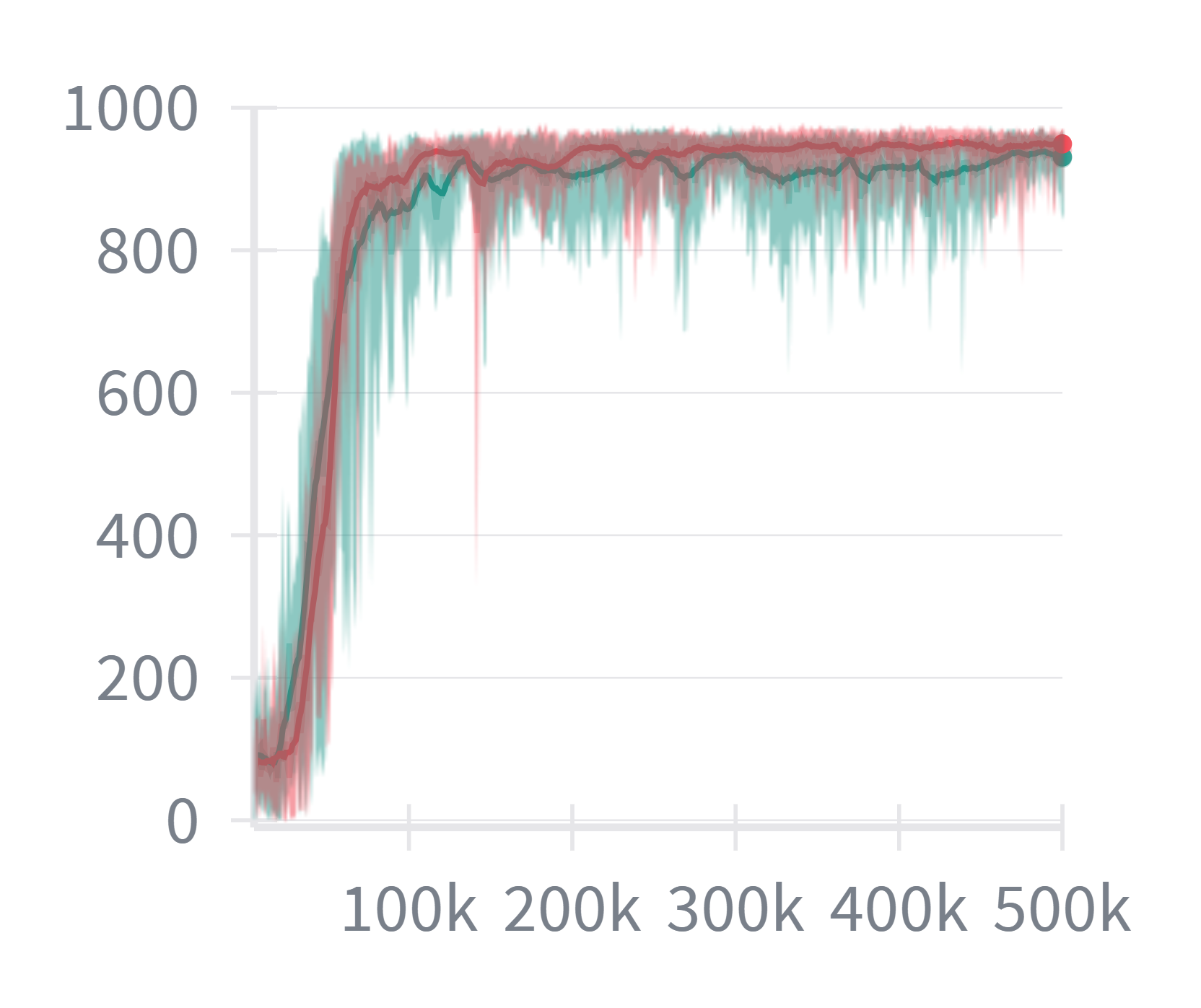}} &
        \subfloat[\textit{Reacher Hard}]{
            \centering
            \includegraphics[width=0.24\textwidth]{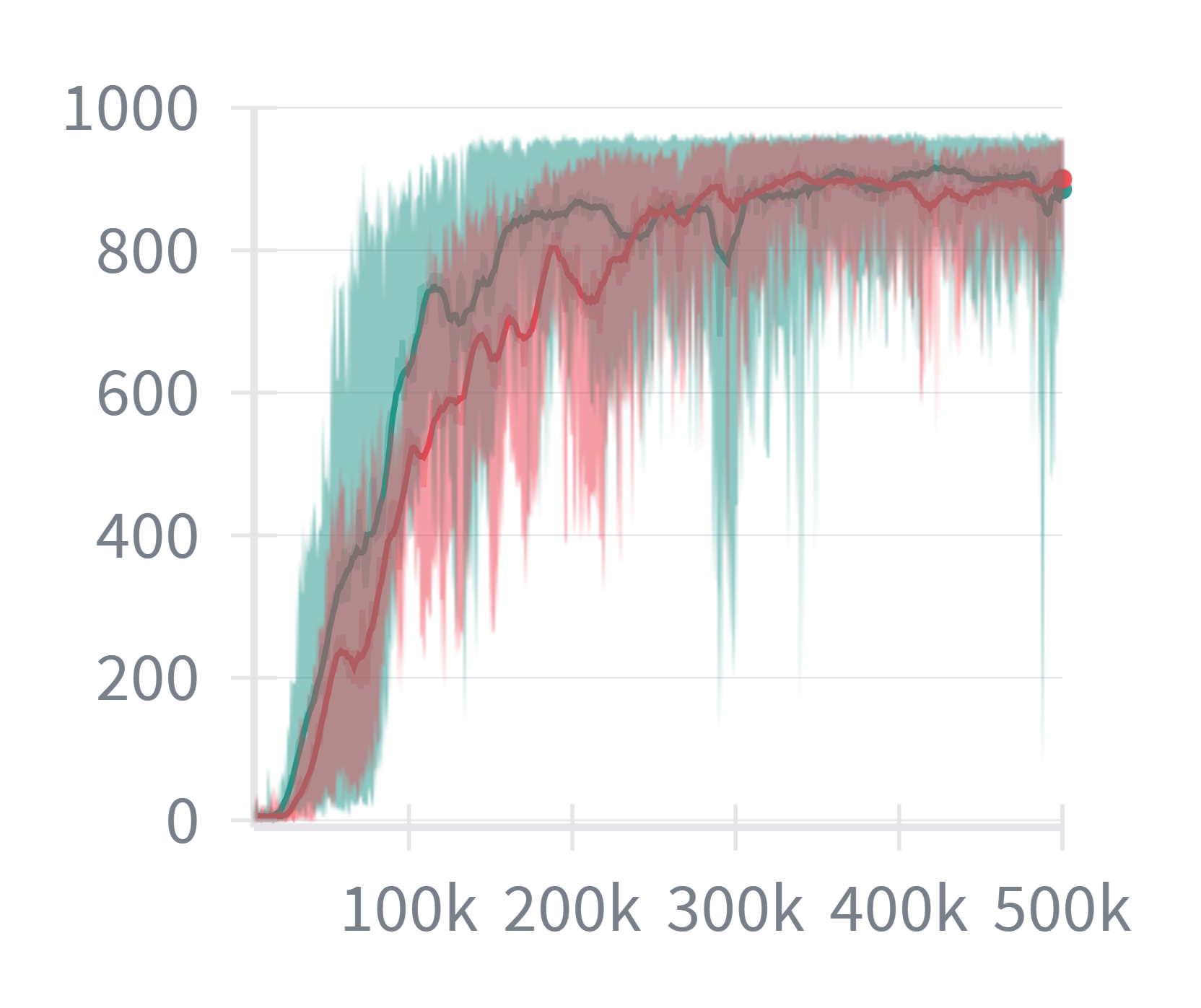}} &
        \subfloat[\textit{Hopper Hop}]{
            \centering
            \includegraphics[width=0.24\textwidth]{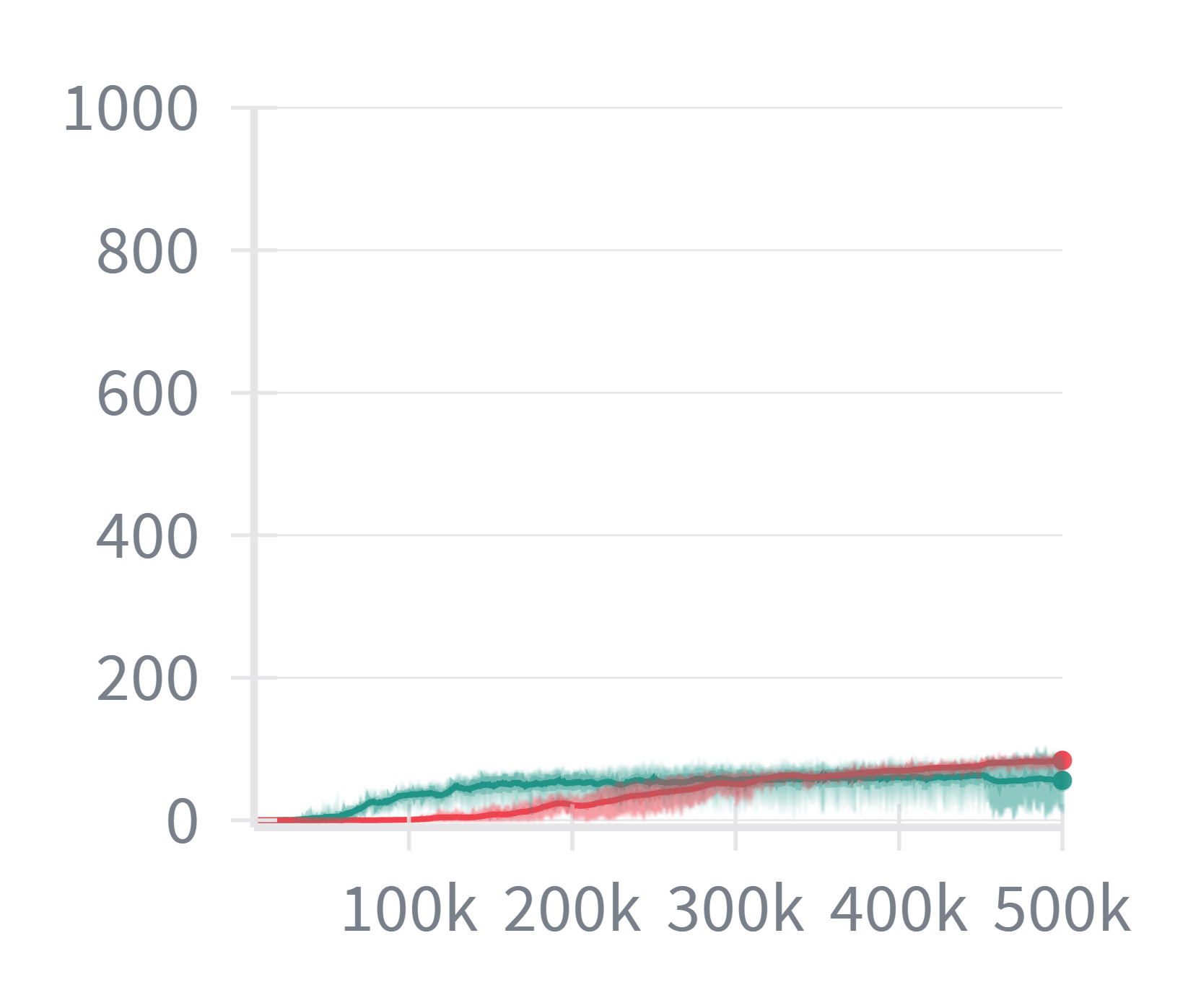}} &
        \subfloat[\textit{Cartpole Swingup}]{
            \centering
            \includegraphics[width=0.24\textwidth]{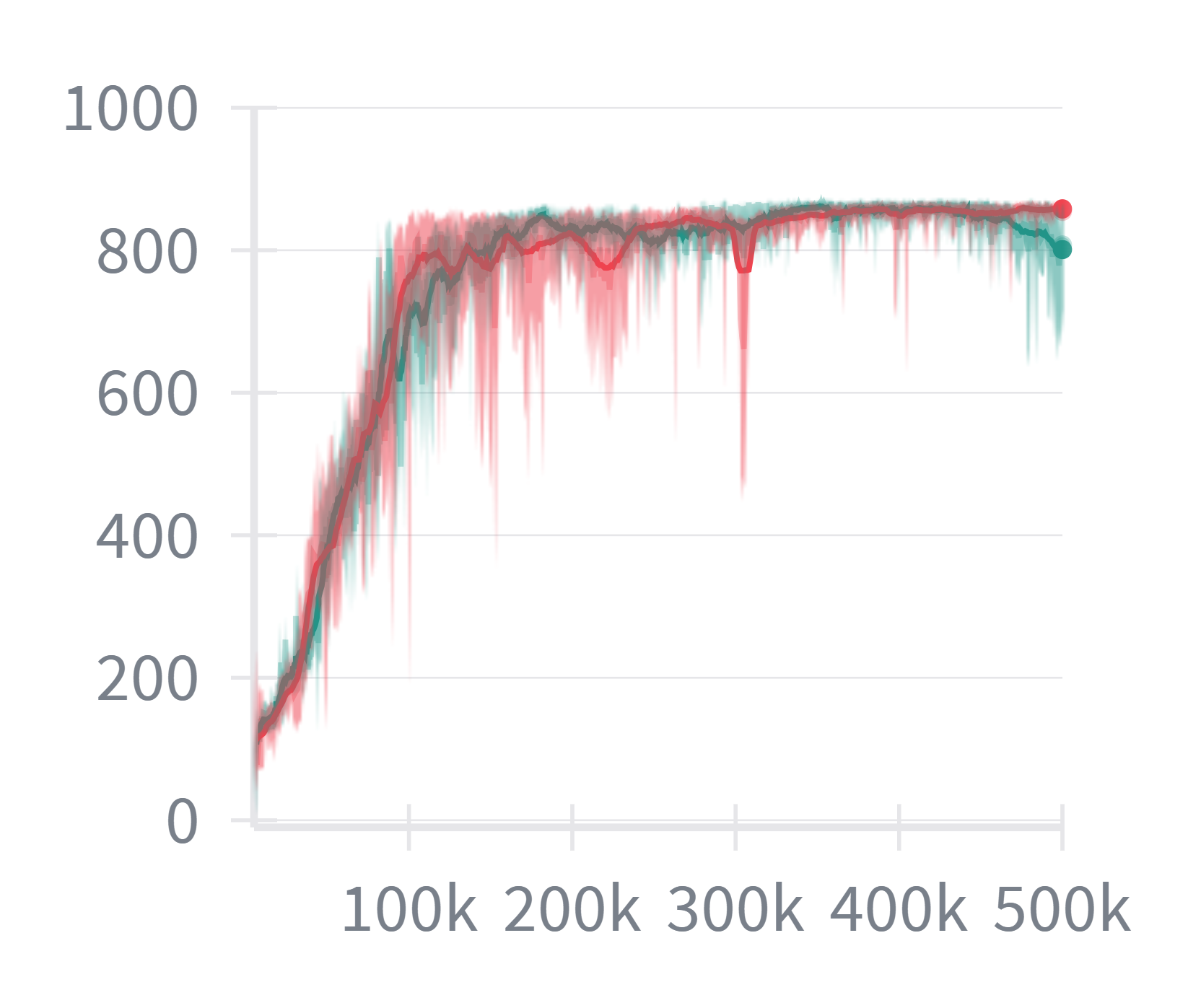}} \\
    \end{tabular}
    \caption{Ablation comparison of the message-passing mechanism}
    \label{fig:abo2}
\end{figure}

\textbf{Experimental design\quad}
In the proposed CausalDreamer, the causal field-node state-space model performs posterior modeling of stochastic states through message passing among nodes. To verify the role of this mechanism, this section designs an ablation experiment. While keeping the node-partition strategy and the overall method framework unchanged, the message-passing structure in posterior inference is replaced with the global multilayer perceptron (MLP) used in DreamerV3. Figure~\ref{fig:abo2} compares the performance of the two configurations on multiple tasks, where the horizontal axis denotes the number of real episodes used for training, and the vertical axis denotes average return.

\textbf{Experimental results\quad}
As shown in Figure~\ref{fig:abo2}, on most tasks, the causal field-node state-space model with the message-passing mechanism performs better overall than its ablated variant, indicating that explicitly introducing information exchange among nodes during posterior modeling is beneficial. The \textbf{ablated variant} replaces the original message-passing-based posterior inference with global-MLP-based posterior inference. In other words, it performs unified inference directly based on the whole observation, without distinguishing the local states of different nodes or their interactions. Although this simplifies the model structure, it also makes the structural information in the observation more likely to be compressed and weakened during unified representation, thereby limiting the posterior distribution's ability to characterize complex dynamic relations.

In contrast, the \textbf{complete causal field-node state-space model} performs posterior modeling at the node level and explicitly characterizes dependencies among nodes through multiple rounds of message passing. As a result, each stochastic node can not only use its corresponding local observation, but also incorporate contextual information from other relevant nodes, thereby more fully reflecting the coupling relationships within the system. For continuous-control tasks involving multivariate interactions, this modeling approach is more conducive to recovering the true dynamic process, and also benefits subsequent observation reconstruction, future prediction, and policy learning based on imagined trajectories.

This ablation experiment shows that message passing is an important component for improving the quality of posterior modeling in the causal field-node state-space model. It helps the model make fuller use of structural information in observations and ultimately leads to more stable performance improvements.

\subsubsection{Cross-Task Knowledge Transfer of CausalDreamer}

\begin{figure}
    \centering
    \setlength{\tabcolsep}{0pt}
    \includegraphics[width=0.35\textwidth]{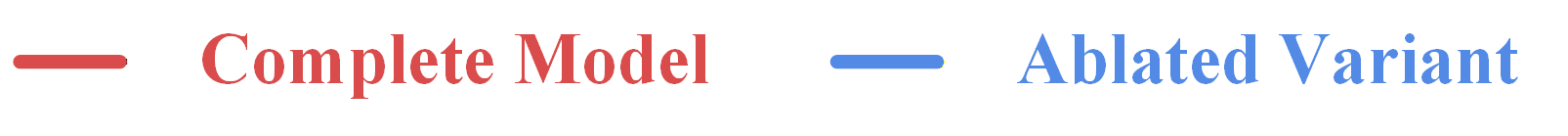}
    
    \begin{tabular}{cc}
        \subfloat[\textit{Walker: Stand $\to$ Walk}]{
            \centering
            \includegraphics[width=0.3\textwidth]{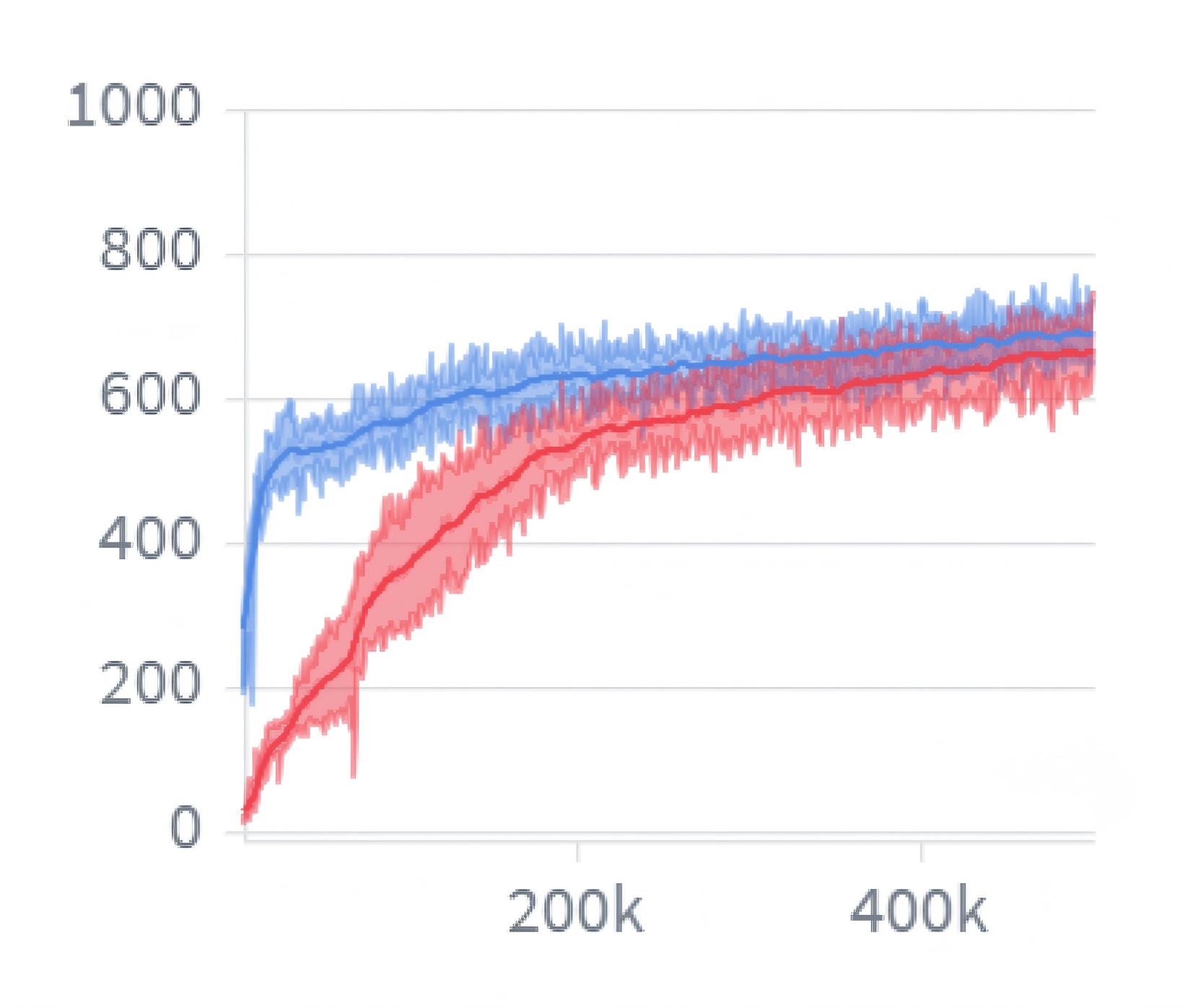}} &
        \subfloat[\textit{Walker: Walk $\to$ Run}]{
            \centering
            \includegraphics[width=0.3\textwidth]{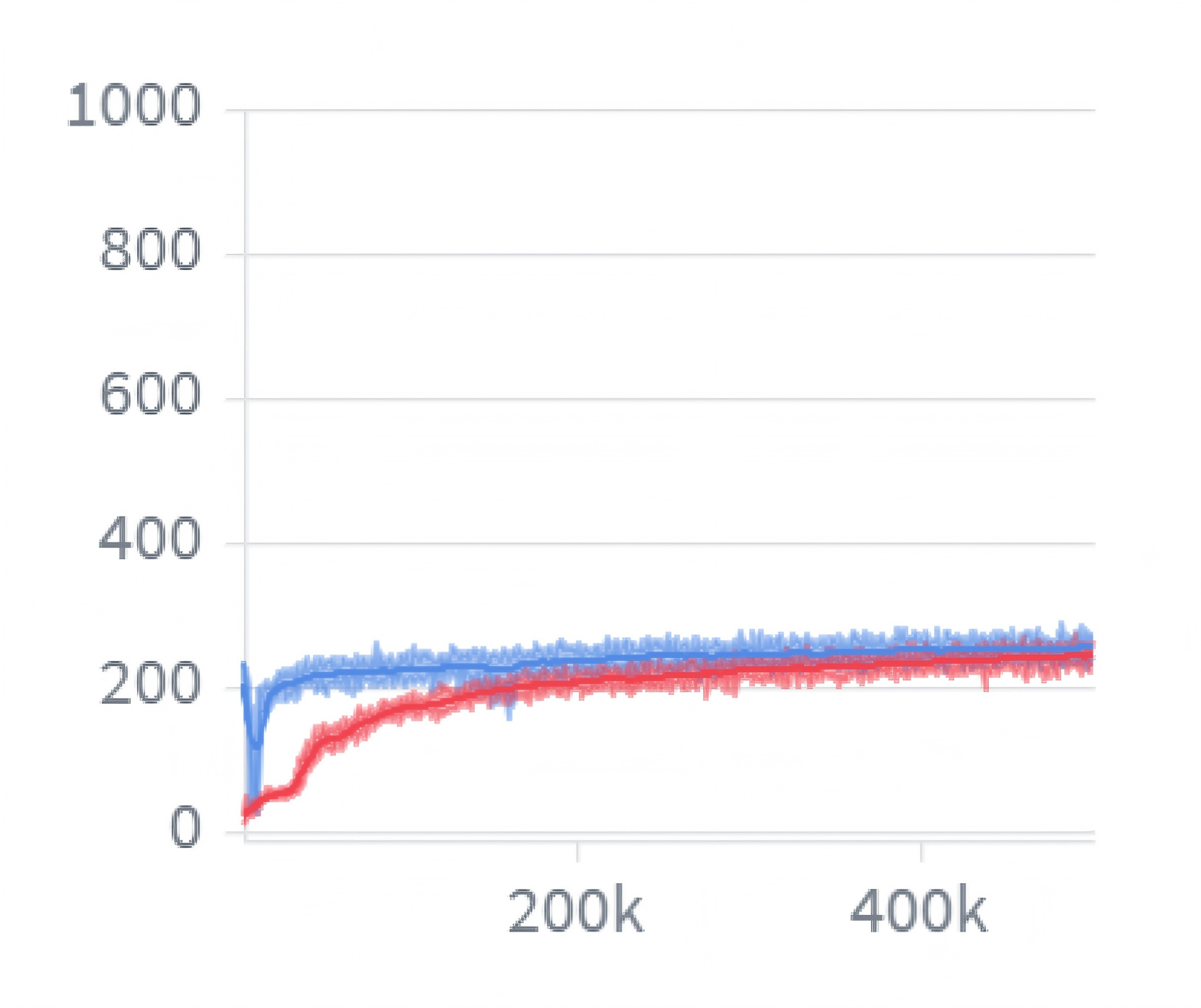}} \\
        \subfloat[\textit{Reacher: Easy $\to$ Hard}]{
            \centering
            \includegraphics[width=0.3\textwidth]{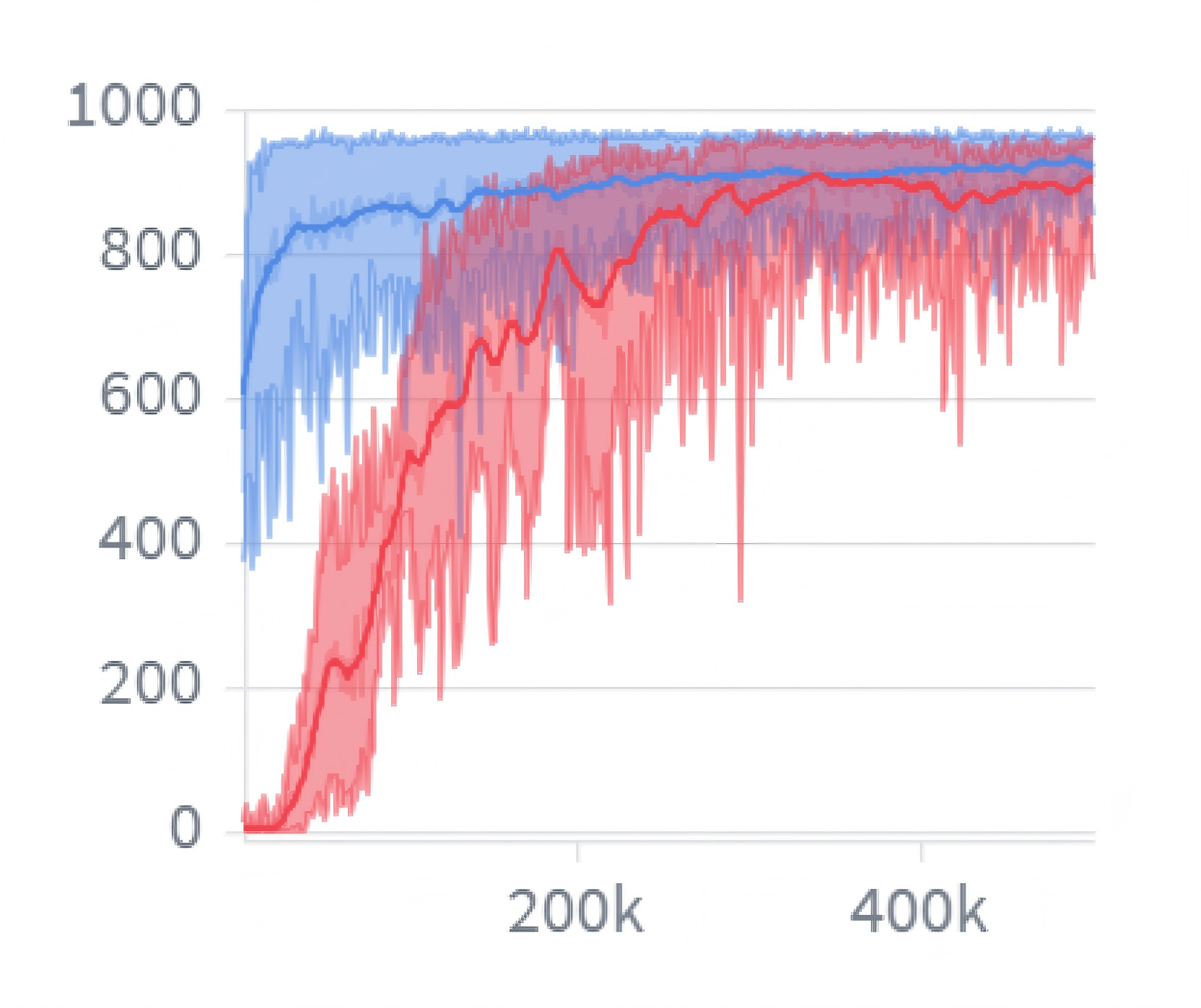}} &
        \subfloat[\textit{Hopper: Stand $\to$ Hop}]{
            \centering
            \includegraphics[width=0.3\textwidth]{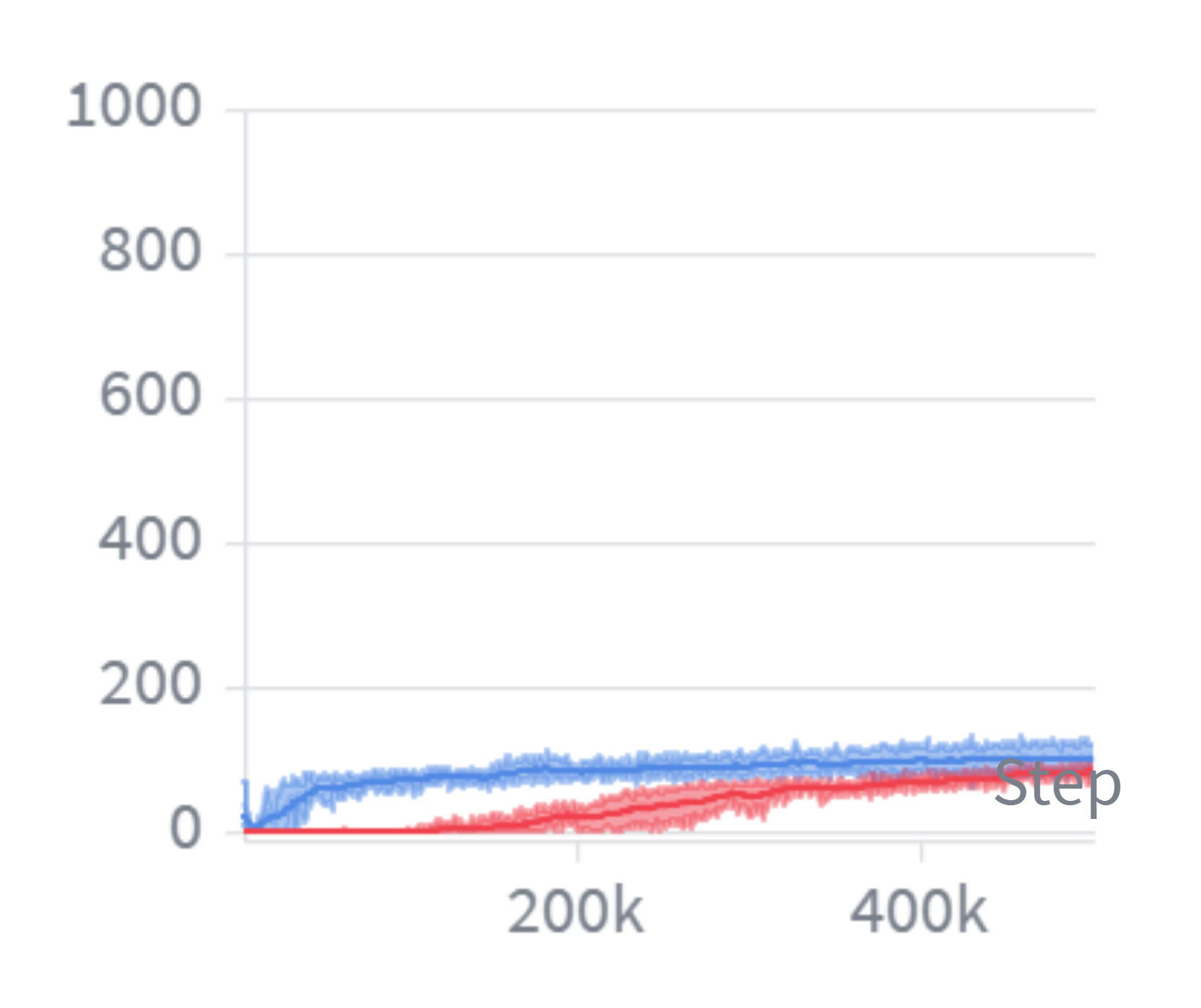}} \\
    \end{tabular}
    \caption{Performance of CausalDreamer in cross-task knowledge reuse from easier tasks to harder tasks}
    \label{fig:abo2_1}
\end{figure}

\begin{figure}
    \centering
    \setlength{\tabcolsep}{0pt}
    \includegraphics[width=0.35\textwidth]{transfer_legend2.png}
    
    \begin{tabular}{cc}
        \subfloat[\textit{Walker: Run $\to$ Stand}]{
            \centering
            \includegraphics[width=0.3\textwidth]{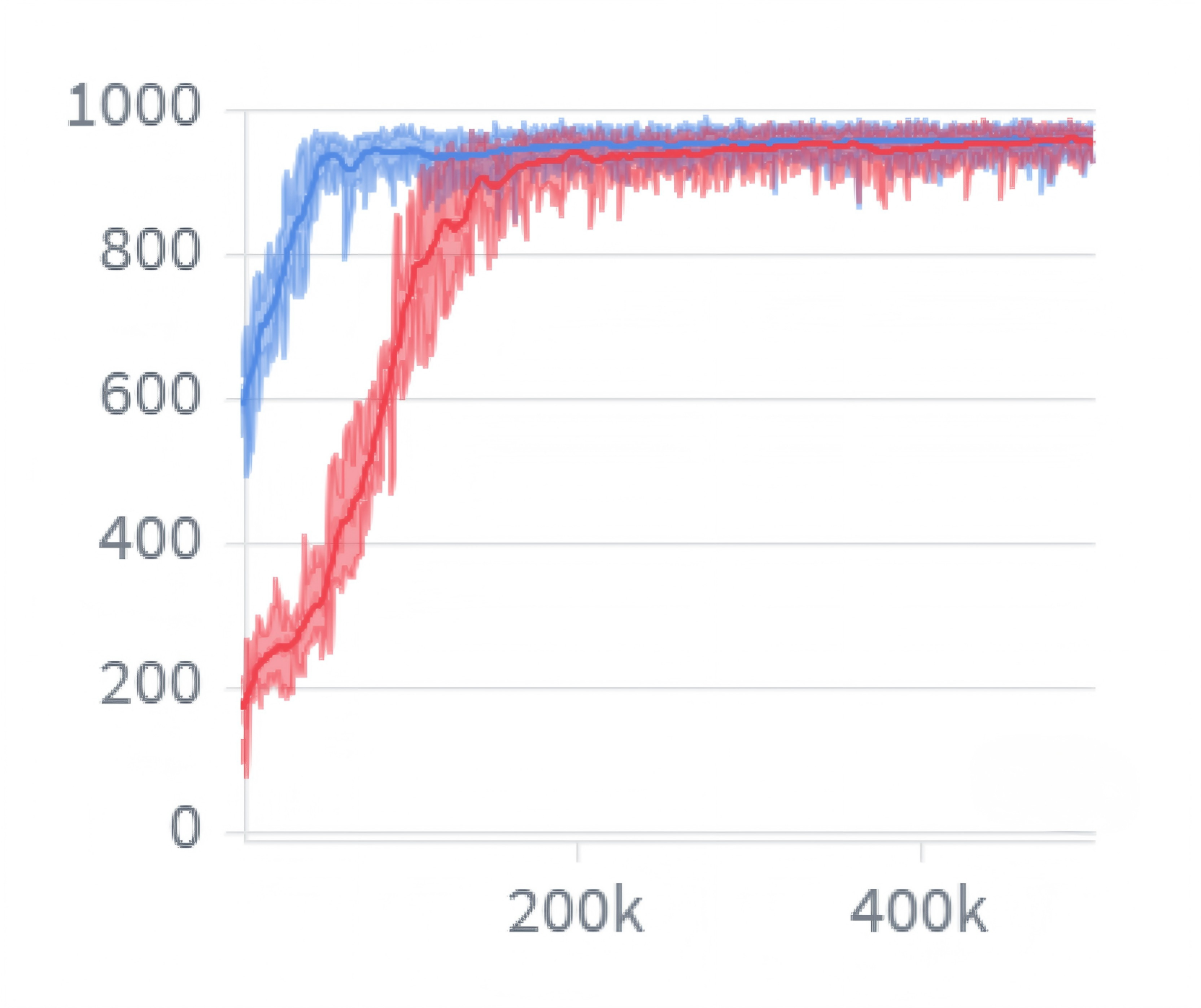}} &
        \subfloat[\textit{Walker: Run $\to$ Walk}]{
            \centering
            \includegraphics[width=0.3\textwidth]{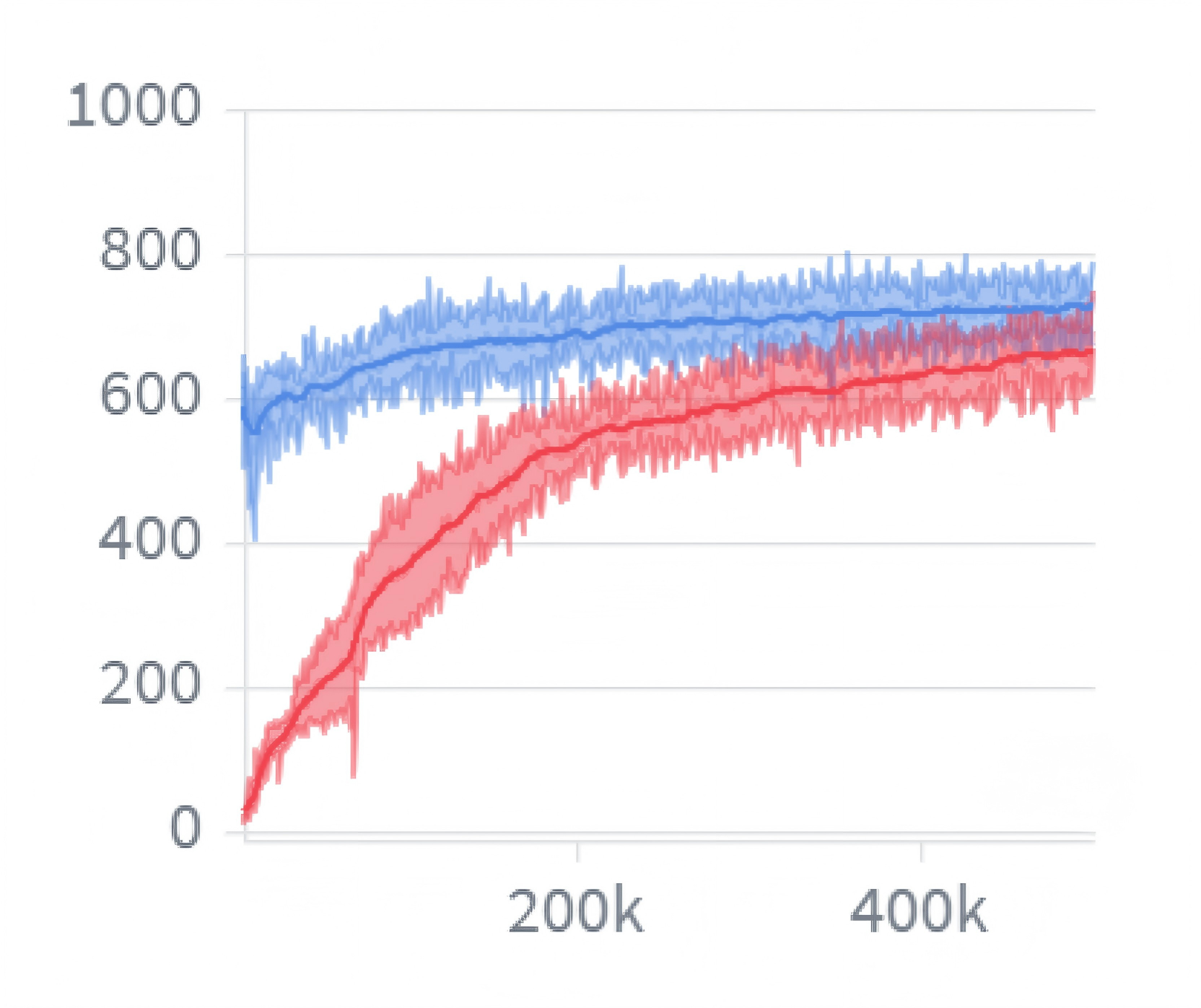}} \\
        \subfloat[\textit{Reacher: Hard $\to$ Easy}]{
            \centering
            \includegraphics[width=0.3\textwidth]{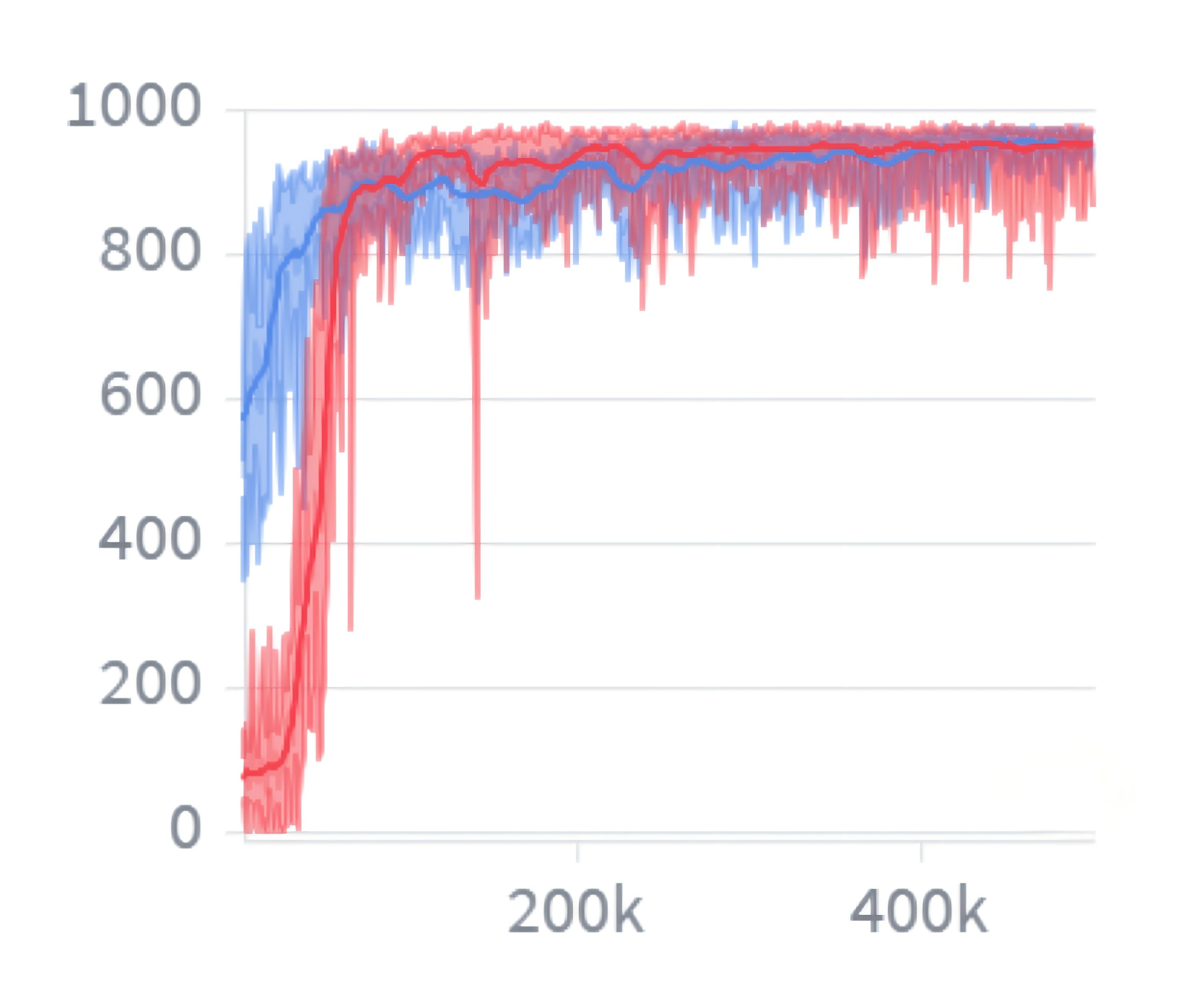}} &
        \subfloat[\textit{Hopper: Hop $\to$ Stand}]{
            \centering
            \includegraphics[width=0.3\textwidth]{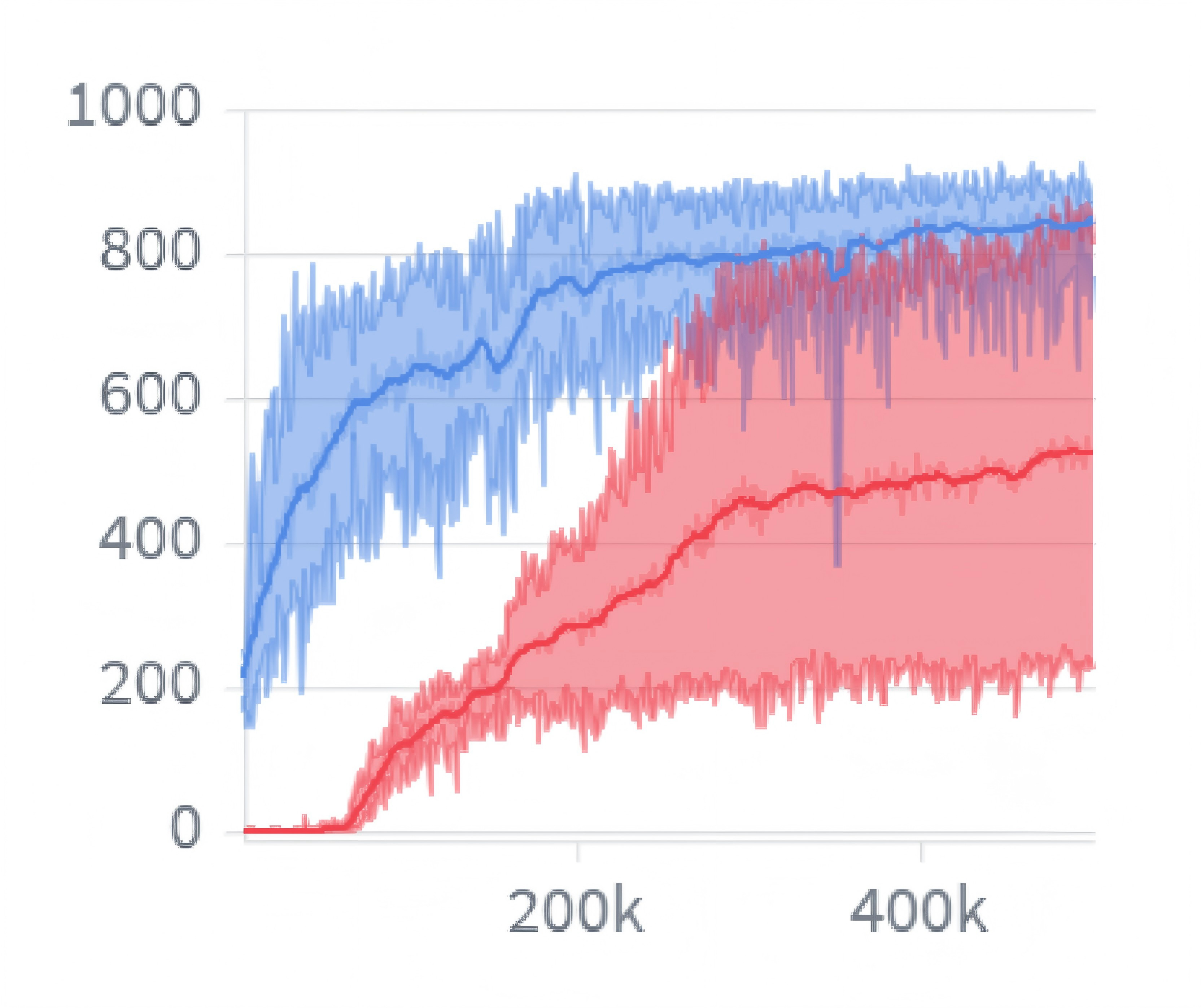}} \\
    \end{tabular}
    \caption{Performance of CausalDreamer in cross-task knowledge reuse from harder tasks to easier tasks}
    \label{fig:abo2_2}
\end{figure}

\textbf{Experimental design\quad}
To examine the cross-task knowledge transfer capability of CausalDreamer, this section selects task pairs from DMC that have consistent dynamics structures but different task difficulties as transfer benchmarks. Specifically, several task pairs are chosen from the same task family with identical environmental dynamics but different objective difficulty, including Stand, Walk, and Run in the Walker family; Easy and Hard in the Reacher family; and Stand and Hop in the Hopper family. In the experiments, the world model is first trained on a source task, and its parameters are then transferred to a target task within the same task family to initialize the target-task world model. The target task is then trained using the same world-model-based imagination-learning process, iterating among world-model updates, trajectory imagination, and policy optimization. The transfer directions include transfer from easier tasks to harder tasks and from harder tasks to easier tasks, in order to evaluate the reuse effect of world-model parameters across tasks of different difficulty. This section compares the transfer-initialized results with the baseline trained from scratch to evaluate knowledge transfer among tasks within the same dynamics family. The results are shown in Figures~\ref{fig:abo2_1} and~\ref{fig:abo2_2}, where the horizontal axis denotes the number of real episodes used for training, and the vertical axis denotes average return.

\textbf{Experimental results\quad}
Figures~\ref{fig:abo2_1} and~\ref{fig:abo2_2} show that, after initializing the target task with world-model parameters trained on the source task, CausalDreamer performs better overall on the target task than the baseline trained from scratch. This is typically reflected in faster early-stage performance improvement, higher sample efficiency, and more stable convergence. This indicates that the knowledge learned from the source task is not limited to a specific policy, but can provide an effective dynamics prior for the target task through world-model parameter transfer. More specifically, when transferring from easier tasks to harder tasks, the advantage is more pronounced, because related tasks often share similar environmental dynamics, while harder tasks impose higher requirements on control precision and policy quality. Existing structural dynamics knowledge can accelerate state modeling, imagination prediction, and policy optimization in the target task. Conversely, when transferring from harder tasks to easier tasks, some tasks may exhibit a brief adaptation phase in early training. However, as the model gradually matches the state distribution of the target task, the transferred prior still brings good convergence behavior. Overall, the world model learned by CausalDreamer contains latent dynamics knowledge that can be reused across related tasks, and field-node encoding together with implicit causal representation provides a more stable representation basis for cross-task parameter transfer.

\section{Conclusion}

Unlike methods that focus on explicit causal modeling for specific delayed tasks, this work targets more general task scenarios and aims to improve the agent's generalization and transfer ability under random delays. To this end, this paper proposes CausalDreamer, a transferable delay-aware reinforcement learning method based on implicit causal graph modeling. The method consists of two modules: an implicit-causal-graph-based world model and world-model-based behavior learning and planning. The former uses a field-node encoder, a causal field-node state-space model, and decoding and prediction modules to perform structured representation learning and environmental dynamics modeling within a unified framework. The latter uses imagined trajectories in latent space for policy optimization, thereby preserving structural knowledge that can be reused within the same dynamics family. Experimental results show that the proposed method generally outperforms baselines such as DreamerV3 on standard tasks without delays, and its performance degradation under random delays is also much smaller. This indicates that the learned representations and dynamic knowledge have good stability. More importantly, the cross-task transfer experiments show that, whether transferring from easier tasks to harder tasks or from harder tasks to easier tasks, the transferred training process generally outperforms training from scratch, and the final performance is usually no worse and sometimes even better. This demonstrates that CausalDreamer does not merely learn local policies tied to a single task, but learns structural knowledge that can be reused within the same dynamics family. Further ablation experiments also show that the field-node encoder and message-passing mechanism play key roles in forming such transferable representations. In summary, the proposed CausalDreamer method can learn stable and transferable structural knowledge in random-delay environments, effectively addressing the difficulty of reusing knowledge across tasks.

\section*{Declaration of generative AI and AI-assisted technologies in the manuscript preparation process}
During the preparation of this work, the authors used Qwen~(\cite{DBLP:journals/corr/abs-2505-09388}) to check grammatical errors and improve the clarity of the text. After using them, the authors reviewed and edited the content as needed and take full responsibility for the content of the publication.

\section*{Acknowledgments}
This work was supported by the National Natural Science Foundation of China (No. 62372459).

\bibliographystyle{elsarticle-num} 
\bibliography{nn_emp}



\end{document}